\documentclass[sigconf]{acmart}

\usepackage{graphicx} 
\usepackage{float} 
\usepackage{amsmath}
\usepackage{amsthm}
\usepackage{multirow,diagbox,enumitem}
\usepackage{arydshln}
\usepackage{graphicx}
\usepackage{wrapfig}
\usepackage{makecell}
\usepackage{subfigure}

\usepackage{algorithm}
\usepackage{algorithmic}

\newtheorem{theorem}{Theorem}

\theoremstyle{remark}
\newtheorem{remark}{Remark}

\usepackage{color, colortbl}
\definecolor{Gray}{rgb}{0.9, 0.9, 0.9}

\newcommand{\model}{LEAP}

\usepackage{titletoc}
\usepackage[page, header, toc, page]{appendix}

\copyrightyear{2026}
\acmYear{2026}
\setcopyright{cc}
\setcctype{by}
\acmConference[KDD 2026] {Proceedings of the 32nd ACM SIGKDD Conference on Knowledge Discovery and Data Mining V.1}{August 9--13, 2025}{Jeju Island, Republic of Korea.}
\acmBooktitle{Proceedings of the 32nd ACM SIGKDD Conference on Knowledge Discovery and Data Mining V.1 (KDD 2026), August 9--13, 2025, Jeju Island, Republic of Korea}
\acmISBN{979-8-4007-2258-5/2026/08}
\acmDOI{10.1145/3770854.3780291}

\begin{document}

\title[LEAP]{Learning and Editing Universal Graph Prompt Tuning via Reinforcement Learning}

\author{Jinfeng Xu}
\email{jinfeng@connect.hku.hk}
\affiliation{%
  \institution{The University of Hong Kong}
  \city{HongKong SAR}
  \country{China}}

\author{Zheyu Chen}
\email{zheyu.chen@bit.edu.cn}
\affiliation{%
  \institution{Beijing Institute of Technology}
  \city{Beijing}
  \country{China}}

\author{Shuo Yang}
\email{shuoyang.ee@gmail.com}
\affiliation{%
  \institution{The University of Hong Kong}
  \city{HongKong SAR}
  \country{China}}

\author{Jinze Li}
\email{lijinze-hku@connect.hku.hk}
\affiliation{%
  \institution{The University of Hong Kong}
  \city{HongKong SAR}
  \country{China}}

\author{Hewei Wang}
\email{heweiw@alumni.cmu.edu}
\affiliation{%
    \institution{Carnegie Mellon University}  
    \city{Pittsburgh, PA}   
    \country{USA}}

\author{Yijie Li}
\email{yijieli@andrew.cmu.edu}
\affiliation{%
  \institution{Carnegie Mellon University}
  \city{Pittsburgh, PA}
  \country{USA}}

\author{Edith C. H. Ngai}
\authornote{Corresponding authors}
\email{chngai@eee.hku.hk}
\affiliation{%
  \institution{The University of Hong Kong}
  \city{HongKong SAR}
  \country{China}}

\renewcommand{\shortauthors}{Jinfeng Xu et al.}


 
\begin{abstract}
The "pre-training, prompt-tuning" has emerged as a pivotal paradigm in advancing the performance of graph representation learning models across a wide range of downstream tasks. This paradigm leverages the power of pre-trained models and task-specific prompts to bridge the gap between general graph representations and task-specific requirements. Early graph prompt tuning approaches relied on task-specific designs for Graph Neural Networks (GNNs), limiting their adaptability across diverse pre-training strategies. In contrast, another promising line of research has investigated universal graph prompt tuning, which operates directly in the input graph's feature space and builds a theoretical foundation that universal graph prompt tuning can theoretically achieve an equivalent effect of any prompting function, eliminating dependence on specific pre-training strategies. Recent works propose selective node-based graph prompt tuning to pursue more ideal prompts. However, we argue that selective node-based graph prompt tuning inevitably compromises the theoretical foundation of universal graph prompt tuning. In this paper, we strengthen the theoretical foundation of universal graph prompt tuning by introducing stricter constraints, demonstrating that adding prompts to all nodes is a necessary condition for achieving the universality of graph prompts. To this end, we propose a novel model and paradigm, \textit{\underline{L}earning and \underline{E}diting Universal Gr\underline{A}ph \underline{P}rompt Tuning} (\model), which preserves the theoretical foundation of universal graph prompt tuning while pursuing more ideal prompts. Specifically, we first build the basic universal graph prompts to preserve the theoretical foundation and then employ actor-critic reinforcement learning to select nodes and edit prompts. Extensive experiments on graph- and node-level tasks across various pre-training strategies in both full-shot and few-shot scenarios show that \model\space consistently outperforms fine-tuning and other prompt-based approaches.
\end{abstract}



\maketitle

\section{Introduction}

In recent years, Graph Neural Networks (GNNs) have demonstrated remarkable success in a wide range of domains, including knowledge graphs \citep{liang2023knowledge,liang2024survey}, social networks \citep{peng2021streaming,cao2024hierarchical}, and recommender systems \citep{xu2024aligngroup,xu2025mentor,xu2025cohesion}, due to their ability to model complex relationships in real-world data effectively. Despite their expressive power, traditional supervised learning on graphs faces two significant challenges: the scarcity of labeled data and poor generalization to out-of-distribution samples \citep{yehudai2021local}. To address these issues, the "\textit{pre-training, fine-tuning}" paradigm has been applied in GNNs \citep{wu2021self}, which leverages self-supervised learning on unlabeled graphs to capture intrinsic structural and semantic properties. These models are then fine-tuned on downstream tasks to achieve outstanding performance for specific downstream tasks. Despite significant progress, the "\textit{pre-training, fine-tuning}" paradigm still suffers from notable limitations. A key challenge is the misalignment between pre-training objectives and downstream tasks, often resulting in sub-optimal performance, a phenomenon referred to as negative transfer \citep{han2021adaptive,li2024adaptergnn}. Additionally, pre-trained models tend to overfit in the few-shot scenario, leading to catastrophic forgetting \citep{liu2021overcoming}, severely hampering their generalization ability.

To address these issues, \textit{"pre-training, prompt-tuning"} has emerged as a transformative paradigm, demonstrating remarkable success in computer vision (CV) \citep{jia2022visual,yao2023visual,xu2025multi} and natural language processing (NLP) \citep{brown2020language,liu2023pre}. By reformulating downstream tasks to better align with the objectives of pre-trained models, prompt learning effectively bridges the gap between pre-training and fine-tuning, enhancing task performance and generalization ability. The existing "\textit{pre-training, prompt-tuning}" approaches are broadly divided into two categories: \textbf{Task-specific Graph Prompt Tuning} and \textbf{Universal Graph Prompt Tuning}. For task-specific graph prompt tuning approaches, GPPT \citep{sun2022gppt} and GraphPrompt \citep{liu2023graphprompt} align pretext and downstream tasks with edge prediction, while All in One \citep{sun2023all} uses a prompt graph for graph-level contrastive learning. SGL-PT \citep{zhu2023sgl} connects downstream tasks with masked node prediction via contrastive and generative objectives. However, these approaches may struggle when GNNs are pre-trained with diverse self-supervised tasks through multi-task learning, rather than simple link prediction. To address these challenges, universal graph prompt tuning approaches, which are agnostic to specific pre-training strategies, offer enhanced compatibility and flexibility. GPF \citep{fang2023universal}, a seminal work in this direction, introduces universal graph prompt tuning, which operates directly in the input graph's feature space and builds a theoretical foundation that universal graph prompt tuning can theoretically achieve an equivalent effect of any prompting function, eliminating dependence on specific pre-training strategies. Building on this, GPF-plus \citep{fang2023universal} extends the GPF by incorporating node-specific prompted features, enabling more adaptable and fine-grained tuning. Recent works explore selective node-based graph prompt tuning to achieve more ideal prompts. SUPT \citep{lee2024subgraph} assigns feature prompts at the subgraph level, capturing nuanced contextual information within subgraphs and achieving substantial performance gains. RELIEF \citep{zhu2024relief} takes a reinforcement learning approach to dynamically select specific nodes for prompt addition and design dynamically, further pursuing high-quality prompt tuning. Although these selective node-based graph prompt tuning approaches achieve certain performance improvements, we point out that selective node-based graph prompt tuning undermines the theoretical foundation (as discussed in detail in Section~\ref{sec:preliminary}). Therefore, selective node-based graph prompt tuning inevitably constrains the theoretical representational capacity of universal graph prompt tuning. 

Undoubtedly, "\textbf{pursuing more ideal prompts without undermining its theoretical foundation}" is the next step for universal graph prompt tuning. To this end, we propose a novel model and paradigm, \textit{\underline{L}earning and \underline{E}diting Universal Gr\underline{A}ph \underline{P}rompt Tuning} (\model), which preserves the theoretical foundation of universal graph prompt tuning while pursuing more ideal prompts. Specifically, we first incorporate distinct learnable prompts to all nodes in the graph to ensure that the theoretical foundation of universal graph prompt tuning remains intact\footnote{Incorporating distinct learnable prompts offers greater flexibility by allowing for different prompted features to be provided for each node. It possesses a larger set of feasible solutions to approximate a specific prompting function than incorporating fixed learnable prompts \citep{fang2023universal}.}. Subsequently, we further refine the prompts by selecting specific nodes for additional prompt editing. Inspired by RELIEF, we formulate the selection and editing process as a sequential decision-making problem. To address this, we leverage an actor-critic reinforcement learning approach to handle the discrete-continuous hybrid action space. Our goal is to maximize the expected cumulative performance gain on the downstream task. Moreover, to ensure that the editing process does not repeatedly focus on a small subset of nodes, which could compromise the model's generalization ability, we incorporate the editing convergence rate as an extra measurement in the reward function. Furthermore, we conduct extensive experiments on graph- and node-level tasks across various pre-training strategies in both full-shot and few-shot scenarios to validate the efficacy of our \model\space consistently outperforms fine-tuning and other prompt-based approaches. Overall, the contributions can be summarized as follows:
\begin{itemize}[leftmargin=*]
    \item To the best of our knowledge, we first highlight that adding prompts to all nodes is a stricter theoretical constraint for universal graph prompt tuning to achieve an equivalent effect to any form of prompting function. This constraint provides a significant foundational guideline for correcting the future research direction of universal graph prompt tuning.
    \item We propose a novel model, Learning and Editing Universal Graph Prompt Tuning (\model), which preserves the theoretical foundation of universal graph prompt tuning while pursuing more ideal prompts. More importantly, LEAP establishes an ideal paradigm for universal graph prompt tuning.
    \item Extensive experiments on both graph- and node-level tasks in both full-shot and few-shot scenarios demonstrate that \model\space achieves superior performance to fine-tuning and other prompt-based approaches with various pre-training strategies.
\end{itemize}

\section{Preliminary}
\label{sec:preliminary}
Let $\mathcal{G}=(\mathcal{V}, \mathcal{E}, \mathbf{X}) \in \mathbb{G}$ represents  an undirected graph instance, where $\mathcal{V}=\{v_1, v_2, \ldots, v_N\}$ and $\mathcal{E} = \{(v_i, v_j) \mid v_i, v_j \in \mathcal{V}\}$ denote the node set and edge set, respectively. $\mathbf{X}=\left\{x_1, x_2, \ldots, x_N\right\} \in \mathbb{R}^{N \times D}$ denotes the node feature matrix, where $x_i \in \mathbb{R}^{1 \times D}$ denotes the feature vector of node $v_i$ with feature dimension $D$. $\mathbf{A} \in\{0,1\}^{N \times N}$ denotes the adjacency matrix, where $\mathbf{A}_{i j}=1$ if $\left(v_i, v_j\right) \in \mathcal{E}$.

\noindent \textbf{Graph Fine-tuning} Given a pre-trained graph model $f_\theta(\cdot)$ and a \textit{task-specific} projection head $g_\phi(\cdot)$, $\theta$ and $\phi$ are \textit{learnable} and fine-tuned on a downstream training dataset $\mathcal{D}=\{(\mathcal{G}_1, y_1), \ldots,(\mathcal{G}_m, y_m)\}$ to maximize the likelihood of predicting the correct labels $y$ of graph $\mathcal{G}$, formally: 
\begin{equation}
\max _{\theta, \phi} \prod_{i=0}^m P_{f_\theta, g_\phi}(y_i | \mathcal{G}_i) = \max _{\theta, \phi} \prod_{i=0}^m P(y_i | g_\phi(f_\theta(\mathbf{X},\mathbf{A}))).
\end{equation}

\noindent \textbf{Graph Prompt-tuning} Given a \textit{frozen} pre-trained model $f_\theta(\cdot)$ and a \textit{learnable} \textit{task-specific} prompt function $\psi(\cdot): \mathbb{G} \rightarrow \mathbb{G}$, which transforms the original graphs $\mathcal{G}$ to prompt graphs $\mathcal{\hat{G}} = \psi(\mathcal{G})$. The \textit{task-specific} projection head $g_\phi(\cdot)$ is also learnable and tuned to coordinate with the prompting function to maximize the likelihood of predicting the correct labels $y$ of graph $\mathcal{G}$, formally: 
\begin{equation}
\max _{\psi, \phi} \prod_{i=0}^m P_{f_\theta, g_\phi}(y_i | \psi(\mathcal{G}_i)) = \max _{\psi, \phi} \prod_{i=0}^m P(y_i | g_\phi(f_\theta(\mathbf{\hat{X}},\mathbf{\hat{A}}))),
\end{equation}
where $\mathbf{\hat{X}}$ and $\mathbf{\hat{A}}$ are the feature matrix and adjacency matrix, respectively, prompted by $\psi(\cdot)$.

\begin{remark}Fine-tuning approaches adjust both the pre-trained graph model $f_\theta(\cdot)$ and the projection head $g_\phi(\cdot)$ to fit downstream tasks, but they suffer from negative transfer \citep{han2021adaptive} and catastrophic forgetting \citep{liu2021overcoming} problems. To this end, prompt-tuning approaches transform graphs $\mathcal{G}$ using a prompting function $\psi(\cdot)$ and then tune both the projection head $g_\phi(\cdot)$ and the prompting function $\psi(\cdot)$ to fit downstream tasks. However, this type of approach remains challenging and varies across different pre-training tasks \citep{fang2023universal}.
\end{remark}

\noindent \textbf{Universal Graph Prompt-tuning} GPF is a seminal work that first proposes universal graph prompt tuning by demonstrating that adding a learnable, uniform feature prompt vector to each node is theoretically equivalent to any form of graph manipulation, independent of the pre-training strategy. Given the node feature matrix $\mathbf{X}$ of a graph $\mathcal{G}$, a learnable prompt feature vector $p \in \mathbb{R}^{D}$ is added to node features. This can be viewed as the prompting function $\psi(\cdot)$, achieving a prompted feature matrix $\mathbf{X}^{*}$, formally: $\mathbf{X}^*=\{x_1+p, x_2+p, \ldots, x_N+p\}=\{x_1^*, x_2^*, \ldots, x_N^*\}$. GPF provides rigorous derivations to prove that universal graph prompt tuning is theoretically equivalent to any form of graph manipulation, formally: $f_\theta(\mathbf{\hat{X}}, \mathbf{\hat{A}}) = f_\theta(\mathbf{X}+p, \mathbf{A})$. Moreover, GPF-plus is a more powerful version of GPF by assigning an independent learnable vector $p_i$ to each node $v_i$ in the graph, formally: $\mathbf{X}^*=\{x_1+p_1, x_2+p_2, \ldots, x_N+p_N\}=\{x_1^*, x_2^*, \ldots, x_N^*\}$. Cause GPF-plus can degenerate to GPF when all vectors $p_i$ are the same. Therefore, theoretical foundation $f_\theta(\mathbf{\hat{X}}, \mathbf{\hat{A}}) = f_\theta(\mathbf{X}+p, \mathbf{A})$ is also applicable for GPF-plus. Although universal graph prompt tuning has a solid theoretical foundation, the challenge of effectively training an ideal prompt remains unresolved. Recent works propose selective node-based graph prompt tuning, where SUPT and RELIEF use MLP and reinforcement learning, respectively, to select nodes for adding prompts. However, we argue that such practices undermine the theoretical foundation of universal graph prompt tuning.

\begin{theorem}
\label{theorem:1}
\textbf{(Necessity of Adding Prompts to All Nodes)}
For any prompting function $\psi(\cdot): \mathbb{G} \rightarrow \mathbb{G}$, there exists a set of node-wise prompt vectors $\{p_i \in \mathbb{R}^D\}_{i=1}^N$ such that $f_\theta(\mathbf{\hat{X}}, \mathbf{\hat{A}}) = f_\theta(\mathbf{X}+p, \mathbf{A})$, \textbf{if and only if} each node $v_i \in \mathcal{V}$ is added a prompt $p_i$ (allowing $p_i \neq p_j$ or $p_i=p_j$).
\end{theorem}

The complete proof of Theorem~\ref{theorem:1} can be found in Appendix~\ref{appendix:proofs}. This theorem extends the original theorem in GPF by emphasizing that per-node prompting is not optional—omitting prompts for any node strictly reduces the expressiveness of universal graph prompt tuning. Furthermore, it provides a significant guideline for refining the future research directions of universal graph prompt tuning.

\section{Methodology}
Based on the stricter constraints for universal graph prompt tuning presented at the end of Section~\ref{sec:preliminary}, we clearly see that ``\textbf{pursuing more ideal prompts without undermining its theoretical foundation}" is the next step for universal graph prompt tuning. In this section, we introduce a basic universal and learnable initial graph prompt to preserve the theoretical foundation of the framework.  We further propose an actor-critic reinforcement learning framework to select and edit some prompts dynamically. Note that we use graph-level classification as an example throughout this section. Our LEAP can be extended to node-level tasks by incorporating prompts into the $n$-hop subgraph induced by the target node \citep{sun2022gppt,liu2023graphprompt}.

\subsection{Basic Learnable Universal Graph Prompt}
To preserve the theoretical foundation of universal graph prompt tuning, we first adopt a parameter-efficient variant of GPF-plus as the basis for learnable universal graph prompts. Specifically, we train only $k$ independent basis prompt vectors $p^b_{1}, p^b_{2}, \dots, p^b_{k}$ and employ an attentive aggregation mechanism, supported by $k$ learnable linear projections $a$, to generate prompts for all nodes. Formally:
\begin{equation}
\resizebox{0.9\hsize}{!}{$\begin{aligned}
p_i=\sum_{j=1}^k \alpha_{i, j} p_j^b, \quad \alpha_{i, j}=\frac{\exp \left(a_j^{\mathrm{T}} x_i\right)}{\sum_{l=1}^k \exp \left(a_l^{\mathrm{T}} x_i\right)}, \quad p_1^b, p_2^b, \ldots p_k^b \in \mathbb{R}^D,
\end{aligned}$}
\end{equation}
where $p_i$ is generated prompt for node $v_i$. $k$ is a hyper-parameter that can be adjusted based on the downstream dataset. In Section~\ref{subsec:additional_experiments}, we provide an empirical discussion on the selection of GPF or full GPF-plus as the basic learnable universal graph prompts.

\subsection{Selecting and Editing Universal Graph Prompt}
Inspired by RELIEF, we formulate the selecting and editing process as a sequential decision-making problem and employ an actor-critic reinforcement learning approach, H-PPO \citep{schulman2017proximal}, to handle the discrete-continuous hybrid action space. Specifically, we use discrete actors to select nodes and continuous actors to edit them. The state space is defined as the node representations of the prompted graph $\mathcal{\hat{G}}$, which are derived from the pre-trained graph model $f_\theta(\cdot)$. Additionally, the reward function is formulated based on loss reduction and editing coverage, with a critic employed to estimate the state value.

\subsubsection{Actors}
Since LEAP utilizes H-PPO, which consists of two parallel actor networks and a single critic network, collectively referred to as the policy network $\Pi_\varsigma(\cdot)$, where $\varsigma$ denotes the parameters of these networks.

\textbf{Discrete Actor:} The purpose of the discrete actor is to select a node $v_a$ from the node set $\mathcal{V}$. In LEAP, given a state $\mathbf{s} \in \mathbb{R}^{N \times D}$, we employ a shallow MLP as an encoder $\operatorname{Enc}_{d}(\cdot)$ and apply the Softmax function to transform its output into a discrete action probability. Formally: $p(a|\mathbf{s}) = \operatorname{Softmax}(\operatorname{Enc}_{d}(\mathbf{s}))$. Then we sample a node $v_a$ with the highest probability.

\textbf{Continuous Actor:} The purpose of the continuous actor is to generate a feature vector $f_a \in \mathbb{R}^{1 \times D}$ to edit the basic universal graph prompt for node $v_a$, which is selected by the discrete actor. In LEAP, given a state $\mathbf{s} \in \mathbb{R}^{N \times D}$, we adopt a shallow MLP as an encoder $\operatorname{Enc}_{c}(\cdot)$ to encode all nodes. The feature vector $z$ corresponding to the selected node $v_a$ is then used to edit the basic universal graph prompt. Formally: $z = \operatorname{Enc}_{c}(\mathbf{s})_{[a,:]}$. Due to the inherent uncertainty of the action, we get the final feature vector $f_a$ by stochastically resampling $z$ from a Gaussian distribution $\mathcal{N}(z, \sigma)$, where $\sigma$ can either be a learnable parameter or predefined. Additionally, $f_a$ is clamped to a desired range $[-\vartheta, \vartheta]$, where $\vartheta$ is a hyper-parameter to control the scale of the prompt edited at each step.

Overall, each step within the discrete-continuous hybrid action space generates a feature vector $f_a$ for the selected node $v_a$. Then $f_a$ is utilized to edit the basic universal graph prompt, formally: $p_a \leftarrow p_a + f_a$. The entire universal graph prompt matrix at step $t$ is expressed as $p^t=\{p_1^t,\dots,p_N^t\}$.

\subsubsection{State} 
The state space is defined as the node representations of the prompted graph $\mathcal{\hat{G}}$, which are derived from the pre-trained graph model $f_\theta(\cdot)$. Note that for each step $t$, the prompted graph $\mathcal{\hat{G}}$ is changed by prompting the node feature matrix $\mathbf{X}$ via $\mathbf{X} + p^{t-1}$. This enables the agent to account for node representations that are closely linked to the pretrained graph model, allowing for precise control. Formally, the state at time step $t$ is denoted as: $\mathbf{s}^t = f_\theta(\mathcal{\hat{G}}) = f_\theta(\mathbf{X} + p^{t-1}, \mathbf{A})$. Note that the state at time step $t$ is built upon the edited universal graph prompt matrix at time step $t-1$.

\subsubsection{Reward}
We aim to design a reward function that provides guiding signals for action values during exploration while being adaptable to all downstream tasks in universal graph prompt tuning. The most intuitive approach is to use loss reduction as our instant reward, as it directly reflects performance improvement and remains task-agnostic. However, relying solely on loss reduction may result in insufficient guiding signals, causing the actors to repeatedly focus on a small subset of nodes, which can harm the model's generalization ability. To this end, we cleverly introduce the editing convergence rate as an additional measure in the reward function. Formally, given two adjacent steps, the reward $r^t$ is expressed as:
\begin{equation}
\label{eq:reward}
\resizebox{0.92\hsize}{!}{$\begin{aligned}
r^t=\lambda_{e} \operatorname{ECR}^t + \mathcal{L}(g_\phi(f_\theta(\mathbf{X} + p^{t-1}, \mathbf{A})), y)-\mathcal{L}(g_\phi(f_\theta(\mathbf{X} + p^{t}, \mathbf{A})), y),
\end{aligned}$}
\end{equation}
where $\mathcal{L}(\cdot)$ denotes a loss function associated with the downstream task. $\operatorname{ECR}^t$ denotes the editing convergence rate at time step $t$. $\lambda_{e}$ is a balancing hyper-parameter. To calculate the editing convergence rate, we maintain a count vector $\mathbf{c} \in \mathbb{R}^{N}$, which is initialized as a zero vector. At each step $t$, the actors select and edit the prompt $p_a$ for node $v_a$. The count vector $\mathbf{c}^t$ is updated as follows: $\mathbf{c}_a = \mathbf{c}_a + 1$, where $\mathbf{c}_a$ denotes the edit count for node $v_a$. $\operatorname{ECR}^t$ is calculated as:
\begin{equation}
\operatorname{ECR}^t=\frac{\sum_{i=1}^N \mathbb{I}\left(\mathbf{c}_i^t \neq 0\right)}{N},
\end{equation}
where $\mathbb{I}(\mathbf{c}_i \neq 0)$ is an indicator function that equals $1$ if $\mathbf{c}_i \neq 0$, and $0$ otherwise. In Section~\ref{subsec:additional_experiments}, we validate the effectiveness of incorporating the editing convergence rate through ablation studies. 

\subsubsection{Critic}
The critic is designed for estimating the state-value function. Given a state $\mathbf{s}^t \in \mathbb{R}^{N \times D}$ at time step $t$. We first apply a flattening operation, $\operatorname{Flat}(\cdot)$, which preserves node-level spatial correlations for the state while enabling graph-level value estimation, formally: $\mathbf{\tilde{s}}^t=\operatorname{Flat}(\mathbf{s}^t) \in \mathbb{R}^{1 \times N D}$. Then, the flattened state vector $\mathbf{\tilde{s}}^t$ is passed through a shallow MLP $\operatorname{Enc}_e(\cdot)$ to produce a real value, treated as the estimation $V_\varphi(\mathbf{s}^t)$, formally: $V_\varphi(\mathbf{s}^t) = \operatorname{Enc}_e(\mathbf{\tilde{s}}^t)$, where $\varphi$ represents parameters of critic network, which is included in $\varsigma$.

\begin{figure*}[!th]
    \centering
    \includegraphics[width=0.9\linewidth]{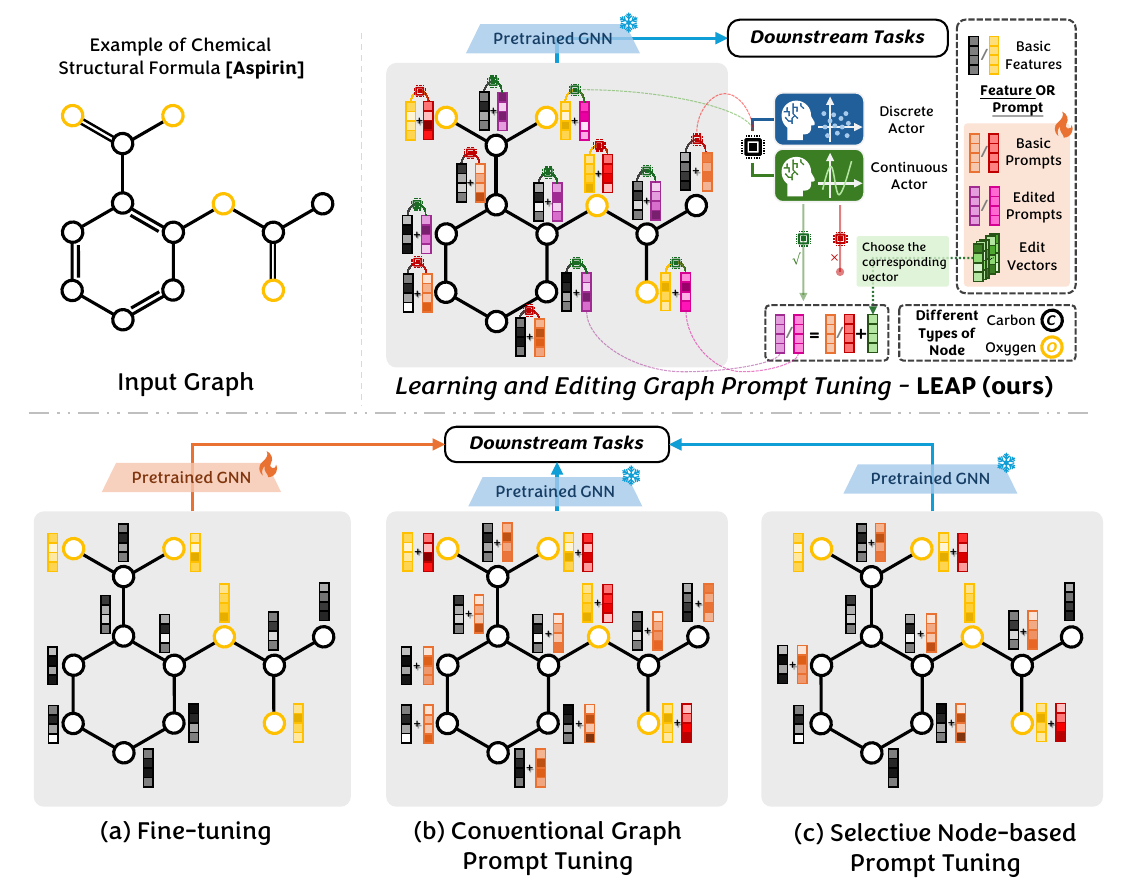}
    \caption{Different paradigms for universal graph prompt tuning and basic fine-tuning. a) Fine-tuning, b) Conventional Graph Prompt Tuning, c) Selective Node-based Prompt Tuning, and d) Learning and Editing Graph Prompt Tuning.} 
    \label{fig:overview}
\end{figure*}

\subsection{Training}
Pre-trained graph model $f_\theta(\cdot)$ is frozen. We represent the policy networks for the discrete actor and continuous actor as $\pi_{\varsigma_d}(\cdot)$ and $\pi_{\varsigma_c}(\cdot)$, respectively. Thus, we optimize policy networks $\pi_{\varsigma_d}(\cdot)$ and $\pi_{\varsigma_c}(\cdot)$ of actors, policy network of critic $V_\varphi(\cdot)$, projection head $g_\phi(\cdot)$, basic universal graph prompt $p^b$ and linear projections $a$.

\subsubsection{Policy Networks Training}
The two actors are independently trained using the PPO surrogate objectives\footnote{Due to space constraints, we provide an introduction to PPO in Appendix~\ref{appendix:ppo}.} $\mathcal{L}_{d}^{\mathrm{PPO}}$ and $\mathcal{L}_{c}^{\mathrm{PPO}}$, respectively. At each time step $t$, we repeat $T-t$ steps and collect $T-t$ rewards using Eq.~\ref{eq:reward} to compute the discounted cumulative return, formally:
\begin{equation}
R^t=\sum_{l=0}^{T-t} \gamma^l r^{t+l},
\end{equation}
where $T$ is a hyper-parameter representing the finite horizon, $\gamma^l$ is the discount factor. Then, we adopt MSE loss $\mathcal{L}_{C}$ to optimize policy networks of critic $V_\varphi(\cdot)$, formally:
\begin{equation}
\mathcal{L}_{C}=\frac{1}{T} \sum_{t=1}^T(V_{\varphi}(\mathbf{s}^t)-R^t)^2.
\end{equation}
Policy networks training is batchified by processing a batch of graphs simultaneously to improve sampling and training efficiency.

\subsubsection{Projection Head and Basic Prompt Training}
We optimize the projection head and basic prompt by aligning predictions with their correct labels, coordinating the projection head with the representations of the final prompted graphs. We set the standard deviation in the continuous policy to $0$, ensuring that the prompt vector values remain identical for the same state and discrete action. This ensures stable prompted graphs and consistent representations, along with the labels, are used to guide the update of the projection heads $a$ and basis prompt vectors $p^b$. Given a downstream training dataset $\mathcal{D}=\left\{\left(\mathcal{G}_1, y_1\right), \ldots,\left(\mathcal{G}_m, y_m\right)\right\}$, projection head $g_\phi(\cdot)$, basic prompt vectors $p^b$ and linear projections $a$ are updated by:
\begin{equation}
\min _{\phi,p^b,a} \frac{1}{m} \sum_{i=1}^m \mathcal{L}(g_\phi(f_\theta(\mathcal{\hat{G}}_i)), y).
\end{equation}

\begin{remark}
Since our learnable basic universal graph prompt shares the same optimization objective as the RL algorithm and mutually enhances the learning process, LEAP maintains strong generalization capabilities in few-shot prompt tuning scenarios without suffering from the overfitting issues commonly observed in the RL algorithm \citep{raileanu2021decoupling,wang2020improving,ghosh2021generalization}. As a result, we deliberately omit policy generalization techniques. To further avoid overfitting and accelerate training, we update the policy networks every $h$ times. 
\end{remark}


\subsection{Training Algorithm}
\label{appendix:algorithm}
For clarity, we provide LEAP's algorithm and complexity analysis. LEAP contains three learnable components: Basic Universal Graph Prompt ($p^b$ and $a$), Policy Networks ($\Pi_\varsigma$), and Projection Head ($g_\phi(\cdot)$). Notably, Policy Networks $\Pi_\varsigma$ includes discrete actor $\pi_{\varsigma_d}(\cdot)$, continuous actor $\pi_{\varsigma_c}(\cdot)$, and critic $V_\varphi(\cdot)$. For simplicity, we use $\varsigma$ to represent the parameters of all the actors and
the critic. The pseudo-code for LEAP is presented in Algorithm~\ref{algorithm:training}. Notably, the operations in line $7$ can be batchified by consuming a batch of graphs.

\begin{algorithm}
\caption{Process of LEAP}
\label{algorithm:training}
\begin{algorithmic} [1] 
\STATE \textbf{Input:} The frozen pre-trained graph model $f_\theta(\cdot)$, Maximum number of nodes $N$, Training graphs $\mathcal{D}=\{(\mathcal{G}_1, y_1), \ldots,(\mathcal{G}_m, y_m)\}$, Policy networks update interval $h$, Training epochs $E$, Policy finite horizon (step) $T$.
\STATE \textbf{Output:} Optimal basic universal graph prompt $p^b$ and $a$, Optimal policy networks $\Pi_\varsigma$, Optimal projection head $g_\phi(\cdot)$.
\STATE Initialize $p^b$, $a$, $\Pi_\varsigma$, $g_\phi(\cdot)$;
\FOR{$epoch=1$ to $E$}
    \STATE Calculate basic universal graph prompt $p$ via $p^b$ and $a$;
    \IF{$epoch$ $\%$ $h = 0$}
        \FOR{$ (\mathcal{G}_i, y_i) \in \mathcal{D}$} 
            \STATE Define $p^0 = p$;
            \FOR{$t=1$ to $T$} 
                \STATE Calculate prompted graph $\mathcal{\hat{G}}$;
                \STATE Construct state $\mathbf{s}^t$;
                \STATE Select node and edit prompt $p^{t-1}$ as $p^t$;
                \STATE Compute reward $r^t$;
            \ENDFOR
            \STATE Update policy networks $\Pi_\varsigma$ and basic universal graph prompt $p$;
        \ENDFOR
    \ENDIF
    \STATE Update basic universal graph prompt $p^b$, $a$, and projection head $g_\phi(\cdot)$;
\ENDFOR

\end{algorithmic}
\end{algorithm}

\section{Discussion}
We highlight that LEAP is not only an effective model but also establishes an ideal paradigm for universal graph prompt tuning. Previous works are categorized into the Conventional Graph Prompt Tuning paradigm, which preserves the theoretical foundation but lacks effective prompt optimization, and the Selective Node-based Graph Prompt Tuning paradigm, which refines prompts but undermines the theoretical foundation. LEAP introduces an ideal paradigm, the \textit{Learning and Editing Graph Prompt Tuning} paradigm, which ensures the theoretical foundation of universal graph prompt tuning while pursuing more ideal prompts. To facilitate a better understanding, we illustrate these paradigms in Figure~\ref{fig:overview}. Detailed discussions are provided in Appendix~\ref{appendix:discussion}.

\begin{table*}[!htbp]
  \caption{ROC-AUC (\%) and standard deviation for graph classification on molecule property prediction benchmark under full-shot scenario with various pre-training and tuning strategies.}
  \vskip -0.1in
  \small
      \centering
      \begin{tabular}{ccccccccccc}
        \toprule
        & \makecell{Tuning \\ Strategy} & BBBP & Tox21 & ToxCast & SIDER & ClinTox & MUV & HIV & BACE & \textbf{Avg.} \\
    
        \midrule
    
        \multirow{7}{*}{\rotatebox{90}{Infomax}} &
        FT & 
        \underline{74.61} {\scriptsize \textcolor{gray}{$\pm$1.00}} &
        83.09 {\scriptsize \textcolor{gray}{$\pm$0.48}} &
        67.03 {\scriptsize \textcolor{gray}{$\pm$0.69}} &
        65.54 {\scriptsize \textcolor{gray}{$\pm$0.65}} &
        71.83 {\scriptsize \textcolor{gray}{$\pm$1.94}} &
        \textbf{81.69} {\scriptsize \textcolor{gray}{$\pm$1.29}} &
        77.58 {\scriptsize \textcolor{gray}{$\pm$0.82}} &
        82.19 {\scriptsize \textcolor{gray}{$\pm$1.33}} & 
        75.45 \\
        & GPF &
        74.19 {\scriptsize \textcolor{gray}{$\pm$0.78}} &
        83.33 {\scriptsize \textcolor{gray}{$\pm$0.39}} &
        67.62 {\scriptsize \textcolor{gray}{$\pm$0.79}} &
        67.22 {\scriptsize \textcolor{gray}{$\pm$0.91}} &
        74.94 {\scriptsize \textcolor{gray}{$\pm$3.09}} &
        81.30 {\scriptsize \textcolor{gray}{$\pm$1.45}} &
        77.05 {\scriptsize \textcolor{gray}{$\pm$0.65}} &
        84.01 {\scriptsize \textcolor{gray}{$\pm$1.01}} & 
        76.21 \\
        & GPF-plus &
        74.42 {\scriptsize \textcolor{gray}{$\pm$0.56}} &
        \underline{83.40} {\scriptsize \textcolor{gray}{$\pm$0.55}} &
        \underline{67.80} {\scriptsize \textcolor{gray}{$\pm$0.41}} &
        \underline{67.32} {\scriptsize \textcolor{gray}{$\pm$0.84}} &
        75.56 {\scriptsize \textcolor{gray}{$\pm$2.64}} &
        81.66 {\scriptsize \textcolor{gray}{$\pm$1.72}} &
        \underline{77.70} {\scriptsize \textcolor{gray}{$\pm$1.11}} &
        84.20 {\scriptsize \textcolor{gray}{$\pm$1.55}} & 
        \underline{76.51} \\
        & SUPT\textsubscript{soft} &
        74.55 {\scriptsize \textcolor{gray}{$\pm$0.58}} &
        83.20 {\scriptsize \textcolor{gray}{$\pm$0.91}} &
        67.32 {\scriptsize \textcolor{gray}{$\pm$0.91}} &
        67.27 {\scriptsize \textcolor{gray}{$\pm$1.04}} &
        \underline{75.86} {\scriptsize \textcolor{gray}{$\pm$2.80}} &
        \textbf{81.69} {\scriptsize \textcolor{gray}{$\pm$1.50}} &
        77.42 {\scriptsize \textcolor{gray}{$\pm$1.02}} &
        84.20 {\scriptsize \textcolor{gray}{$\pm$2.13}} & 
        76.44 \\
        & SUPT\textsubscript{hard} &
        73.99 {\scriptsize \textcolor{gray}{$\pm$1.19}} &
        83.36 {\scriptsize \textcolor{gray}{$\pm$0.72}} &
        67.51 {\scriptsize \textcolor{gray}{$\pm$0.80}} &
        67.23 {\scriptsize \textcolor{gray}{$\pm$1.11}} &
        75.60 {\scriptsize \textcolor{gray}{$\pm$2.29}} &
        81.46 {\scriptsize \textcolor{gray}{$\pm$1.54}} &
        77.31 {\scriptsize \textcolor{gray}{$\pm$0.99}} &
        \underline{84.27} {\scriptsize \textcolor{gray}{$\pm$1.20}} & 
        76.34 \\
        & RELIEF &
        74.18 {\scriptsize \textcolor{gray}{$\pm$0.40}} &
        83.30 {\scriptsize \textcolor{gray}{$\pm$0.31}} &
        67.30 {\scriptsize \textcolor{gray}{$\pm$0.35}} &
        67.02 {\scriptsize \textcolor{gray}{$\pm$0.49}} &
        75.29 {\scriptsize \textcolor{gray}{$\pm$1.03}} &
        80.02 {\scriptsize \textcolor{gray}{$\pm$0.52}} &
        76.84 {\scriptsize \textcolor{gray}{$\pm$0.68}} &
        83.97 {\scriptsize \textcolor{gray}{$\pm$0.69}} & 
        75.99 \\
        & LEAP &
        \textbf{74.89} {\scriptsize \textcolor{gray}{$\pm$0.92}} & 
        \textbf{83.65} {\scriptsize \textcolor{gray}{$\pm$0.69}} & 
        \textbf{67.99} {\scriptsize \textcolor{gray}{$\pm$0.77}} & 
        \textbf{67.50} {\scriptsize \textcolor{gray}{$\pm$0.81}} & 
        \textbf{75.98} {\scriptsize \textcolor{gray}{$\pm$1.66}} & 
        \underline{81.49} {\scriptsize \textcolor{gray}{$\pm$1.44}} & 
        \textbf{77.76} {\scriptsize \textcolor{gray}{$\pm$1.80}} & 
        \textbf{84.35} {\scriptsize \textcolor{gray}{$\pm$1.00}} & 
        \textbf{76.70}\\
    
        \addlinespace[1pt]
        \midrule[0.1pt]
        \addlinespace[2pt]
    
        \multirow{7}{*}{\rotatebox{90}{AttrMasking}} &
        FT & 
        72.71 {\scriptsize \textcolor{gray}{$\pm$0.83}} &
        83.04 {\scriptsize \textcolor{gray}{$\pm$0.20}} &
        66.92 {\scriptsize \textcolor{gray}{$\pm$0.44}} &
        68.57 {\scriptsize \textcolor{gray}{$\pm$0.72}} &
        76.02 {\scriptsize \textcolor{gray}{$\pm$2.71}} &
        80.22 {\scriptsize \textcolor{gray}{$\pm$1.82}} &
        78.13 {\scriptsize \textcolor{gray}{$\pm$0.56}} &
        83.19 {\scriptsize \textcolor{gray}{$\pm$1.88}} & 
        76.10 \\
        & GPF &
        73.08 {\scriptsize \textcolor{gray}{$\pm$0.66}} &
        \underline{83.71} {\scriptsize \textcolor{gray}{$\pm$0.61}} &
        67.30 {\scriptsize \textcolor{gray}{$\pm$0.59}} &
        69.69 {\scriptsize \textcolor{gray}{$\pm$0.79}} &
        76.51 {\scriptsize \textcolor{gray}{$\pm$1.55}} &
        80.48 {\scriptsize \textcolor{gray}{$\pm$1.03}} &
        78.91 {\scriptsize \textcolor{gray}{$\pm$1.20}} &
        85.36 {\scriptsize \textcolor{gray}{$\pm$0.72}} & 
        76.88 \\
        & GPF-plus &
        73.15 {\scriptsize \textcolor{gray}{$\pm$0.75}} &
        83.47 {\scriptsize \textcolor{gray}{$\pm$0.38}} &
        67.52 {\scriptsize \textcolor{gray}{$\pm$0.52}} &
        69.82 {\scriptsize \textcolor{gray}{$\pm$0.75}} &
        \underline{77.20} {\scriptsize \textcolor{gray}{$\pm$2.04}} &
        \underline{80.74} {\scriptsize \textcolor{gray}{$\pm$1.17}} &
        78.77 {\scriptsize \textcolor{gray}{$\pm$1.08}} &
        \underline{86.00} {\scriptsize \textcolor{gray}{$\pm$0.57}} & 
        \underline{77.08} \\
        & SUPT\textsubscript{soft} &
        \underline{73.51} {\scriptsize \textcolor{gray}{$\pm$1.01}} &
        \underline{83.71} {\scriptsize \textcolor{gray}{$\pm$0.72}} &
        67.54 {\scriptsize \textcolor{gray}{$\pm$0.87}} &
        \underline{69.90} {\scriptsize \textcolor{gray}{$\pm$0.92}} &
        \underline{77.20} {\scriptsize \textcolor{gray}{$\pm$2.31}} &
        80.35 {\scriptsize \textcolor{gray}{$\pm$1.39}} &
        \underline{78.95} {\scriptsize \textcolor{gray}{$\pm$1.23}} &
        85.35 {\scriptsize \textcolor{gray}{$\pm$1.18}} & 
        77.06 \\
        & SUPT\textsubscript{hard} &
        73.28 {\scriptsize \textcolor{gray}{$\pm$1.08}} &
        83.52 {\scriptsize \textcolor{gray}{$\pm$0.94}} &
        \underline{67.59} {\scriptsize \textcolor{gray}{$\pm$0.75}} &
        69.57 {\scriptsize \textcolor{gray}{$\pm$0.80}} &
        76.96 {\scriptsize \textcolor{gray}{$\pm$1.78}} &
        80.33 {\scriptsize \textcolor{gray}{$\pm$0.99}} &
        78.60 {\scriptsize \textcolor{gray}{$\pm$0.93}} &
        85.18 {\scriptsize \textcolor{gray}{$\pm$1.33}} & 
        76.88 \\
        & RELIEF &
        73.16 {\scriptsize \textcolor{gray}{$\pm$0.47}} &
        83.04 {\scriptsize \textcolor{gray}{$\pm$0.43}} &
        67.47 {\scriptsize \textcolor{gray}{$\pm$0.22}} &
        69.06 {\scriptsize \textcolor{gray}{$\pm$0.39}} &
        76.48 {\scriptsize \textcolor{gray}{$\pm$1.21}} &
        80.26 {\scriptsize \textcolor{gray}{$\pm$0.56}} &
        78.17 {\scriptsize \textcolor{gray}{$\pm$0.50}} &
        85.20 {\scriptsize \textcolor{gray}{$\pm$0.28}} & 
        76.61 \\
        & LEAP &
        \textbf{73.84} {\scriptsize \textcolor{gray}{$\pm$0.70}} & 
        \textbf{83.94} {\scriptsize \textcolor{gray}{$\pm$0.66}} & 
        \textbf{67.83} {\scriptsize \textcolor{gray}{$\pm$0.71}} & 
        \textbf{70.12} {\scriptsize \textcolor{gray}{$\pm$0.83}} & 
        \textbf{77.27} {\scriptsize \textcolor{gray}{$\pm$1.90}} & 
        \textbf{80.80} {\scriptsize \textcolor{gray}{$\pm$1.05}} & 
        \textbf{79.29} {\scriptsize \textcolor{gray}{$\pm$1.16}} & 
        \textbf{86.16} {\scriptsize \textcolor{gray}{$\pm$1.23}} & 
        \textbf{77.41}\\
    
        \addlinespace[1pt]
        \midrule[0.1pt]
        \addlinespace[2pt]
    
        \multirow{7}{*}{\rotatebox{90}{ContextPred}} &
        FT & 
        75.33 {\scriptsize \textcolor{gray}{$\pm$0.69}} &
        83.11 {\scriptsize \textcolor{gray}{$\pm$0.62}} &
        67.50 {\scriptsize \textcolor{gray}{$\pm$0.65}} &
        66.89 {\scriptsize \textcolor{gray}{$\pm$0.89}} &
        76.19 {\scriptsize \textcolor{gray}{$\pm$1.80}} &
        83.52 {\scriptsize \textcolor{gray}{$\pm$1.35}} &
        79.15 {\scriptsize \textcolor{gray}{$\pm$0.71}} &
        85.10 {\scriptsize \textcolor{gray}{$\pm$0.91}} & 
        77.10 \\
        & GPF &
        74.85 {\scriptsize \textcolor{gray}{$\pm$0.80}} &
        84.20 {\scriptsize \textcolor{gray}{$\pm$0.38}} &
        68.19 {\scriptsize \textcolor{gray}{$\pm$0.48}} &
        67.40 {\scriptsize \textcolor{gray}{$\pm$0.58}} &
        76.52 {\scriptsize \textcolor{gray}{$\pm$1.91}} &
        84.08 {\scriptsize \textcolor{gray}{$\pm$1.00}} &
        79.04 {\scriptsize \textcolor{gray}{$\pm$1.50}} &
        85.82 {\scriptsize \textcolor{gray}{$\pm$0.78}} & 
        77.51 \\
        & GPF-plus &
        75.42 {\scriptsize \textcolor{gray}{$\pm$0.66}} &
        \underline{84.63} {\scriptsize \textcolor{gray}{$\pm$0.60}} &
        \underline{68.31} {\scriptsize \textcolor{gray}{$\pm$0.54}} &
        67.39 {\scriptsize \textcolor{gray}{$\pm$0.70}} &
        76.90 {\scriptsize \textcolor{gray}{$\pm$2.07}} &
        84.08 {\scriptsize \textcolor{gray}{$\pm$1.16}} &
        79.12 {\scriptsize \textcolor{gray}{$\pm$1.22}} &
        86.07 {\scriptsize \textcolor{gray}{$\pm$0.87}} & 
        77.74 \\
        & SUPT\textsubscript{soft} &
        75.08 {\scriptsize \textcolor{gray}{$\pm$1.00}} &
        84.36 {\scriptsize \textcolor{gray}{$\pm$0.82}} &
        68.00 {\scriptsize \textcolor{gray}{$\pm$0.91}} &
        \textbf{68.08} {\scriptsize \textcolor{gray}{$\pm$1.02}} &
        \underline{76.91} {\scriptsize \textcolor{gray}{$\pm$2.20}} &
        \textbf{84.26} {\scriptsize \textcolor{gray}{$\pm$1.19}} &
        \underline{79.31} {\scriptsize \textcolor{gray}{$\pm$1.01}} &
        \underline{86.19} {\scriptsize \textcolor{gray}{$\pm$1.04}} & 
        \underline{77.77} \\
        & SUPT\textsubscript{hard} &
        \underline{75.62} {\scriptsize \textcolor{gray}{$\pm$0.92}} &
        84.30 {\scriptsize \textcolor{gray}{$\pm$0.59}} &
        68.12 {\scriptsize \textcolor{gray}{$\pm$0.70}} &
        \underline{68.00} {\scriptsize \textcolor{gray}{$\pm$0.59}} &
        76.58 {\scriptsize \textcolor{gray}{$\pm$2.02}} &
        84.15 {\scriptsize \textcolor{gray}{$\pm$1.24}} &
        79.15 {\scriptsize \textcolor{gray}{$\pm$1.11}} &
        85.94 {\scriptsize \textcolor{gray}{$\pm$1.43}} & 
        77.73 \\
        & RELIEF &
        75.18 {\scriptsize \textcolor{gray}{$\pm$0.53}} &
        84.17 {\scriptsize \textcolor{gray}{$\pm$0.25}} &
        67.96 {\scriptsize \textcolor{gray}{$\pm$0.30}} &
        67.33 {\scriptsize \textcolor{gray}{$\pm$0.42}} &
        76.32 {\scriptsize \textcolor{gray}{$\pm$0.93}} &
        83.91 {\scriptsize \textcolor{gray}{$\pm$0.66}} &
        78.97 {\scriptsize \textcolor{gray}{$\pm$0.54}} &
        85.77 {\scriptsize \textcolor{gray}{$\pm$0.56}} & 
        77.45 \\
        & LEAP &
        \textbf{76.01} {\scriptsize \textcolor{gray}{$\pm$0.65}} & 
        \textbf{84.89} {\scriptsize \textcolor{gray}{$\pm$0.72}} & 
        \textbf{68.46} {\scriptsize \textcolor{gray}{$\pm$0.72}} & 
        \textbf{68.08} {\scriptsize \textcolor{gray}{$\pm$0.80}} & 
        \textbf{77.04} {\scriptsize \textcolor{gray}{$\pm$1.91}} & 
        \underline{84.21} {\scriptsize \textcolor{gray}{$\pm$1.30}} & 
        \textbf{79.42} {\scriptsize \textcolor{gray}{$\pm$1.06}} & 
        \textbf{86.30} {\scriptsize \textcolor{gray}{$\pm$1.19}} & 
        \textbf{78.05}\\
    
        \addlinespace[1pt]
        \midrule[0.1pt]
        \addlinespace[2pt]
    
        \multirow{7}{*}{\rotatebox{90}{GCL}} &
        FT & 
        76.13 {\scriptsize \textcolor{gray}{$\pm$1.49}} &
        79.05 {\scriptsize \textcolor{gray}{$\pm$0.63}} &
        64.80 {\scriptsize \textcolor{gray}{$\pm$0.57}} &
        63.52 {\scriptsize \textcolor{gray}{$\pm$1.40}} &
        77.32 {\scriptsize \textcolor{gray}{$\pm$2.01}} &
        73.40 {\scriptsize \textcolor{gray}{$\pm$1.41}} &
        \textbf{78.31} {\scriptsize \textcolor{gray}{$\pm$0.90}} &
        75.70 {\scriptsize \textcolor{gray}{$\pm$1.64}} & 
        73.53 \\
        & GPF &
        77.20 {\scriptsize \textcolor{gray}{$\pm$1.08}} &
        79.42 {\scriptsize \textcolor{gray}{$\pm$0.40}} &
        64.71 {\scriptsize \textcolor{gray}{$\pm$0.72}} &
        64.20 {\scriptsize \textcolor{gray}{$\pm$0.71}} &
        79.11 {\scriptsize \textcolor{gray}{$\pm$2.50}} &
        73.85 {\scriptsize \textcolor{gray}{$\pm$1.01}} &
        77.22 {\scriptsize \textcolor{gray}{$\pm$1.44}} &
        78.01 {\scriptsize \textcolor{gray}{$\pm$0.86}} & 
        74.22 \\
        & GPF-plus &
        \underline{77.44} {\scriptsize \textcolor{gray}{$\pm$0.76}} &
        79.33 {\scriptsize \textcolor{gray}{$\pm$0.58}} &
        64.73 {\scriptsize \textcolor{gray}{$\pm$0.65}} &
        64.39 {\scriptsize \textcolor{gray}{$\pm$0.82}} &
        79.36 {\scriptsize \textcolor{gray}{$\pm$1.54}} &
        \underline{74.21} {\scriptsize \textcolor{gray}{$\pm$1.59}} &
        77.70 {\scriptsize \textcolor{gray}{$\pm$1.01}} &
        78.01 {\scriptsize \textcolor{gray}{$\pm$0.90}} & 
        \underline{74.40} \\
        & SUPT\textsubscript{soft} &
        76.09 {\scriptsize \textcolor{gray}{$\pm$1.10}} &
        \textbf{80.05} {\scriptsize \textcolor{gray}{$\pm$0.71}} &
        \underline{64.83} {\scriptsize \textcolor{gray}{$\pm$0.95}} &
        \underline{64.90} {\scriptsize \textcolor{gray}{$\pm$1.04}} &
        78.99 {\scriptsize \textcolor{gray}{$\pm$2.14}} &
        73.69 {\scriptsize \textcolor{gray}{$\pm$1.72}} &
        \underline{78.15} {\scriptsize \textcolor{gray}{$\pm$1.39}} &
        78.00 {\scriptsize \textcolor{gray}{$\pm$1.13}} & 
        74.34 \\
        & SUPT\textsubscript{hard} &
        76.25 {\scriptsize \textcolor{gray}{$\pm$0.98}} &
        79.92 {\scriptsize \textcolor{gray}{$\pm$0.85}} &
        64.69 {\scriptsize \textcolor{gray}{$\pm$0.88}} &
        64.63 {\scriptsize \textcolor{gray}{$\pm$0.69}} &
        78.86 {\scriptsize \textcolor{gray}{$\pm$2.05}} &
        73.00 {\scriptsize \textcolor{gray}{$\pm$1.50}} &
        78.06 {\scriptsize \textcolor{gray}{$\pm$1.62}} &
        \underline{78.19} {\scriptsize \textcolor{gray}{$\pm$1.07}} & 
        74.20 \\
        & RELIEF &
        76.93 {\scriptsize \textcolor{gray}{$\pm$0.70}} &
        78.95 {\scriptsize \textcolor{gray}{$\pm$0.19}} &
        64.73 {\scriptsize \textcolor{gray}{$\pm$0.36}} &
        64.20 {\scriptsize \textcolor{gray}{$\pm$0.52}} &
        \underline{79.70} {\scriptsize \textcolor{gray}{$\pm$1.36}} &
        72.55 {\scriptsize \textcolor{gray}{$\pm$0.79}} &
        77.80 {\scriptsize \textcolor{gray}{$\pm$1.06}} &
        76.03 {\scriptsize \textcolor{gray}{$\pm$0.49}} & 
        73.86 \\
        & LEAP &
        \textbf{77.69} {\scriptsize \textcolor{gray}{$\pm$0.84}} & 
        \underline{80.02} {\scriptsize \textcolor{gray}{$\pm$0.51}} & 
        \textbf{64.92} {\scriptsize \textcolor{gray}{$\pm$0.60}} & 
        \textbf{64.99} {\scriptsize \textcolor{gray}{$\pm$0.74}} & 
        \textbf{80.08} {\scriptsize \textcolor{gray}{$\pm$1.51}} & 
        \textbf{74.28} {\scriptsize \textcolor{gray}{$\pm$1.38}} & 
        78.12 {\scriptsize \textcolor{gray}{$\pm$1.14}} &  
        \textbf{78.34} {\scriptsize \textcolor{gray}{$\pm$1.30}} & 
        \textbf{74.81}\\
        
        \bottomrule
      \end{tabular}
  \label{tab:graph_full}
\end{table*}

\begin{table*}[!t]
    \caption{Accuracy (\%) and Macro F1-score (\%) with respective standard deviation for node classification under full-shot scenario with two pre-training and various tuning strategies.}
    \vskip -0.1in
    \small
        \centering
        \begin{tabular}{cc<{\hspace{3pt}}|>{\hspace{3pt}}c@{\hspace{3pt}}c|c@{\hspace{3pt}}c|c@{\hspace{3pt}}c|c@{\hspace{3pt}}c|c@{\hspace{3pt}}c}

        \toprule
        
        & \multirow{2}{*}{\makecell{Tuning \\ Strategy}} & \multicolumn{2}{c|}{Cora} & \multicolumn{2}{c|}{CiteSeer} & \multicolumn{2}{c|}{PubMed} & \multicolumn{2}{c|}{Computers} & \multicolumn{2}{c}{Photos} \\
        & & Accuracy & Macro F1 & Accuracy & Macro F1 & Accuracy & Macro F1 & Accuracy & Macro F1 & Accuracy & Macro F1 \\
    
        \midrule

        \multirow{8}[2]{*}{\rotatebox{90}{MaskedEdge}} &
        FT & 
        78.87 {\scriptsize \textcolor{gray}{$\pm$1.33}} &
        75.35 {\scriptsize \textcolor{gray}{$\pm$0.85}} &
        70.28 {\scriptsize \textcolor{gray}{$\pm$1.28}} &
        67.61 {\scriptsize \textcolor{gray}{$\pm$0.91}} &
        78.99 {\scriptsize \textcolor{gray}{$\pm$0.92}} &
        77.43 {\scriptsize \textcolor{gray}{$\pm$0.75}} &
        86.41 {\scriptsize \textcolor{gray}{$\pm$1.23}} &
        83.95 {\scriptsize \textcolor{gray}{$\pm$1.16}} &
        90.39 {\scriptsize \textcolor{gray}{$\pm$0.98}} &
        88.93 {\scriptsize \textcolor{gray}{$\pm$0.73}} \\
        
        & GPPT &
        81.16 {\scriptsize \textcolor{gray}{$\pm$0.74}} &
        78.90 {\scriptsize \textcolor{gray}{$\pm$1.30}} &
        70.39 {\scriptsize \textcolor{gray}{$\pm$1.12}} &
        68.02 {\scriptsize \textcolor{gray}{$\pm$0.96}} &
        77.88 {\scriptsize \textcolor{gray}{$\pm$1.32}} &
        76.25 {\scriptsize \textcolor{gray}{$\pm$1.34}} &
        84.29 {\scriptsize \textcolor{gray}{$\pm$1.20}} &
        81.60 {\scriptsize \textcolor{gray}{$\pm$0.98}} &
        88.95 {\scriptsize \textcolor{gray}{$\pm$1.21}} &
        85.92 {\scriptsize \textcolor{gray}{$\pm$0.98}} \\
        
        & GPF &
        81.60 {\scriptsize \textcolor{gray}{$\pm$1.22}} &
        79.22 {\scriptsize \textcolor{gray}{$\pm$1.10}} &
        70.40 {\scriptsize \textcolor{gray}{$\pm$1.24}} &
        68.38 {\scriptsize \textcolor{gray}{$\pm$1.28}} &
        79.35 {\scriptsize \textcolor{gray}{$\pm$1.04}} &
        78.04 {\scriptsize \textcolor{gray}{$\pm$1.14}} &
        86.32 {\scriptsize \textcolor{gray}{$\pm$1.30}} &
        83.91 {\scriptsize \textcolor{gray}{$\pm$1.28}} &
        90.45 {\scriptsize \textcolor{gray}{$\pm$1.42}} &
        89.11 {\scriptsize \textcolor{gray}{$\pm$1.40}} \\
        
        & GPF-plus &
        \underline{82.02} {\scriptsize \textcolor{gray}{$\pm$1.02}} &
        \underline{79.91} {\scriptsize \textcolor{gray}{$\pm$0.88}} &
        \underline{70.98} {\scriptsize \textcolor{gray}{$\pm$1.51}} &
        \underline{68.70} {\scriptsize \textcolor{gray}{$\pm$1.67}} &
        79.52 {\scriptsize \textcolor{gray}{$\pm$1.38}} &
        78.41 {\scriptsize \textcolor{gray}{$\pm$1.02}} &
        86.82 {\scriptsize \textcolor{gray}{$\pm$1.03}} &
        84.41 {\scriptsize \textcolor{gray}{$\pm$1.27}} &
        \underline{90.70} {\scriptsize \textcolor{gray}{$\pm$1.62}} &
        \underline{89.35} {\scriptsize \textcolor{gray}{$\pm$1.65}} \\
        
        & SUPT\textsubscript{soft} &
        81.80 {\scriptsize \textcolor{gray}{$\pm$1.18}} &
        79.37 {\scriptsize \textcolor{gray}{$\pm$1.35}} &
        70.50 {\scriptsize \textcolor{gray}{$\pm$1.43}} &
        68.33 {\scriptsize \textcolor{gray}{$\pm$1.30}} &
        \underline{79.98} {\scriptsize \textcolor{gray}{$\pm$1.29}} &
        \underline{78.77} {\scriptsize \textcolor{gray}{$\pm$1.38}} &
        \textbf{87.03} {\scriptsize \textcolor{gray}{$\pm$0.97}} &
        \textbf{84.73} {\scriptsize \textcolor{gray}{$\pm$0.90}} &
        90.04 {\scriptsize \textcolor{gray}{$\pm$1.01}} &
        89.85 {\scriptsize \textcolor{gray}{$\pm$1.18}} \\
        
        & SUPT\textsubscript{hard} &
        80.09 {\scriptsize \textcolor{gray}{$\pm$1.20}} &
        77.70 {\scriptsize \textcolor{gray}{$\pm$1.33}} &
        70.22 {\scriptsize \textcolor{gray}{$\pm$1.13}} &
        68.10 {\scriptsize \textcolor{gray}{$\pm$1.02}} &
        78.49 {\scriptsize \textcolor{gray}{$\pm$0.86}} &
        77.20 {\scriptsize \textcolor{gray}{$\pm$1.01}} &
        86.59 {\scriptsize \textcolor{gray}{$\pm$0.89}} &
        84.20 {\scriptsize \textcolor{gray}{$\pm$0.92}} &
        90.11 {\scriptsize \textcolor{gray}{$\pm$0.97}} &
        89.97 {\scriptsize \textcolor{gray}{$\pm$1.03}} \\
        
        & RELIEF &
        79.00 {\scriptsize \textcolor{gray}{$\pm$0.76}} &
        76.12 {\scriptsize \textcolor{gray}{$\pm$0.67}} &
        70.30 {\scriptsize \textcolor{gray}{$\pm$0.92}} &
        68.13 {\scriptsize \textcolor{gray}{$\pm$0.84}} &
        79.06 {\scriptsize \textcolor{gray}{$\pm$0.88}} &
        77.97 {\scriptsize \textcolor{gray}{$\pm$0.91}} &
        85.92 {\scriptsize \textcolor{gray}{$\pm$0.69}} &
        83.74 {\scriptsize \textcolor{gray}{$\pm$0.72}} &
        90.00 {\scriptsize \textcolor{gray}{$\pm$0.67}} &
        89.91 {\scriptsize \textcolor{gray}{$\pm$0.63}} \\
        
        & \model  & 
        \textbf{82.81} {\scriptsize \textcolor{gray}{$\pm$1.00}} & 
        \textbf{80.77} {\scriptsize \textcolor{gray}{$\pm$0.91}} & 
        \textbf{71.54} {\scriptsize \textcolor{gray}{$\pm$1.38}} & 
        \textbf{69.09} {\scriptsize \textcolor{gray}{$\pm$1.27}} & 
        \textbf{80.48} {\scriptsize \textcolor{gray}{$\pm$1.20}} & 
        \textbf{79.51} {\scriptsize \textcolor{gray}{$\pm$1.14}} & 
        \underline{86.89} {\scriptsize \textcolor{gray}{$\pm$1.61}} & 
        \underline{84.56} {\scriptsize \textcolor{gray}{$\pm$1.72}} & 
        \textbf{91.04} {\scriptsize \textcolor{gray}{$\pm$1.09}} & 
        \textbf{90.63} {\scriptsize \textcolor{gray}{$\pm$1.13}} \\
        
        \addlinespace[1pt]
        \midrule[0.1pt]
        \addlinespace[2pt]
    
        \multirow{8}[2]{*}{\rotatebox{90}{ContraEdge}} &
        FT & 
        80.45 {\scriptsize \textcolor{gray}{$\pm$1.56}} &
        80.02 {\scriptsize \textcolor{gray}{$\pm$1.29}} &
        72.56 {\scriptsize \textcolor{gray}{$\pm$1.37}} &
        70.04 {\scriptsize \textcolor{gray}{$\pm$1.12}} &
        79.21 {\scriptsize \textcolor{gray}{$\pm$0.97}} &
        77.71 {\scriptsize \textcolor{gray}{$\pm$0.89}} &
        88.71 {\scriptsize \textcolor{gray}{$\pm$1.00}} &
        86.19 {\scriptsize \textcolor{gray}{$\pm$1.07}} &
        91.02 {\scriptsize \textcolor{gray}{$\pm$1.01}} &
        89.55 {\scriptsize \textcolor{gray}{$\pm$0.94}} \\
        
        & GPrompt &
        79.58 {\scriptsize \textcolor{gray}{$\pm$1.70}} &
        78.14 {\scriptsize \textcolor{gray}{$\pm$1.55}} &
        72.40 {\scriptsize \textcolor{gray}{$\pm$1.11}} &
        69.88 {\scriptsize \textcolor{gray}{$\pm$1.29}} &
        78.30 {\scriptsize \textcolor{gray}{$\pm$1.80}} &
        77.69 {\scriptsize \textcolor{gray}{$\pm$1.67}} &
        87.43 {\scriptsize \textcolor{gray}{$\pm$1.42}} &
        84.89 {\scriptsize \textcolor{gray}{$\pm$1.63}} &
        90.55 {\scriptsize \textcolor{gray}{$\pm$1.43}} &
        89.06 {\scriptsize \textcolor{gray}{$\pm$1.18}} \\
        
        & GPF &
        80.00 {\scriptsize \textcolor{gray}{$\pm$1.02}} &
        79.49 {\scriptsize \textcolor{gray}{$\pm$1.19}} &
        73.01 {\scriptsize \textcolor{gray}{$\pm$1.70}} &
        70.59 {\scriptsize \textcolor{gray}{$\pm$1.64}} &
        79.03 {\scriptsize \textcolor{gray}{$\pm$1.59}} &
        77.60 {\scriptsize \textcolor{gray}{$\pm$1.46}} &
        88.94 {\scriptsize \textcolor{gray}{$\pm$1.30}} &
        86.48 {\scriptsize \textcolor{gray}{$\pm$1.21}} &
        91.35 {\scriptsize \textcolor{gray}{$\pm$1.55}} &
        89.87 {\scriptsize \textcolor{gray}{$\pm$1.75}} \\
        
        & GPF-plus &
        \underline{80.98} {\scriptsize \textcolor{gray}{$\pm$1.41}} &
        \underline{80.37} {\scriptsize \textcolor{gray}{$\pm$1.22}} &
        \underline{73.40} {\scriptsize \textcolor{gray}{$\pm$1.40}} &
        \underline{71.01} {\scriptsize \textcolor{gray}{$\pm$1.22}} &
        79.28 {\scriptsize \textcolor{gray}{$\pm$1.99}} &
        77.99 {\scriptsize \textcolor{gray}{$\pm$1.87}} &
        89.13 {\scriptsize \textcolor{gray}{$\pm$1.50}} &
        86.66 {\scriptsize \textcolor{gray}{$\pm$1.52}} &
        \underline{91.49} {\scriptsize \textcolor{gray}{$\pm$1.30}} &
        \underline{90.05} {\scriptsize \textcolor{gray}{$\pm$1.61}} \\
        
        & SUPT\textsubscript{soft} &
        80.50 {\scriptsize \textcolor{gray}{$\pm$1.12}} &
        79.99 {\scriptsize \textcolor{gray}{$\pm$0.98}} &
        72.69 {\scriptsize \textcolor{gray}{$\pm$1.48}} &
        70.21 {\scriptsize \textcolor{gray}{$\pm$1.01}} &
        \underline{79.81} {\scriptsize \textcolor{gray}{$\pm$1.68}} &
        \underline{78.40} {\scriptsize \textcolor{gray}{$\pm$1.82}} &
        \textbf{89.25} {\scriptsize \textcolor{gray}{$\pm$0.95}} &
        \textbf{86.83} {\scriptsize \textcolor{gray}{$\pm$0.79}} &
        91.13 {\scriptsize \textcolor{gray}{$\pm$1.54}} &
        89.68 {\scriptsize \textcolor{gray}{$\pm$1.09}} \\
        
        & SUPT\textsubscript{hard} &
        80.44 {\scriptsize \textcolor{gray}{$\pm$1.01}} &
        79.89 {\scriptsize \textcolor{gray}{$\pm$1.31}} &
        72.53 {\scriptsize \textcolor{gray}{$\pm$1.25}} &
        70.26 {\scriptsize \textcolor{gray}{$\pm$1.18}} &
        79.40 {\scriptsize \textcolor{gray}{$\pm$2.27}} &
        78.08 {\scriptsize \textcolor{gray}{$\pm$2.30}} &
        89.02 {\scriptsize \textcolor{gray}{$\pm$0.84}} &
        86.59 {\scriptsize \textcolor{gray}{$\pm$0.79}} &
        91.05 {\scriptsize \textcolor{gray}{$\pm$1.05}} &
        89.57 {\scriptsize \textcolor{gray}{$\pm$0.89}} \\
      
        & RELIEF &
        80.33 {\scriptsize \textcolor{gray}{$\pm$0.72}} &
        80.00 {\scriptsize \textcolor{gray}{$\pm$0.89}} &
        72.43 {\scriptsize \textcolor{gray}{$\pm$0.94}} &
        69.96 {\scriptsize \textcolor{gray}{$\pm$0.82}} &
        79.21 {\scriptsize \textcolor{gray}{$\pm$0.66}} &
        77.59 {\scriptsize \textcolor{gray}{$\pm$0.59}} &
        88.98 {\scriptsize \textcolor{gray}{$\pm$1.01}} &
        86.59 {\scriptsize \textcolor{gray}{$\pm$1.05}} &
        90.88 {\scriptsize \textcolor{gray}{$\pm$0.65}} &
        89.42 {\scriptsize \textcolor{gray}{$\pm$0.59}} \\

        & \model & 
        \textbf{81.41} {\scriptsize \textcolor{gray}{$\pm$1.30}} & 
        \textbf{80.70} {\scriptsize \textcolor{gray}{$\pm$1.49}} & 
        \textbf{73.81} {\scriptsize \textcolor{gray}{$\pm$1.35}} & 
        \textbf{71.44} {\scriptsize \textcolor{gray}{$\pm$1.29}} & 
        \textbf{80.22} {\scriptsize \textcolor{gray}{$\pm$1.42}} & 
        \textbf{78.87} {\scriptsize \textcolor{gray}{$\pm$1.31}} & 
        \underline{89.20} {\scriptsize \textcolor{gray}{$\pm$1.25}} & 
        \underline{86.74} {\scriptsize \textcolor{gray}{$\pm$1.26}} & 
        \textbf{91.67} {\scriptsize \textcolor{gray}{$\pm$1.19}} &
        \textbf{90.26} {\scriptsize \textcolor{gray}{$\pm$1.12}} \\

        \bottomrule
        \end{tabular}
    \label{tab:node_full}
\end{table*}

\section{Experiment}
\label{sec:experiments}
\subsection{Evaluation Tasks}
To comprehensively evaluate the performance and generalization of LEAP\footnote{The code and appendix are available at: \href{https://github.com/Jinfeng-Xu/LEAP}{https://github.com/Jinfeng-Xu/LEAP}.}, we provide extensive experiments on both graph- and node-level tasks across various pre-training strategies in both full-shot and few-shot scenarios. Detailed descriptions of tasks are in Appendix~\ref{subsec:task}.

\subsubsection{Model Architecture and Datasets}
For the graph-level task, we use a 5-layer GIN \citep{xu2019powerful} as the base GNN model, pre-trained on chemistry datasets \citep{hu2020strategies} and prompt-tuned on molecular property prediction benchmarks from MoleculeNet \citep{wu2018moleculenet}, following setups in prior feature prompting works \citep{fang2023universal,lee2024subgraph}. For the node-level task, we use a 2-layer GIN, widely adopted in previous works \citep{kipf2017semi,shchur2018pitfalls}. Following prior works \citep{sun2022gppt,liu2023graphprompt}, we pre-train the model using methods designed for edge-level tasks, enabling transfer to node classification. We then evaluate on Cora, Citeseer, Pubmed \citep{kipf2017semi}, and Amazon-Co-Buy (Computer and Photo) \citep{shchur2018pitfalls} for downstream node classification. Detailed descriptions of architectures and datasets are in Appendix~\ref{appendix:architectures}\footnote{Our RL approach differs from RELIEF, which relies on RL to generate universal graph prompts. Instead, our RL focuses on editing universal graph prompts, resulting in low sensitivity to hyper-parameters.} and Appendix~\ref{appendix:datasets}.

\subsubsection{Pre-training Strategies}
For the graph-level task, we employ four widely used strategies to pre-train the graph models, including Deep Graph Infomax (Infomax) \citep{velivckovic2019deep}, Attribute Masking (AttrMasking) \citep{hu2020strategies}, Context Prediction (ContextPred) \citep{hu2020strategies}, and Graph Contrastive Learning (GCL) \citep{you2020graph}. For the node-level task, we employ two widely used strategies to pre-train the graph models, including Masked Edge Prediction (MaskedEdge) \citep{sun2022gppt} and Edge Connectivity Contrastive Learning (ContraEdge) \citep{liu2023graphprompt}. Detailed descriptions of pre-training strategies are in Appendix~\ref{appendix:strategies}.

\subsubsection{Baselines}
For the graph-level task, we compare LEAP against Fine-tuning (FT), GPF, GPG-plus \citep{fang2023universal}, SUPT\textsubscript{soft}, SUPT\textsubscript{hard} \citep{lee2024subgraph}, and RELIEF \citep{zhu2024relief}. For the node-level task, we include all the above baselines and additionally compare with GPPT \citep{sun2022gppt} for the MaskedEdge pre-training strategy and GPrompt \citep{liu2023graphprompt} for the ContraEdge pre-training strategy. Detailed descriptions of baselines are in Appendix~\ref{appendix:baselines}.

\subsubsection{Implementations}
We use early stopping based on validation set performance. Each experiment is repeated five times with different random seeds. For the graph-level task, ROC-AUC is used as the evaluation metric, while for the node-level task, we use Accuracy and Macro F1-score, following previous work \citep{zhu2024relief}. For LEAP, the $k$ for the universal graph prompt is chosen from $[5, 10, 20]$, the prompt edit range $\vartheta$ from $[0.1, 0.5, 1]$, $\lambda_c$ for the reward function from $[1e^{-5}, 1e^{-4}, 1e^{-3}]$, and $h$ for the policy network update interval from $[1, 2, 3, 4]$. Detailed hyper-parameter settings are provided in Appendix~\ref{appendix:hyper-parameter}. All the experiments were conducted on an NVIDIA L20 PCIe 48GB GPU.

\begin{table*}[!ht]
  \caption{ROC-AUC (\%) and standard deviation for graph classification on molecule property prediction benchmark under both full-shot and few-shot scenarios with various pre-training and variants.}
      \centering
      \small
      \begin{tabular}{cccccccccc}
        \toprule
        \multirow{2}{*}{Datasets} & \multirow{2}{*}{Variants} & \multicolumn{2}{c}{Infomax} & \multicolumn{2}{c}{AttrMasking} & \multicolumn{2}{c}{ContextPred}& \multicolumn{2}{c}{GCL} \\
         & & full-shot& few-shot& full-shot& few-shot& full-shot& few-shot& full-shot& few-shot\\

        \midrule





        \multirow{4}{*}{ToxCast} & LEAP & \textbf{67.99}& \textbf{58.79}& \underline{67.83}& \textbf{58.43}& \textbf{68.46}& \textbf{58.53}& \textbf{64.92}& \textbf{54.89}\\
        & LEAP\textsubscript{GPF} & 67.70& 58.40& 67.39& 58.19& \underline{68.31}& 58.30& 64.73& 54.50\\
        & LEAP\textsubscript{GPF-plus} & \underline{67.94}& \underline{58.52}& \textbf{67.84}& \underline{58.32}& \textbf{68.46}& \underline{58.42}& \underline{64.86}& \underline{54.67}\\
        & LEAP $w/o$ ECR & 67.32& 58.47& 67.42& 58.28& 68.10& 58.34& 64.70& 54.48\\

        \addlinespace[1pt]
        \midrule[0.1pt]
        \addlinespace[2pt]



        \multirow{4}{*}{ClinTox} & LEAP & \textbf{75.98}& \textbf{66.44}& \textbf{77.27}& \textbf{73.33}& \textbf{77.04}& \textbf{62.29}& \textbf{80.08}& \textbf{79.93}\\
        & LEAP\textsubscript{GPF} & 75.39& 65.82& 77.15& 71.56& 76.81& 61.98& 79.23& 79.18\\
        & LEAP\textsubscript{GPF-plus} & \underline{75.87}& \underline{65.99}& \underline{77.24}& \underline{72.91}& \underline{76.99}& \underline{62.22}& \underline{80.01}& \underline{79.58}\\
        & LEAP $w/o$ ECR & 75.36& 65.80& 77.00& 72.46& 76.44& 62.05& 79.23& 79.32\\

        \addlinespace[1pt]
        \midrule[0.1pt]
        \addlinespace[2pt]

        \multirow{4}{*}{MUV} & LEAP & \underline{81.49}& \textbf{67.28}& \textbf{80.80}& \textbf{63.42}& \textbf{84.21}& \textbf{64.92}& \textbf{74.28}& \textbf{53.59}\\
        & LEAP\textsubscript{GPF} & 81.28& 67.20& 80.65& 62.94& 84.02& 64.49& 74.01& 52.80\\
        & LEAP\textsubscript{GPF-plus} & \textbf{81.62}& \underline{67.22}& \underline{80.75}& \underline{63.19}& \underline{84.15}& \underline{64.62}& \underline{74.26}& \underline{53.25}\\
        & LEAP $w/o$ ECR & 80.83& 67.15& 80.33& 62.90& 83.98& 64.40& 73.79& 53.21\\

        \addlinespace[1pt]
        \midrule[0.1pt]
        \addlinespace[2pt]



        \multirow{4}{*}{BACE} & LEAP & \underline{84.35}& \textbf{67.31}& \textbf{86.16}& \textbf{69.45}& \textbf{86.30}& \textbf{70.33}& \textbf{78.34}& \textbf{62.50}\\
        & LEAP\textsubscript{GPF} & 84.16& 66.09& 85.85& 68.58& 85.77& 69.20& 78.04& 60.71\\
        & LEAP\textsubscript{GPF-plus} & \textbf{84.37}& 67.01& \underline{86.06}& \underline{69.22}& \underline{86.03}& 69.78& \underline{78.30}& 61.50\\
        & LEAP $w/o$ ECR & 84.08& \underline{67.08}& 85.47& 68.98& 85.42& \underline{70.00}& 77.35& \underline{62.12}\\

      \bottomrule
      \end{tabular}
  \label{tab:ablation_graph}
\end{table*}

\begin{table*}[!ht]
    \caption{Accuracy (\%) and Macro F1-score (\%) with respective standard deviation for node classification under both full-shot and few-shot scenarios with two pre-training and various variants.}
   \small
        \centering
       \begin{tabular}{cccccccccc}

        \toprule
        
        \multirow{3}{*}{Datasets} & \multirow{3}{*}{Variants} & \multicolumn{4}{c}{MaskedEdge} & \multicolumn{4}{c}{ContraEdge}\\
        & & \multicolumn{2}{c}{full-shot}& \multicolumn{2}{c}{few-shot}& \multicolumn{2}{c}{full-shot}& \multicolumn{2}{c}{few-shot}\\
        & & Accuracy & Macro F1 & Accuracy & Macro F1& Accuracy & Macro F1& Accuracy & Macro F1\\
    
        \midrule

        \multirow{4}{*}{Cora} & LEAP & \textbf{82.81}& \textbf{80.77}& \textbf{58.32}& \textbf{56.81}& \textbf{81.41}& \textbf{80.70}& \textbf{61.33}& \textbf{61.12}\\
        & LEAP\textsubscript{GPF} & 81.90& 79.46& 56.32& 55.20& 80.75& 79.81& 59.57& 59.02\\
        & LEAP\textsubscript{GPF-plus} & \underline{82.68}& \underline{80.50}& \underline{57.39}& \underline{55.72}& \underline{81.11}& \underline{80.37}& \underline{61.01}& \underline{60.54}\\
        & LEAP $w/o$ ECR & 80.97& 78.78& 56.65& 55.53& 80.38& 80.04& 60.88& 60.37\\

        \addlinespace[1pt]
        \midrule[0.1pt]
        \addlinespace[2pt]

        \multirow{4}{*}{CiteSeer} & LEAP & \textbf{71.54}& \textbf{69.09}& \textbf{64.24}& \textbf{57.67}& \textbf{73.81}& \textbf{71.44}& \textbf{65.92}& \textbf{57.90}\\
        & LEAP\textsubscript{GPF} & 70.80& 68.56& 62.02& 56.66& 73.12& 70.65& 64.00& 56.49\\
        & LEAP\textsubscript{GPF-plus} & \underline{71.39}& \underline{68.99}& \underline{64.10}& \underline{57.59}& \underline{73.72}& \underline{71.32}& \underline{65.45}& \underline{57.40}\\
        & LEAP $w/o$ ECR & 70.45& 68.09& 63.02& 57.26& 72.66& 70.30& \underline{65.45}& 57.28\\
        
        \addlinespace[1pt]
        \midrule[0.1pt]
        \addlinespace[2pt]





        \multirow{4}{*}{Photos} & LEAP & \textbf{91.04}& \textbf{89.63}& \textbf{85.19}& \textbf{83.35}& \textbf{91.67}& \textbf{90.26}& \textbf{86.41}& \textbf{83.95}\\
        & LEAP\textsubscript{GPF} & 90.30& 89.06& 83.48& 81.20& 91.35& 89.89& 85.66& 82.84\\
        & LEAP\textsubscript{GPF-plus} & \underline{90.90}& \underline{89.51}& \underline{84.67}& \underline{82.75}& \underline{91.60}& \underline{90.18}& \underline{86.18}& \underline{83.53}\\
        & LEAP $w/o$ ECR & 90.19& 88.99& 84.56& 82.71& 91.00& 89.63& 86.10& 83.46\\
        
        \bottomrule
        \end{tabular}
    \label{tab:ablation_node}
\end{table*}

\subsection{Overall Performance}
Based on the results in Table~\ref{tab:graph_full} and Table~\ref{tab:node_full}, our LEAP achieves superior performance compared to baselines on graph- and node-level tasks in the full-shot scenario, with improvements in $28/32$ and $16/20$ tasks (metrics), respectively. Due to space constraints, the results for graph- and node-level tasks under few-shot settings are provided in Table~\ref{tab:graph_few} and Table~\ref{tab:node_few} in Appendix~\ref{appendix:few-shot}. Notably, RELIEF achieves satisfactory performance in the few-shot scenario by leveraging reinforcement learning to dynamically select specific nodes for prompt addition. However, as emphasized in Theorem~\ref{theorem:1}, its selective node-based graph prompt tuning compromises the theoretical foundation of universal graph prompt tuning, leading to unsatisfactory performance in the full-shot scenario.

\subsection{Ablation Study}
\label{appendix:ablation_study}
To analyze the effectiveness of LEAP, we conduct extensive ablation studies to evaluate the necessity and contribution of each individual component within the model. Specifically, we compare our model with the following variants: 1) LEAP\textsubscript{GPF}, which replaces the basic universal graph prompt with GPF. 2) LEAP\textsubscript{GPF-plus}, which replaces the basic universal graph prompt with full GPF-plus. 3) LEAP $w/o$ ECR, which removes the editing coverage rate from the reward function. 
The evaluation results are demonstrated in Table~\ref{tab:ablation_graph} and Table~\ref{tab:ablation_node}. Specifically, we have the following observations. LEAP performs superior in most scenarios. We attribute this to learning different prompts for each node being inherently more challenging than learning $k$ basic universal graph prompt vectors and $k$ linear projections. Thus, LEAP $w/o$ ECR demonstrates the least satisfactory performance in the full-shot scenario. We ascribe this to the absence of the editing convergence rate constraint, which leads the model to repeatedly select and modify a small subset of nodes. 

For more comprehensive experimental results covering all datasets, please refer to Appendix~\ref{appendix:ablation_study}.

\subsection{Training Curve}
\begin{figure*}[!h]
    \centering
    \subfigure {
        \label{fig:training curve1}
        \includegraphics[width=0.2\linewidth]{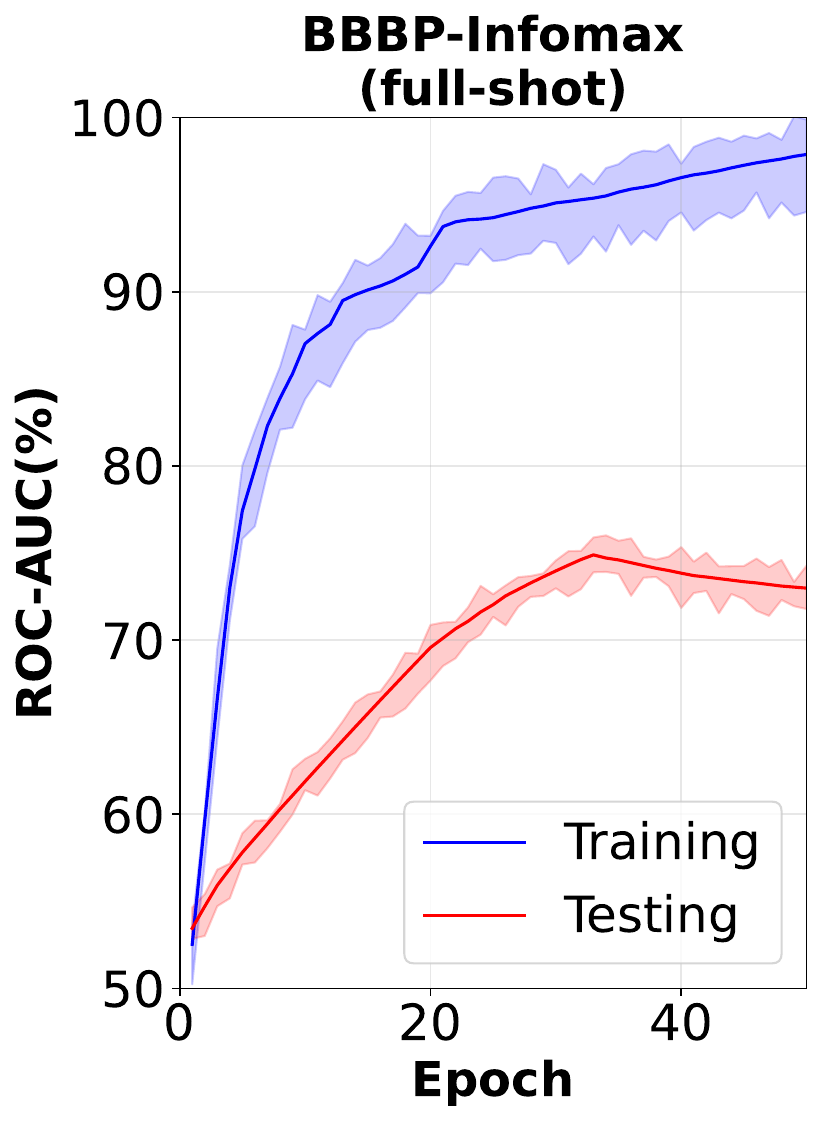}
        } 
    \subfigure {
        \label{fig:training curve2}
        \includegraphics[width=0.2\linewidth]{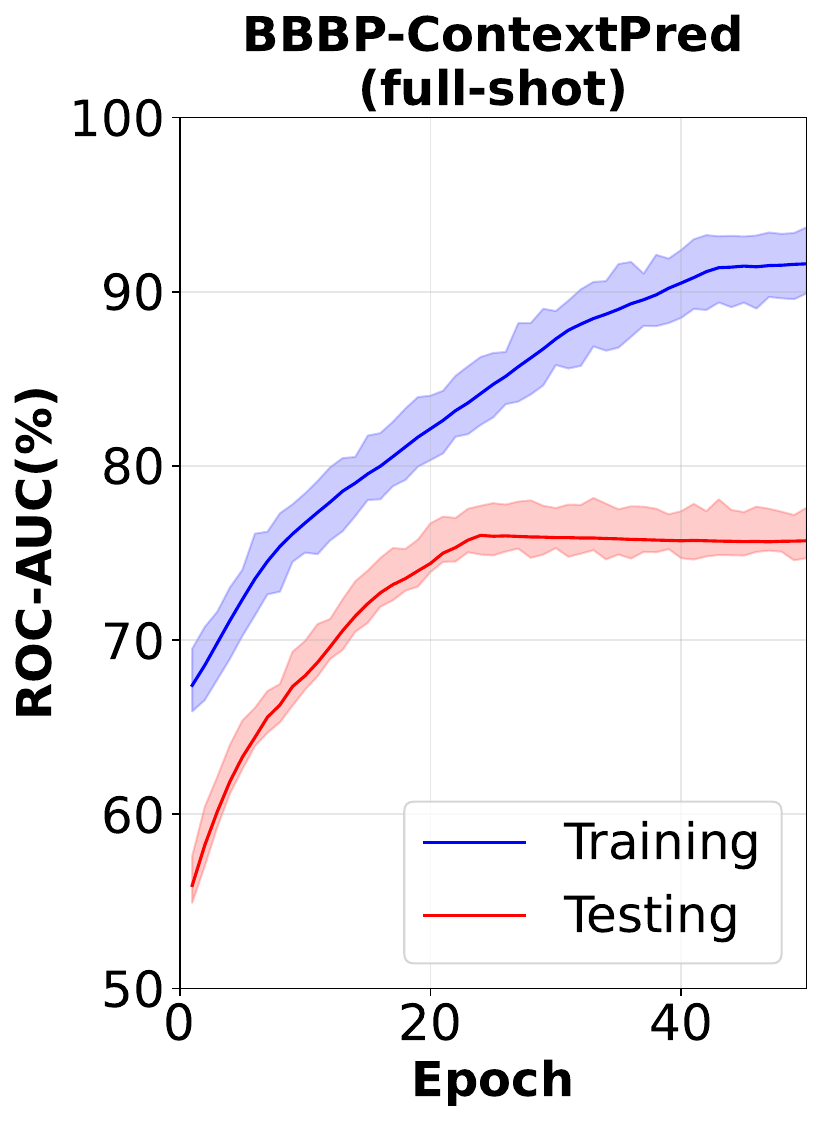}
        }
    \subfigure {
        \label{fig:training curve3}
        \includegraphics[width=0.2\linewidth]{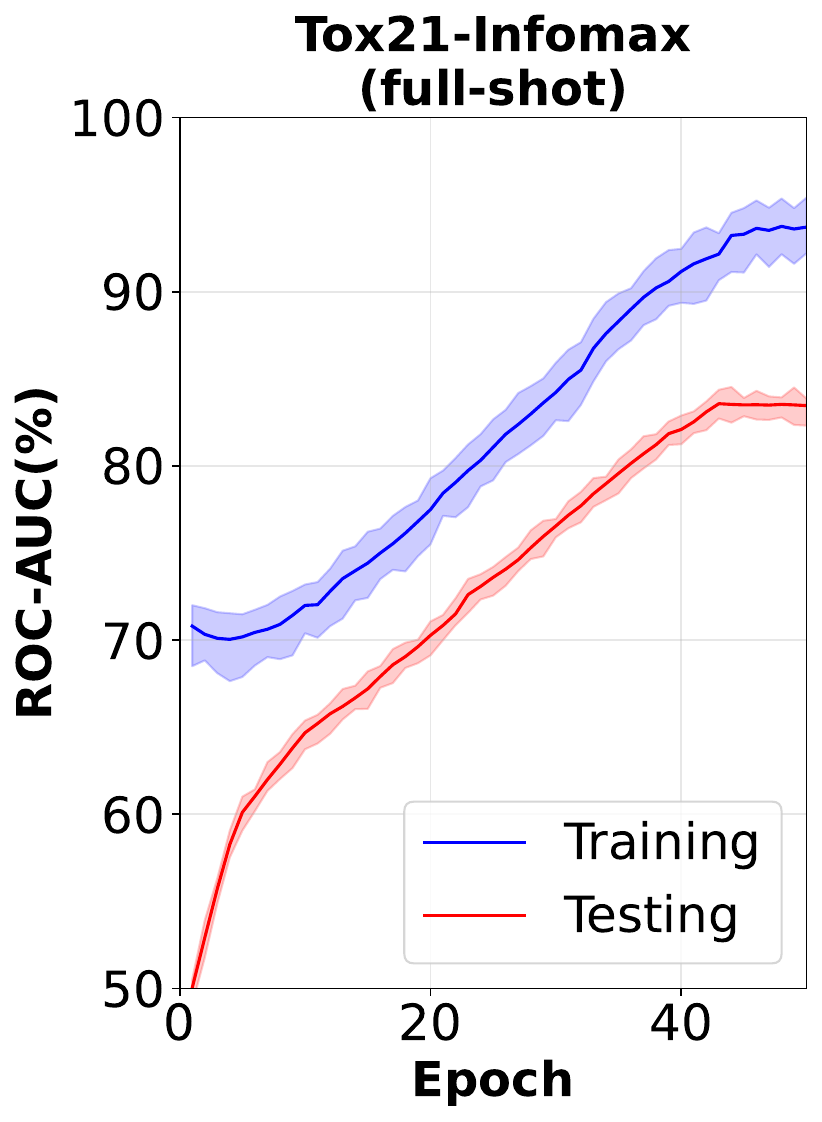}
        }
    \subfigure {
        \label{fig:training curve4}
        \includegraphics[width=0.2\linewidth]{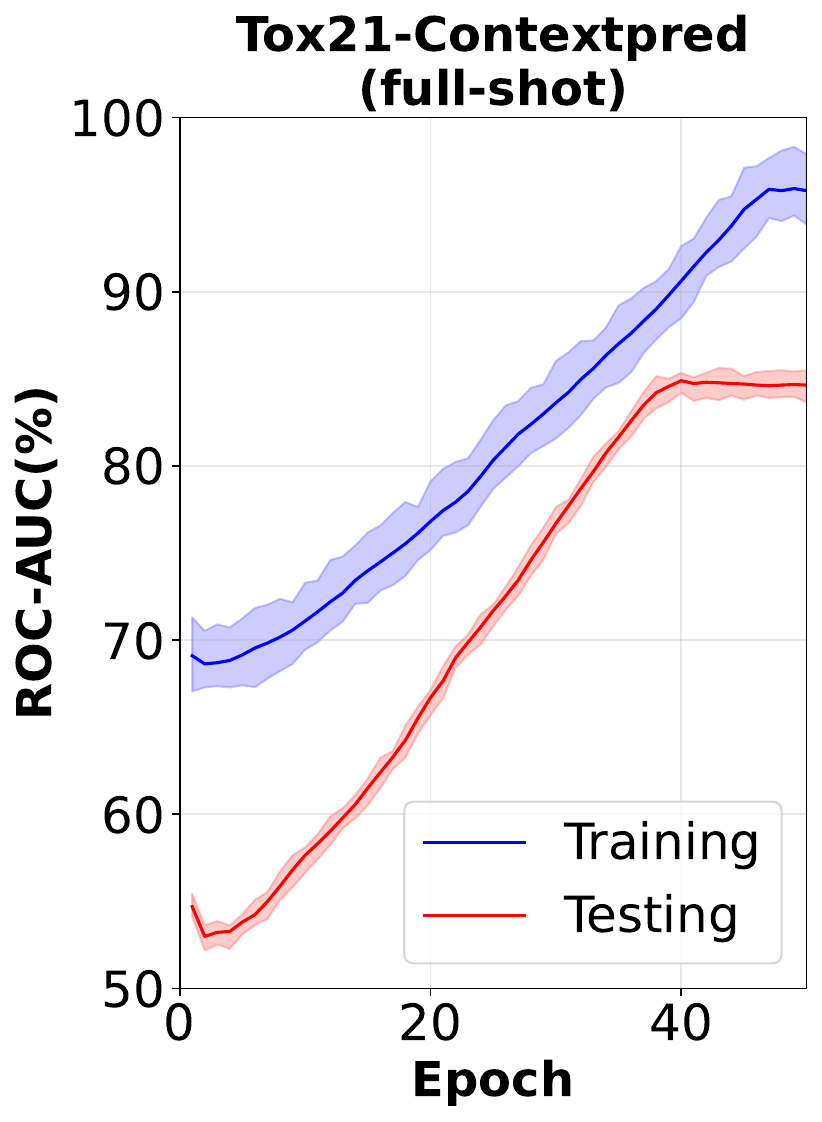}
        }\vskip -0.1in
    \subfigure {
        \label{fig:training curve5}
        \includegraphics[width=0.2\linewidth]{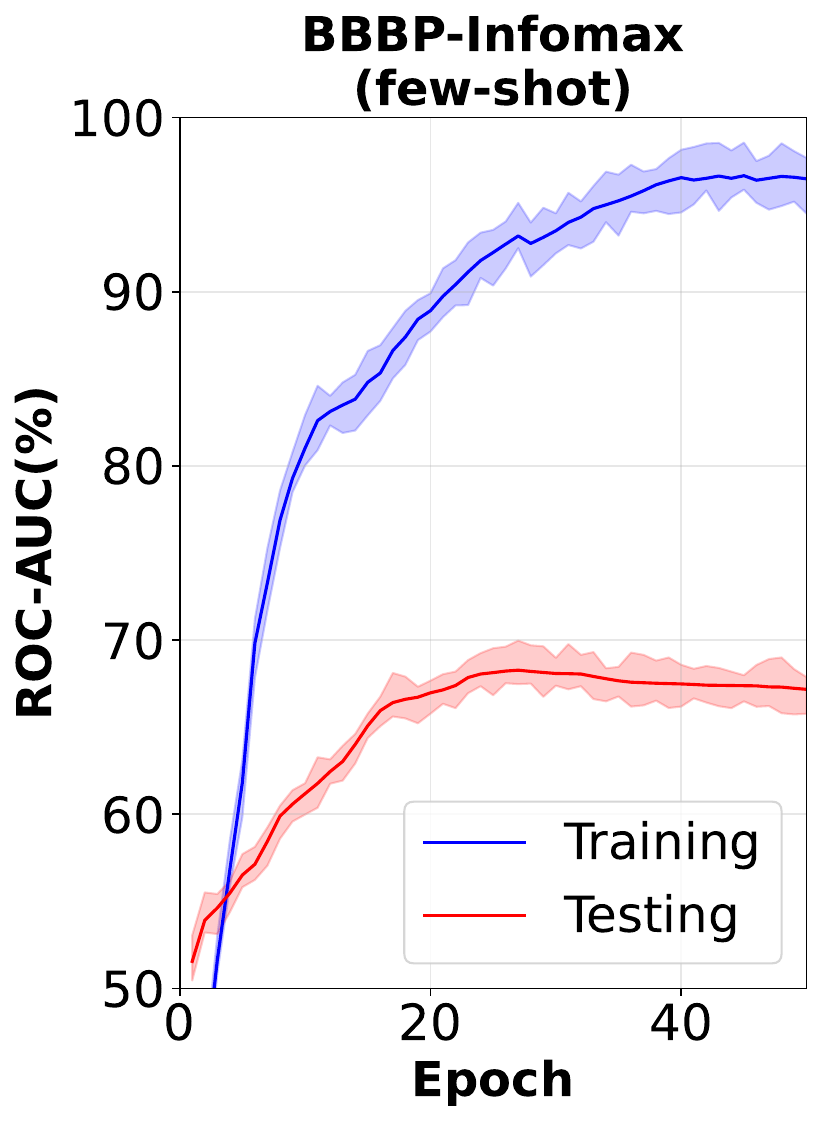}
        } 
    \subfigure {
        \label{fig:training curve6}
        \includegraphics[width=0.2\linewidth]{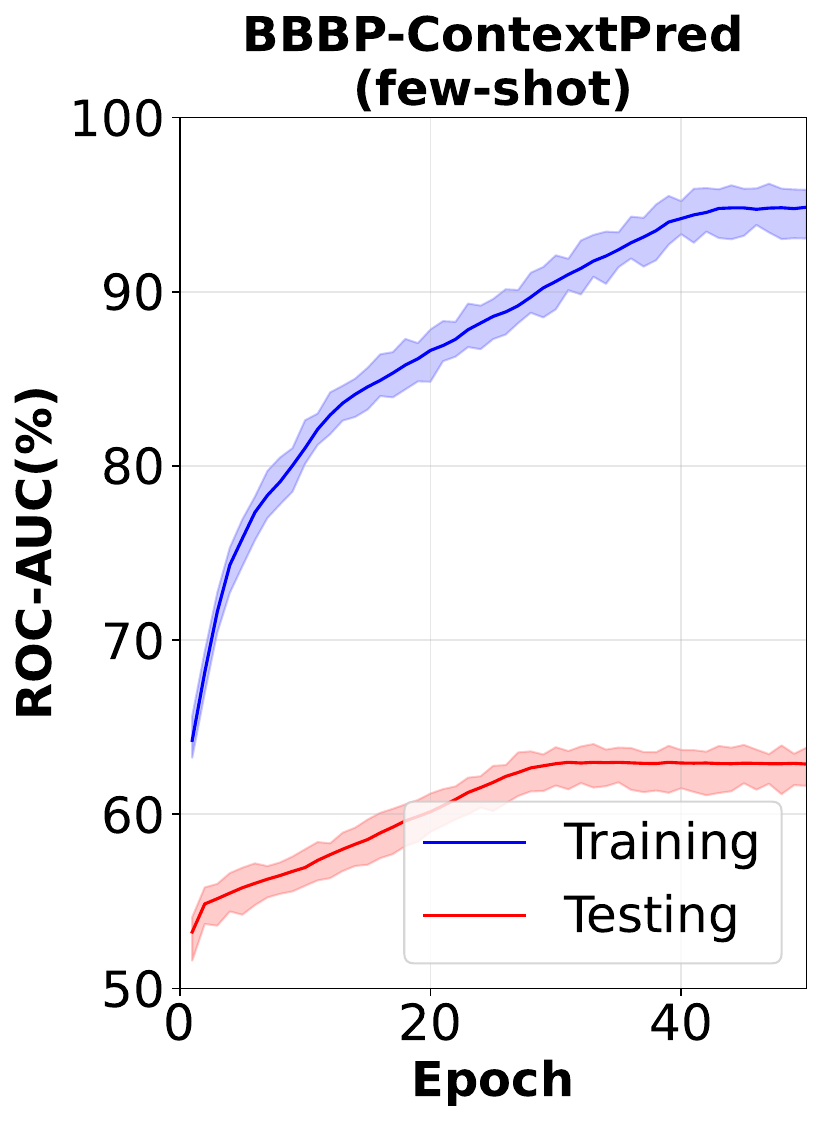}
        }
    \subfigure {
        \label{fig:training curve7}
        \includegraphics[width=0.2\linewidth]{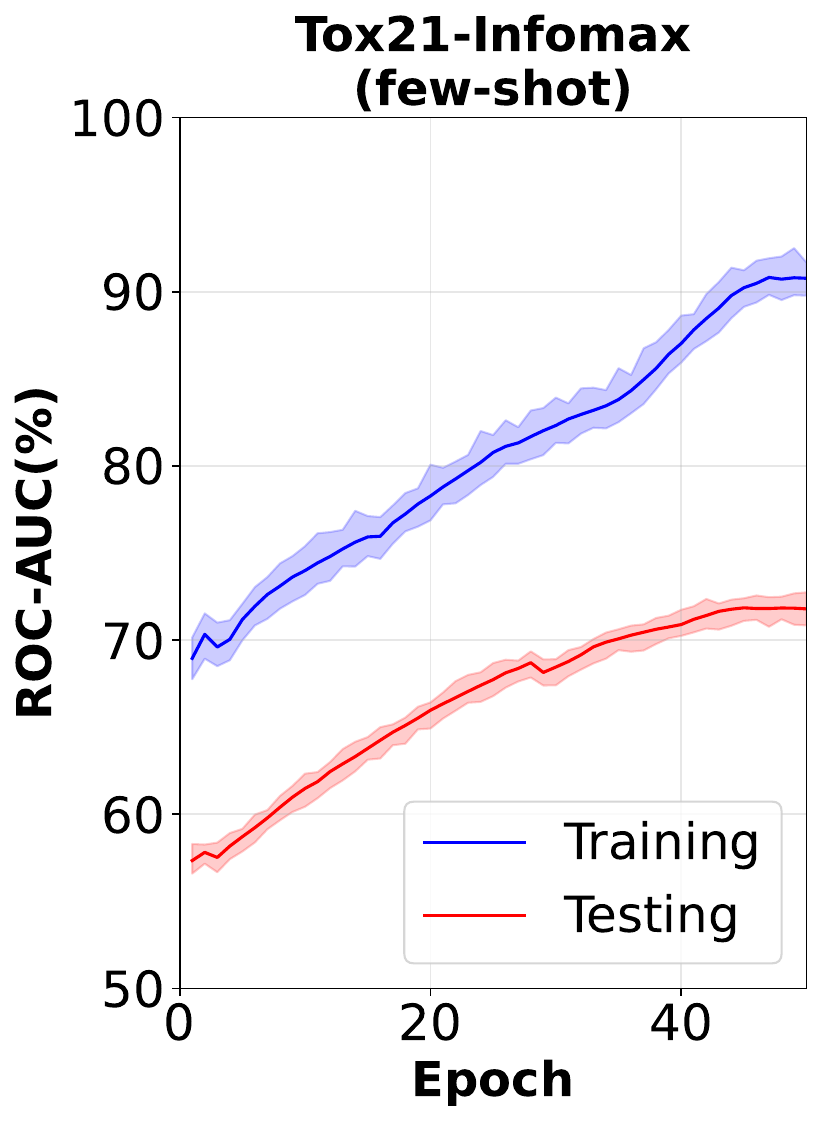}
        }
    \subfigure {
        \label{fig:training curve8}
        \includegraphics[width=0.2\linewidth]{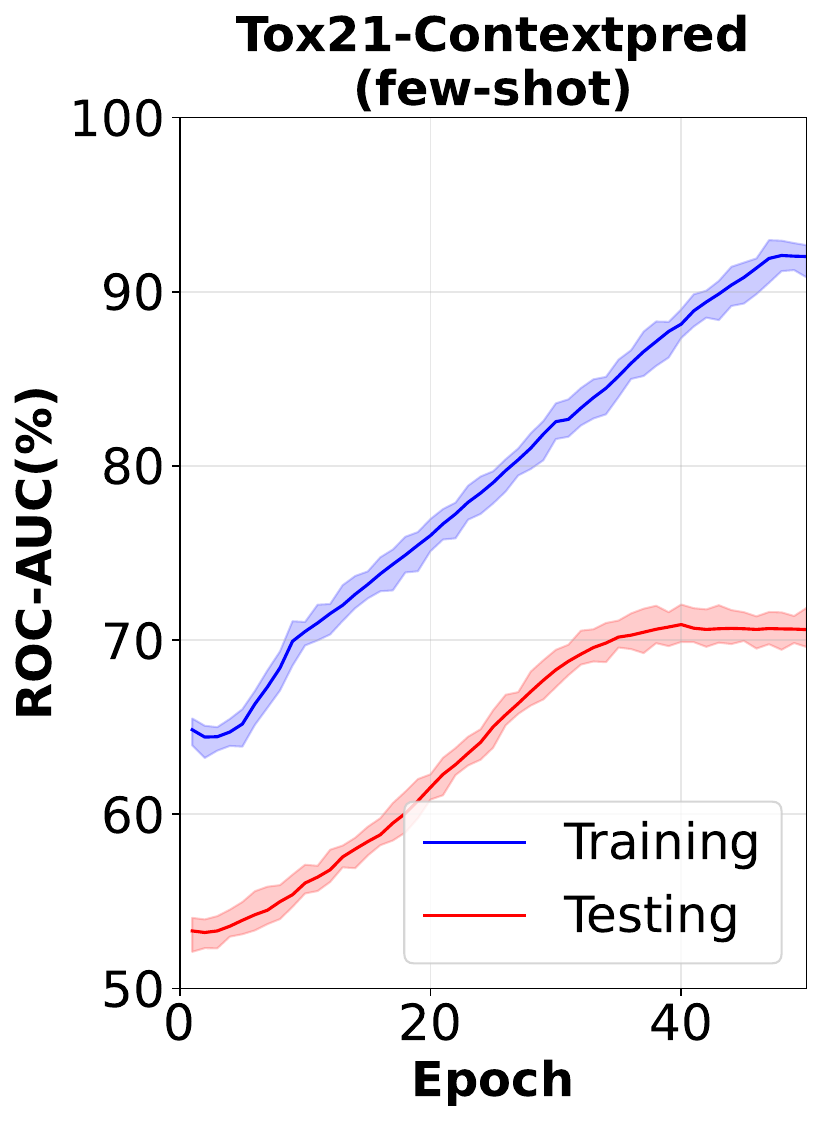}
        }
    \vskip -0.1in
    \caption{The ROC-AUC curves for the training and testing sets.}
    \label{fig:training curve}
\end{figure*}

We illustrate the ROC-AUC curve of LEAP for the training and testing sets across five random seeds on the BBBP and Tox21 datasets pre-trained with Infomax and ContextPred under both full-shot and few-shot scenarios in Figure~\ref{fig:training curve}. The ROC-AUC curve exhibits a smooth and consistently increasing trend, which reflects the stability and effectiveness of the training process.

\subsection{Efficiency Study}
Table~\ref{tab:time} presents a comparison of average training time (in seconds) for 50 epochs across all pre-training strategies for LEAP and all baselines in the few-shot scenario. LEAP achieves an acceptable computational cost compared to GPF, GPF-plus, and SUPT variants, while demonstrating significant computational advantages over RELIEF. In addition, we further provide a theoretical level efficiency analysis in Appendix~\ref{appendix:complexity}.

\begin{table}[htbp]
    \caption{Training time comparisons (s). 50-shot for graph-level task and 10-shot for node-level task.}
    \resizebox{\linewidth}{!}{
        \centering
       \begin{tabular}{cccccccc}

        \toprule
        
        Task-level & Dataset & GPF & GPF-plus & SUPT\textsubscript{soft} & SUPT\textsubscript{hard} & RELIEF & LEAP\\

        \midrule

        \multirow{8}{*}{Graph} & BBBP & 55.3& 57.6& 50.8& 50.5& 123.7& 67.2\\
        & Tox21 & 56.0& 58.3& 51.3& 51.3& 127.9& 68.0\\
        & ToxCast & 55.1& 58.3& 50.4& 50.3& 111.4& 62.8\\
        & SIDER & 47.3& 48.2& 45.5& 45.2& 93.9& 52.0\\
        & ClinTox & 50.3& 51.0& 48.2& 47.9& 104.2& 56.4\\
        & MUV & 108.3& 110.0& 102.3& 102.2& 225.6& 125.3\\
        & HIV & 127.2& 129.8& 117.2& 117.0& 264.8& 150.8\\
        & BACE & 55.4& 56.2& 50.3& 50.0& 128.3& 67.5\\
        
        \addlinespace[1pt]
        \midrule[0.1pt]
        \addlinespace[2pt]

        \multirow{5}{*}{Node} & Cora & 24.2& 25.3& 21.0& 21.2& 47.9& 29.9\\
        & CiteSeer & 28.3& 29.9& 27.1& 26.9& 56.9& 35.7\\
        & PubMed & 36.7& 37.7& 35.4& 35.4& 74.5& 43.0\\
        & Computers & 50.3& 51.7& 46.7& 46.5& 112.2& 61.8\\
        & Photos & 43.4& 44.1& 41.0& 40.8& 84.2& 49.3\\
        \bottomrule
        \end{tabular}
    }
    \label{tab:time}
    \vspace{-1mm}
\end{table}

\subsection{Hyper-parameter Analysis}
We analyze the critical hyper-parameters in LEAP, including the basic universal graph prompt vector number $k$, the prompt edit range $\vartheta$, the policy networks update interval $h$, and the policy finite horizon (step) $T$. For both graph- and node-level tasks, we report the average performance across all datasets for each pre-training strategy. \textbf{Please refer to Appendix~\ref{appendix:hyper-parameter_analysis} for the figures and detailed analyses mentioned in this subsection.}

\subsubsection{Basic Universal Graph Prompt Vector Number $k$}
From Figure~\ref{fig:k-graph} and Figure~\ref{fig:k-node}, we observe that the optimal value of the number of basic universal graph prompt vectors $k$ is $10$. Moreover, selecting $k$ as $5$ or $20$ has no significant impact on performance, demonstrating LEAP's robustness to hyper-parameters $k$.

\subsubsection{Prompt Edit Range $\vartheta$}
From Figure~\ref{fig:range-graph} and Figure~\ref{fig:range-node}, we observe that the optimal value of prompt edit range $\vartheta$ is $0.5$. Moreover, varying $\vartheta$ within reasonable limits has minimal impact on the model's performance, highlighting LEAP's robustness to the hyper-parameter $\vartheta$.

\subsubsection{Policy Networks Update Interval $h$}
From Figure~\ref{fig:h-graph} and Figure~\ref{fig:h-node}, we observe that the optimal value of policy network update interval $h$ is $3$ for the graph-level task and $4$ for the node-level task. Moreover, varying $h$ within reasonable limits has minimal impact on the model's performance, showing LEAP's robustness to the hyper-parameter $h$.

\subsubsection{Policy Finite Horizon (Step) $T$}
From Figure~\ref{fig:t-graph} and Figure~\ref{fig:t-node}, we observe that the optimal value of policy finite horizon (step) $T$ is $N/4$ in the full-shot scenario and $N/2$ in the few-shot scenario. Moreover, varying $T$ within reasonable limits has little effect on the model's performance, demonstrating LEAP's robustness to the hyper-parameter $T$.

\subsection{Additional Experiments}
\label{subsec:additional_experiments}
Comprehensive and additional experiments are presented in the Appendices, with key conclusions summarized here.

\textbf{Few-shot Scenario} We provide extra experiments in few-shot scenarios. Experiments demonstrate that LEAP achieves the best performance compared to baselines on both graph- and node-level tasks across various pre-training strategies in the few-shot scenario. Detailed experiment results and analysis are in Appendix~\ref{appendix:few-shot}.

\textbf{Ablation Study} We conduct ablation studies to demonstrate the effectiveness of each component in LEAP. These include replacing the basic universal graph prompt with GPF or the parameter-dense GPF-plus, as well as removing the editing coverage rate from the reward function. The results validate the significance of each component. Detailed experiment results and analysis are in Appendix~\ref{appendix:ablation_study}.

\textbf{Hyper-parameter Analysis}
We analyze the sensitivity of key hyper-parameters in LEAP, and the experimental results confirm its robust performance across diverse hyper-parameter settings. Detailed experiment results and analysis are in Appendix~\ref{appendix:hyper-parameter_analysis}. 


\section{Related Work}
Due to page limitations, we review recent works and their contributions in Appendix~\ref{appendix:related_work}.

\section{Conclusion}
In this paper, we first establish that assigning prompts to all nodes is a critical theoretical constraint for universal graph prompt tuning to achieve equivalent performance to any prompting function. This insight serves as a foundational guideline for refining future research directions in universal graph prompt tuning. To address this, we propose a novel model, Learning and Editing Universal Graph Prompt Tuning (\model), which preserves the theoretical foundation of universal graph prompt tuning while pursuing more ideal prompts. Extensive experiments on both graph- and node-level tasks, across diverse pre-training strategies in full-shot and few-shot scenarios, demonstrate the effectiveness of \model. We highlight that LEAP is not only an effective model but also establishes an ideal paradigm for universal graph prompt tuning.


\begin{acks}
This work was supported by the Hong Kong UGC General Research Fund no. 17203320 and 17209822, and the project grants from the HKU-SCF FinTech Academy.
\end{acks}

\balance
\bibliographystyle{ACM-Reference-Format}


\newpage
\nobalance
\appendix

\section{Supplementary Materials for Methodology}

\subsection{PPO Surrogate Objectives}
\label{appendix:ppo}
\textbf{Discrete Action Policy:}

The surrogate objective $\mathcal{L}_{d}^{\mathrm{PPO}}$ for the discrete policy $\pi_{\varsigma_d}(\cdot)$ is defined as:
\begin{equation}
\begin{aligned}
\mathbb{E}_{(\mathbf{s}, a) \sim \pi_{\text{old}}} &\left[\min \left(\kappa A(\mathbf{s}, a), \operatorname{clip}\left(\kappa, 1-\epsilon, 1+\epsilon\right) A(\mathbf{s}, a)\right)\right] \\ 
+& \beta_d \mathbb{E}_{\pi}\left[H\left(\pi_{\varsigma_d}(a | s)\right)\right],
\end{aligned}
\end{equation}
where $\pi_{\text{old}}$ represents the frozen parameters of the policy from the previous iteration. $\kappa = \frac{\pi_{\varsigma_d}(a | \mathbf{s})}{\pi_{\text{old}}(a |\mathbf{s})}$. $A(\mathbf{s}, a)$ is the advantage function for discrete actions, computed via Generalized Advantage Estimation (GAE). $\operatorname{clip}(\cdot)$ restricts the policy ratio to $[1-\epsilon, 1+\epsilon]$ to prevent destabilizing policy updates. $\beta_d$ is the entropy regularization coefficient, and $H(\cdot)$ measures the policy entropy to encourage exploration.

\noindent \textbf{Continuous Action Policy:}

The surrogate objective $\mathcal{L}_{c}^{\mathrm{PPO}}$ for the continuous policy $\pi_{\varsigma_c}(\cdot)$ is defined as:
\begin{equation}
\begin{aligned}
\mathbb{E}_{(\mathbf{s}, a, z) \sim \pi_{\text{old}}}&\left[\min \left(\Gamma A(\mathbf{s}, a, z), \operatorname{clip}\left(\Gamma, 1-\epsilon, 1+\epsilon\right) A(\mathbf{s}, a, z)\right)\right] \\+&\beta_c \mathbb{E}_\pi\left[H\left(\pi_{\varsigma_c}(z |\mathbf{s}, a)\right)\right],
\end{aligned}
\end{equation}
where $A(\mathbf{s}, a, z)$ incorporates the joint effect of discrete and continuous actions on the advantage estimation. $\Gamma = \frac{\pi_{\varsigma_c}(z |\mathbf{s}, a)}{\pi_{\text{old}}(z |\mathbf{s}, a)}$. $\beta_c$ is the entropy regularization coefficient.

\noindent \textbf{Advantage Function Computation:}

The advantage functions $A(\mathbf{s}, a)$ and $A(\mathbf{s}, a, z)$ are derived from the critic policy network $V_{\varphi}(\cdot)$ using GAE:
\begin{equation}
A(\mathbf{s}^t, a^t, z^t)=\sum_{l=0}^{T-t}(\gamma \lambda)^l \delta^{t+l}, \quad \delta^t=r^t+\gamma V_{\varphi}(\mathbf{s}^{t+1})-V_{\varphi}(\mathbf{s}^t),
\end{equation}
with $\gamma \in[0,1)$ as the discount factor and $\lambda \in[0,1]$ as the GAE smoothing coefficient.

\section{Experiment Settings}
\label{appendix:experiment_settings}
\subsection{Evaluation Tasks}
\label{subsec:task}
For the graph-level task in the few-shot scenario, we follow previous works \citep{fang2023universal,lee2024subgraph} to adopt the 50-shot setting to ensure experimental fairness. For the node-level task, inspired by previous works \citep{sun2022gppt,sun2023all}, we generate induced graphs for nodes, enabling universal prompt tuning to be extended to node-level tasks by operating on $n$-hop subgraphs. For the node-level task in the few-shot scenario, we follow previous work \citep{zhu2024relief} to adopt the 10-shot setting to ensure experimental fairness. This ensures an appropriate number of nodes in the induced subgraphs.

\subsection{Model Architectures}
\label{appendix:architectures}
We follow previous works \citep{fang2023universal,lee2024subgraph,zhu2024relief} to adopt 5-layer GIN and 2-layer GIN for graph- and node-level tasks, respectively, to ensure experimental fairness. Hidden dimensions for GIN are set as $300$ and $128$ for graph- and node-level tasks, respectively. We adopt weight decay $5e^{-4}$ for the projection head in the graph-level task and utilize Singular Value
Decomposition (SVD) to reduce the initial features of each dataset
to $100$ dimensions for the node-level task. In addition, there are some consistent settings for both graph- and node-level tasks. Since our RL approach differs from RELIEF, which requires generating universal graph prompts independently, we only need to edit basic universal graph prompts. As a result, our RL approach is not highly sensitive to hyper-parameters. For all scenarios, we keep the relevant hyper-parameters fixed. Specifically, we set $0.5$ for dropout rate, $5e^{-4}$ for learning rate of actors and critic, $0.99$ for discount factor $\gamma$, $0.95$ for the GAE smoothing coefficient $\lambda$, $0.2$ for PPO clip range, $0.01$ for entropy regularization coefficients $\beta_d$ and $\beta_c$. Parameter settings are summarized in Table~\ref{tab:parameter}.

\begin{table}[!htbp]
\small
  \caption{Parameter settings for graph- and node-level tasks.}
  \resizebox{\linewidth}{!}{
      \centering
      \begin{tabular}{c|cc}
        \toprule
        Parameter & Graph-level& Node-level\\  \midrule
        GIN Layer Number & $5$& $2$\\
        GIN Hidden Dimension & $300$& $128$\\
        Projection Head Weight Decay & $5e^{-4}$ & -\\
        SVD Dimensions & - & $100$\\
        Dropout Rate & \multicolumn{2}{c}{$0.5$} \\
        Learning Rate for Both Actors and Critic & \multicolumn{2}{c}{$5e^{-4}$} \\
        Discount Factor $\gamma$ & \multicolumn{2}{c}{$0.99$} \\ 
        GAE Smoothing Coefficient $\lambda$ & \multicolumn{2}{c}{$0.95$} \\ 
        PPO Clip Range & \multicolumn{2}{c}{$0.2$} \\ 
        Entropy Regularization Coefficients $\beta_d$ and $\beta_c$& \multicolumn{2}{c}{$0.01$} \\ 
        \bottomrule
      \end{tabular}
  }
  \label{tab:parameter}
\end{table}

\subsection{Datasets}
\label{appendix:datasets}
For the graph-level task, we follow previous works \citep{zhu2024relief,lee2024subgraph,fang2023universal} to utilize Chemical datasets\footnote{\href{https://github.com/snap-stanford/pretrain-gnns/tree/master/chem/model_gin/}{https://github.com/snap-stanford/pretrain-gnns/tree/master/chem/model$\_$gin/}.} \citep{hu2020strategies} to pre-train GNN models. These datasets consist of both unlabeled and labeled molecules, enabling comprehensive model training. Specifically, $2$ million unlabeled molecules from the ZINC15 database are employed for node-level self-supervised pre-training. For graph-level multi-task supervised pre-training, a preprocessed subset of the ChEMBL database is used, which comprises $456$k labeled molecules. This subset provides a diverse chemical space that enhances the model’s generalization across various biochemical tasks. For downstream graph classification tasks, we utilize $8$ binary classification datasets focused on molecular property prediction related to biophysics and physiology \citep{wu2018moleculenet}. We split each dataset into training, validation, and testing sets with ratios of $80$\%, $10$\%, and $10$\%, respectively. Molecules are divided based on their scaffolds, i.e., molecular graph substructures \cite{ramsundar2019deep}, and the clusters are reorganized by prioritizing the most frequently occurring scaffolds in the training set. This approach ensures that the validation and testing sets contain structurally distinct molecules \cite{hu2020strategies}, enabling an assessment of the model’s out-of-distribution generalization capabilities. The statistics of these downstream datasets are summarized in Table~\ref{tab:graph_dataset}.

\begin{table*}[!htbp]
\small
  \caption{Statistics of the downstream datasets used in the graph-level task.}
      \centering
      \begin{tabular}{c|cccccccc}
        \toprule
        Datasets & BBBP& Tox21& ToxCast& SIDER& ClinTox& MUV& HIV& BACE\\ \midrule
        \# Molecules & 2,039& 7,831& 8,575& 1,427& 1,478& 93,087& 41,127& 1,513\\
        \# Tasks & 1& 12& 617& 27& 2& 17& 1& 1\\
        \bottomrule
      \end{tabular}
  \label{tab:graph_dataset}
\end{table*}

For the graph-level task, we follow previous works \citep{zhu2024relief,sun2022gppt} to utilize Cora, Citeseer, Pubmed \cite{kipf2017semi}, and Amazon-Co-Buy (Computer and Photo) \cite{shchur2018pitfalls}, where the first three datasets represent each node as a publication with a sparse bag-of-words feature vector, and edges signify citation links, the last two datasets are segments of the Amazon co-purchase graph, with nodes representing products, edges indicating frequent co-purchases, node features encoded as bag-of-words from product reviews, and class labels derived from product categories. Taking into account the average degree of different datasets, we set varying values of $n$ for generating $n$-hop subgraphs: specifically $4$, $3$, $2$, $2$, and $2$ for Cora, Citeseer, Pubmed, Computers, and Photo, respectively. The statistics of these datasets are summarized in Table~\ref{tab:node_dataset}.

\begin{table}[!htbp]
\small
  \caption{Statistics of the datasets used in the node-level task.}
      \centering
      \begin{tabular}{c|ccccc}
        \toprule
        Datasets & Cora& Citeseer& Pubmed& Computers& Photos\\ \midrule
        \# Nodes & 2,708& 3,327& 19,717& 13,752& 7,650\\
        \# Edges & 5,429& 9,104& 88,648& 491,722& 238,162\\
        \# Features & 1,433& 3,703& 500& 767& 745\\
        \# Classes & 7& 6& 3& 10& 8\\
        \bottomrule
      \end{tabular}
  \label{tab:node_dataset}
\end{table}

\subsection{Pre-training Strategies}
\label{appendix:strategies}
To ensure fair evaluation, we follow previous works \citep{fang2023universal,zhu2024relief} by adopting four pre-training strategies for graph-level tasks and two pre-training strategies for node-level tasks. Specifically, for the graph-level task, we employ five widely used strategies to pre-train the graph models, including Deep Graph Infomax (Infomax) \citep{velivckovic2019deep}, Attribute Masking (AttrMasking) \citep{hu2020strategies}, Context Prediction (ContextPred) \citep{hu2020strategies}, and Graph Contrastive Learning (GCL) \citep{you2020graph}:
\begin{itemize}[leftmargin=*]
    \item \textbf{Infomax} Deep Graph Infomax learns expressive representations for graphs or nodes by maximizing the mutual information between graph-level representations and substructure-level representations across different granularities.
    \item \textbf{AttrMasking} Attribute Masking involves masking node or edge attributes and tasking GNNs with predicting the masked attributes based on the surrounding structural information.
    \item \textbf{ContextPred} Context Prediction leverages subgraphs to predict their surrounding graph structures, aiming to map nodes with similar structural contexts to closely aligned embeddings in the representation space.
    \item \textbf{GCL} Graph Contrastive Learning aims to embed augmented versions of an anchor graph closer to each other (positive samples) while pushing apart the embeddings of different graphs (negative samples). For generating positive and negative samples, we adopt the augmentation strategies \citep{you2020graph}. 
\end{itemize}
For the node-level task, we utilize two widely used strategies to pre-train the graph models: Masked Edge Prediction (MaskedEdge) \citep{sun2022gppt} and Edge Connectivity Contrastive Learning (ContraEdge) \citep{liu2023graphprompt}:
\begin{itemize}[leftmargin=*]
    \item \textbf{MaskedEdge} Masked Edge Prediction, proposed by GPPT \citep{sun2022gppt}, leverages a binary cross-entropy loss to optimize the predicted interaction probabilities between nodes and their adjacency matrix, thereby capturing rich edge information.
    \item \textbf{ContraEdge} Edge Connectivity Contrastive Learning employed by GraphPrompt \cite{liu2023graphprompt} determines positive and negative node pairs based on edge connectivity, encouraging the model to bring embeddings of positively connected node pairs closer while pushing apart embeddings of negatively connected pairs.
\end{itemize}

\subsection{Baselines}
\label{appendix:baselines}
For the graph-level task, we compare LEAP against Fine-tuning (FT), GPF, GPG-plus \citep{fang2023universal}, SUPT\textsubscript{soft}, SUPT\textsubscript{hard} \citep{lee2024subgraph}, and RELIEF \citep{zhu2024relief}. In the node-level task, we include all the above baselines and additionally compare with GPPT \citep{sun2022gppt} for the MaskedEdge pre-training strategy and GPrompt \citep{liu2023graphprompt} for the ContraEdge pre-training strategy. Specifically, given a pre-trained graph model $f_\theta(\cdot)$ and a task-specific projection head $g_\phi(\cdot)$, FT tunes the parameters of the pre-trained graph model $f_\theta(\cdot)$ and a task-specific projection head $g_\phi(\cdot)$ simultaneously during the downstream training stage. GPF, GPG-plus, SUPT\textsubscript{soft}, SUPT\textsubscript{hard}, and RELIEF freeze the parameters of the pre-trained model $f_\theta(\cdot)$ and introduce extra prompt vectors $p$. These baselines tune the parameters of the prompt vectors $p$ and a task-specific projection head $g_\phi(\cdot)$ simultaneously during the downstream training stage. For the node-level task, we include all baselines used in the graph-level task and additionally incorporate GPPT and GraphPrompt as baselines. These methods are specifically designed for graph models pre-trained at the edge level and are well-suited for transfer to node classification. Notably, we did not include All in One \citep{sun2023all}, as it relies on a frozen graph model specifically pre-trained using GCL.

\subsection{Hyper-parameter Settings}
\label{appendix:hyper-parameter}
We detail the tuning range of all hyper-parameters in our LEAP to enhance the reproducibility. Following previous works \cite{zhu2024relief,sun2022gppt,fang2023universal}, we tune training epochs within $[50, 100]$ and $[50,100,300]$ for graph- and node-level tasks, respectively. Moreover, there are some consistent hyper-parameter tuning ranges for both graph- and node-level tasks. Specifically, we conduct a grid search for the graph loader batch size within $[8,16,32,64]$, the basic universal graph prompt vector number $k$ within $[5,10,20]$, the prompt edit range $\vartheta$ within $[0.1,0.5,1]$, reward function balancing hyper-parameter $\lambda_c$ within $[1e^{-5}, 1e^{-4}, 1e^{-3}]$, policy networks update interval $h$ within $[1,2,3,4]$, projection head learning rate within $[5e^{-4},1e^{-3},5e^{-3}]$, projection head layer number within $[1,2,3]$, PPO mini-batch size within $[32,64,128,256]$, and policy finite horizon (step) $T$ within $[N/8, N/4, N/2, N, 2N]$. Hyper-parameter settings are summarized in Table~\ref{tab:hyper-parameter}.

\begin{table}[!htbp]
\small
  \caption{Hyper-parameter settings.}
  \resizebox{\linewidth}{!}{
      \centering
      \begin{tabular}{c|cc}
        \toprule
        Parameter & Graph-level& Node-level\\  \midrule
        Training Epochs & $[50,100]$& $[50,100,300]$\\
        Graph Loader Batch Size & \multicolumn{2}{c}{$[8,16,32,64]$} \\
        Basic Universal Graph Prompt Vector Number $k$ & \multicolumn{2}{c}{$[5,10,20]$} \\ 
        Prompt Edit Range $\vartheta$ & \multicolumn{2}{c}{$[0.1,0.5,1]$} \\ 
        Reward Balancing Parameter $\lambda_c$ & \multicolumn{2}{c}{$[1e^{-5},1e^{-4},1e^{-3}]$} \\
        Policy Networks Update Interval $h$ & \multicolumn{2}{c}{$[1,2,3,4]$} \\ 
        Projection Head Learning Rate & \multicolumn{2}{c}{$[5e^{-4},1e^{-3},5e^{-3}]$} \\  
        Projection Head Layer Number& \multicolumn{2}{c}{$[1,2,3]$} \\ 
        PPO Mini-batch Size & \multicolumn{2}{c}{$[32,64,128,256]$} \\
        Policy Finite Horizon (Step) $T$ & \multicolumn{2}{c}{$[N/8, N/4, N/2, N, 2N]$}\\
        \bottomrule
      \end{tabular}
  }
  \label{tab:hyper-parameter}
  \vskip -0.1in
\end{table}

\begin{table*}[!htbp]
  \caption{ROC-AUC (\%) and standard deviation for graph classification on molecule property prediction benchmark under 50-shot scenario with various pre-training and tuning strategies. The best results for each dataset and pre-training strategy are bolded, while the runners-up are underlined.}
  \small
      \centering
      \begin{tabular}{ccccccccccc}
        \toprule
        & \makecell{Tuning \\ Strategy} & BBBP & Tox21 & ToxCast & SIDER & ClinTox & MUV & HIV & BACE & \textbf{Avg.} \\
    
        \midrule
    
        \multirow{7}{*}{\rotatebox{90}{Infomax}} &
        FT & 
        65.03 {\scriptsize \textcolor{gray}{$\pm$1.13}} &
        71.41 {\scriptsize \textcolor{gray}{$\pm$0.69}} &
        58.04 {\scriptsize \textcolor{gray}{$\pm$0.36}} &
        53.47 {\scriptsize \textcolor{gray}{$\pm$0.48}} &
        61.11 {\scriptsize \textcolor{gray}{$\pm$2.40}} &
        65.20 {\scriptsize \textcolor{gray}{$\pm$0.45}} &
        65.87 {\scriptsize \textcolor{gray}{$\pm$2.41}} &
        65.65 {\scriptsize \textcolor{gray}{$\pm$0.89}} & 
        63.22 \\
        & GPF &
        66.85 {\scriptsize \textcolor{gray}{$\pm$1.14}} &
        71.19 {\scriptsize \textcolor{gray}{$\pm$0.35}} &
        58.38 {\scriptsize \textcolor{gray}{$\pm$0.20}} &
        53.66 {\scriptsize \textcolor{gray}{$\pm$0.45}} &
        65.00 {\scriptsize \textcolor{gray}{$\pm$2.75}} &
        67.17 {\scriptsize \textcolor{gray}{$\pm$0.13}} &
        66.51 {\scriptsize \textcolor{gray}{$\pm$1.91}} &
        65.59 {\scriptsize \textcolor{gray}{$\pm$1.29}} & 
        64.29 \\
        & GPF-plus &
        66.94 {\scriptsize \textcolor{gray}{$\pm$1.70}} &
        71.50 {\scriptsize \textcolor{gray}{$\pm$0.32}} &
        58.55 {\scriptsize \textcolor{gray}{$\pm$0.33}} &
        \underline{53.89} {\scriptsize \textcolor{gray}{$\pm$0.72}} &
        65.85 {\scriptsize \textcolor{gray}{$\pm$1.99}} &
        \underline{67.22} {\scriptsize \textcolor{gray}{$\pm$0.11}} &
        66.91 {\scriptsize \textcolor{gray}{$\pm$1.44}} &
        66.23 {\scriptsize \textcolor{gray}{$\pm$1.95}} & 
        64.64 \\
        & SUPT\textsubscript{soft} &
        66.21 {\scriptsize \textcolor{gray}{$\pm$0.91}} &
        71.40 {\scriptsize \textcolor{gray}{$\pm$0.26}} &
        \underline{58.73} {\scriptsize \textcolor{gray}{$\pm$0.29}} &
        53.49 {\scriptsize \textcolor{gray}{$\pm$0.49}} &
        65.82 {\scriptsize \textcolor{gray}{$\pm$3.05}} &
        66.69 {\scriptsize \textcolor{gray}{$\pm$0.31}} &
        66.59 {\scriptsize \textcolor{gray}{$\pm$1.88}} &
        66.44 {\scriptsize \textcolor{gray}{$\pm$2.01}} & 
        64.42 \\
        & SUPT\textsubscript{hard} &
        66.00 {\scriptsize \textcolor{gray}{$\pm$1.02}} &
        71.23 {\scriptsize \textcolor{gray}{$\pm$0.40}} &
        58.60 {\scriptsize \textcolor{gray}{$\pm$0.26}} &
        53.73 {\scriptsize \textcolor{gray}{$\pm$0.40}} &
        65.83 {\scriptsize \textcolor{gray}{$\pm$2.55}} &
        66.75 {\scriptsize \textcolor{gray}{$\pm$1.19}} &
        66.81 {\scriptsize \textcolor{gray}{$\pm$2.26}} &
        66.80 {\scriptsize \textcolor{gray}{$\pm$0.75}} & 
        64.47 \\
        & RELIEF &
        \underline{67.21} {\scriptsize \textcolor{gray}{$\pm$1.35}} &
        \underline{71.51} {\scriptsize \textcolor{gray}{$\pm$0.28}} &
        58.70 {\scriptsize \textcolor{gray}{$\pm$0.23}} &
        53.81 {\scriptsize \textcolor{gray}{$\pm$0.47}} &
        \underline{66.20} {\scriptsize \textcolor{gray}{$\pm$2.15}} &
        67.09 {\scriptsize \textcolor{gray}{$\pm$0.16}} &
        \underline{67.04} {\scriptsize \textcolor{gray}{$\pm$1.51}} &
        \textbf{67.56} {\scriptsize \textcolor{gray}{$\pm$0.81}} & 
        \underline{64.89} \\
        & LEAP &
        \textbf{68.27} {\scriptsize \textcolor{gray}{$\pm$1.09}} & \textbf{71.85} {\scriptsize \textcolor{gray}{$\pm$0.21}} & \textbf{58.79} {\scriptsize \textcolor{gray}{$\pm$0.22}} & \textbf{54.00} {\scriptsize \textcolor{gray}{$\pm$0.56}} & \textbf{66.44} {\scriptsize \textcolor{gray}{$\pm$2.29}} & \textbf{67.28} {\scriptsize \textcolor{gray}{$\pm$0.49}} & \textbf{67.15} {\scriptsize \textcolor{gray}{$\pm$1.38}} & \underline{67.31} {\scriptsize \textcolor{gray}{$\pm$0.91}} & \textbf{65.14}\\
    
        \addlinespace[1pt]
        \midrule[0.1pt]
        \addlinespace[2pt]
    
        \multirow{7}{*}{\rotatebox{90}{AttrMasking}} &
        FT & 
        66.26 {\scriptsize \textcolor{gray}{$\pm$0.69}} &
        72.36 {\scriptsize \textcolor{gray}{$\pm$0.28}} &
        57.50 {\scriptsize \textcolor{gray}{$\pm$0.36}} &
        55.24 {\scriptsize \textcolor{gray}{$\pm$0.58}} &
        64.55 {\scriptsize \textcolor{gray}{$\pm$3.56}} &
        62.09 {\scriptsize \textcolor{gray}{$\pm$1.01}} &
        58.96 {\scriptsize \textcolor{gray}{$\pm$1.39}} &
        66.86 {\scriptsize \textcolor{gray}{$\pm$1.50}} & 
        62.98 \\
        & GPF &
        66.41 {\scriptsize \textcolor{gray}{$\pm$1.03}} &
        72.55 {\scriptsize \textcolor{gray}{$\pm$0.24}} &
        58.15 {\scriptsize \textcolor{gray}{$\pm$0.22}} &
        55.61 {\scriptsize \textcolor{gray}{$\pm$1.27}} &
        69.52 {\scriptsize \textcolor{gray}{$\pm$2.72}} &
        62.70 {\scriptsize \textcolor{gray}{$\pm$0.30}} &
        58.80 {\scriptsize \textcolor{gray}{$\pm$1.50}} &
        68.26 {\scriptsize \textcolor{gray}{$\pm$1.14}} & 
        64.00 \\
        & GPF-plus &
        66.83 {\scriptsize \textcolor{gray}{$\pm$0.89}} &
        \underline{72.71} {\scriptsize \textcolor{gray}{$\pm$0.45}} &
        58.21 {\scriptsize \textcolor{gray}{$\pm$0.30}} &
        55.98 {\scriptsize \textcolor{gray}{$\pm$0.99}} &
        70.94 {\scriptsize \textcolor{gray}{$\pm$2.01}} &
        62.99 {\scriptsize \textcolor{gray}{$\pm$0.37}} &
        59.13 {\scriptsize \textcolor{gray}{$\pm$1.31}} &
        68.90 {\scriptsize \textcolor{gray}{$\pm$1.00}} & 
        64.46 \\
        & SUPT\textsubscript{soft} &
        66.33 {\scriptsize \textcolor{gray}{$\pm$1.17}} &
        72.63 {\scriptsize \textcolor{gray}{$\pm$0.48}} &
        \underline{58.24} {\scriptsize \textcolor{gray}{$\pm$0.40}} &
        55.79 {\scriptsize \textcolor{gray}{$\pm$1.11}} &
        68.89 {\scriptsize \textcolor{gray}{$\pm$3.15}} &
        62.85 {\scriptsize \textcolor{gray}{$\pm$0.20}} &
        58.98 {\scriptsize \textcolor{gray}{$\pm$1.26}} &
        67.05 {\scriptsize \textcolor{gray}{$\pm$1.61}} & 
        63.85 \\
        & SUPT\textsubscript{hard} &
        66.80 {\scriptsize \textcolor{gray}{$\pm$1.20}} &
        72.74 {\scriptsize \textcolor{gray}{$\pm$0.30}} &
        58.19 {\scriptsize \textcolor{gray}{$\pm$0.35}} &
        55.88 {\scriptsize \textcolor{gray}{$\pm$1.03}} &
        69.41 {\scriptsize \textcolor{gray}{$\pm$2.09}} &
        62.94 {\scriptsize \textcolor{gray}{$\pm$0.27}} &
        59.11 {\scriptsize \textcolor{gray}{$\pm$1.10}} &
        68.98 {\scriptsize \textcolor{gray}{$\pm$0.79}} & 
        64.26 \\
        & RELIEF &
        \underline{66.86} {\scriptsize \textcolor{gray}{$\pm$0.61}} &
        72.44 {\scriptsize \textcolor{gray}{$\pm$0.38}} &
        \textbf{58.43} {\scriptsize \textcolor{gray}{$\pm$0.27}} &
        \underline{56.04} {\scriptsize \textcolor{gray}{$\pm$0.85}} &
        \underline{72.79} {\scriptsize \textcolor{gray}{$\pm$1.99}} &
        \underline{63.09} {\scriptsize \textcolor{gray}{$\pm$0.14}} &
        \textbf{59.40} {\scriptsize \textcolor{gray}{$\pm$0.62}} &
        \underline{69.37} {\scriptsize \textcolor{gray}{$\pm$0.70}} & 
        \underline{64.80} \\
        & LEAP &
        \textbf{66.95} {\scriptsize \textcolor{gray}{$\pm$0.35}} & \textbf{72.79} {\scriptsize \textcolor{gray}{$\pm$0.22}} & \textbf{58.43} {\scriptsize \textcolor{gray}{$\pm$0.20}} & \textbf{56.17} {\scriptsize \textcolor{gray}{$\pm$0.91}} & \textbf{73.33} {\scriptsize \textcolor{gray}{$\pm$2.22}} & \textbf{63.42} {\scriptsize \textcolor{gray}{$\pm$0.20}} & \underline{59.33} {\scriptsize \textcolor{gray}{$\pm$0.82}} & \textbf{69.45} {\scriptsize \textcolor{gray}{$\pm$0.88}} & \textbf{64.98}\\
    
        \addlinespace[1pt]
        \midrule[0.1pt]
        \addlinespace[2pt]
    
        \multirow{7}{*}{\rotatebox{90}{ContextPred}} &
        FT & 
        62.69 {\scriptsize \textcolor{gray}{$\pm$0.97}} &
        70.28 {\scriptsize \textcolor{gray}{$\pm$0.46}} &
        57.90 {\scriptsize \textcolor{gray}{$\pm$0.44}} &
        56.33 {\scriptsize \textcolor{gray}{$\pm$0.70}} &
        61.27 {\scriptsize \textcolor{gray}{$\pm$2.53}} &
        63.30 {\scriptsize \textcolor{gray}{$\pm$1.07}} &
        59.35 {\scriptsize \textcolor{gray}{$\pm$1.78}} &
        66.74 {\scriptsize \textcolor{gray}{$\pm$2.01}} & 
        62.23 \\
        & GPF &
        61.77 {\scriptsize \textcolor{gray}{$\pm$1.03}} &
        70.42 {\scriptsize \textcolor{gray}{$\pm$0.33}} &
        \underline{58.48} {\scriptsize \textcolor{gray}{$\pm$0.24}} &
        56.37 {\scriptsize \textcolor{gray}{$\pm$0.40}} &
        60.85 {\scriptsize \textcolor{gray}{$\pm$2.33}} &
        64.28 {\scriptsize \textcolor{gray}{$\pm$0.38}} &
        \underline{59.64} {\scriptsize \textcolor{gray}{$\pm$2.00}} &
        68.42 {\scriptsize \textcolor{gray}{$\pm$1.37}} & 
        62.53 \\
        & GPF-plus &
        62.12 {\scriptsize \textcolor{gray}{$\pm$1.30}} &
        70.48 {\scriptsize \textcolor{gray}{$\pm$0.37}} &
        58.34 {\scriptsize \textcolor{gray}{$\pm$0.26}} &
        \underline{56.79} {\scriptsize \textcolor{gray}{$\pm$0.39}} &
        62.03 {\scriptsize \textcolor{gray}{$\pm$2.40}} &
        \underline{64.59} {\scriptsize \textcolor{gray}{$\pm$0.42}} &
        59.41 {\scriptsize \textcolor{gray}{$\pm$1.88}} &
        69.18 {\scriptsize \textcolor{gray}{$\pm$1.05}} & 
        62.87 \\
        & SUPT\textsubscript{soft} &
        61.93 {\scriptsize \textcolor{gray}{$\pm$1.34}} &
        70.40 {\scriptsize \textcolor{gray}{$\pm$0.15}} &
        57.91 {\scriptsize \textcolor{gray}{$\pm$0.39}} &
        56.41 {\scriptsize \textcolor{gray}{$\pm$0.57}} &
        59.50 {\scriptsize \textcolor{gray}{$\pm$2.13}} &
        64.42 {\scriptsize \textcolor{gray}{$\pm$1.13}} &
        59.31 {\scriptsize \textcolor{gray}{$\pm$1.55}} &
        68.75 {\scriptsize \textcolor{gray}{$\pm$1.30}} & 
        62.33 \\
        & SUPT\textsubscript{hard} &
        62.11 {\scriptsize \textcolor{gray}{$\pm$1.19}} &
        70.31 {\scriptsize \textcolor{gray}{$\pm$0.20}} &
        58.15 {\scriptsize \textcolor{gray}{$\pm$0.23}} &
        56.31 {\scriptsize \textcolor{gray}{$\pm$0.49}} &
        59.72 {\scriptsize \textcolor{gray}{$\pm$2.39}} &
        61.19 {\scriptsize \textcolor{gray}{$\pm$0.89}} &
        59.28 {\scriptsize \textcolor{gray}{$\pm$1.35}} &
        69.64 {\scriptsize \textcolor{gray}{$\pm$1.22}} & 
        62.09 \\
        & RELIEF &
        \underline{62.86} {\scriptsize \textcolor{gray}{$\pm$0.90}} &
        \underline{70.56} {\scriptsize \textcolor{gray}{$\pm$0.25}} &
        58.32 {\scriptsize \textcolor{gray}{$\pm$0.24}} &
        56.73 {\scriptsize \textcolor{gray}{$\pm$0.39}} &
        \textbf{62.66} {\scriptsize \textcolor{gray}{$\pm$1.39}} &
        64.37 {\scriptsize \textcolor{gray}{$\pm$0.80}} &
        \underline{59.64} {\scriptsize \textcolor{gray}{$\pm$1.92}} &
        \underline{70.07} {\scriptsize \textcolor{gray}{$\pm$0.90}} & 
        \underline{63.15} \\
        & LEAP &
        \textbf{62.98} {\scriptsize \textcolor{gray}{$\pm$0.81}} & \textbf{70.68} {\scriptsize \textcolor{gray}{$\pm$0.17}} & \textbf{58.53} {\scriptsize \textcolor{gray}{$\pm$0.35}} & \textbf{56.96} {\scriptsize \textcolor{gray}{$\pm$0.58}} & \underline{62.29} {\scriptsize \textcolor{gray}{$\pm$1.65}} & \textbf{64.92} {\scriptsize \textcolor{gray}{$\pm$0.61}} & \textbf{59.99} {\scriptsize \textcolor{gray}{$\pm$1.58}} & \textbf{70.33} {\scriptsize \textcolor{gray}{$\pm$0.84}} & \textbf{63.34}\\
    
        \addlinespace[1pt]
        \midrule[0.1pt]
        \addlinespace[2pt]
    
        \multirow{7}{*}{\rotatebox{90}{GCL}} &
        FT & 
        62.81 {\scriptsize \textcolor{gray}{$\pm$1.01}} &
        61.31 {\scriptsize \textcolor{gray}{$\pm$0.67}} &
        54.04 {\scriptsize \textcolor{gray}{$\pm$0.77}} &
        51.66 {\scriptsize \textcolor{gray}{$\pm$0.64}} &
        75.35 {\scriptsize \textcolor{gray}{$\pm$2.18}} &
        52.01 {\scriptsize \textcolor{gray}{$\pm$2.06}} &
        59.10 {\scriptsize \textcolor{gray}{$\pm$1.50}} &
        51.00 {\scriptsize \textcolor{gray}{$\pm$3.21}} & 
        58.41 \\
        & GPF &
        61.86 {\scriptsize \textcolor{gray}{$\pm$1.64}} &
        61.09 {\scriptsize \textcolor{gray}{$\pm$0.88}} &
        54.40 {\scriptsize \textcolor{gray}{$\pm$0.37}} &
        51.55 {\scriptsize \textcolor{gray}{$\pm$0.50}} &
        78.91 {\scriptsize \textcolor{gray}{$\pm$2.22}} &
        52.32 {\scriptsize \textcolor{gray}{$\pm$1.69}} &
        60.02 {\scriptsize \textcolor{gray}{$\pm$1.30}} &
        58.09 {\scriptsize \textcolor{gray}{$\pm$1.77}} & 
        59.78 \\
        & GPF-plus &
        62.99 {\scriptsize \textcolor{gray}{$\pm$1.63}} &
        61.28 {\scriptsize \textcolor{gray}{$\pm$0.49}} &
        54.55 {\scriptsize \textcolor{gray}{$\pm$0.32}} &
        51.95 {\scriptsize \textcolor{gray}{$\pm$0.95}} &
        79.16 {\scriptsize \textcolor{gray}{$\pm$1.91}} &
        52.70 {\scriptsize \textcolor{gray}{$\pm$1.95}} &
        60.21 {\scriptsize \textcolor{gray}{$\pm$1.45}} &
        60.61 {\scriptsize \textcolor{gray}{$\pm$2.56}} & 
        60.43 \\
        & SUPT\textsubscript{soft} &
        62.64 {\scriptsize \textcolor{gray}{$\pm$0.98}} &
        \textbf{61.65} {\scriptsize \textcolor{gray}{$\pm$0.20}} &
        54.50 {\scriptsize \textcolor{gray}{$\pm$0.56}} &
        51.49 {\scriptsize \textcolor{gray}{$\pm$1.32}} &
        78.28 {\scriptsize \textcolor{gray}{$\pm$2.50}} &
        52.52 {\scriptsize \textcolor{gray}{$\pm$1.72}} &
        59.75 {\scriptsize \textcolor{gray}{$\pm$1.79}} &
        59.55 {\scriptsize \textcolor{gray}{$\pm$2.39}} & 
        60.05 \\
        & SUPT\textsubscript{hard} &
        62.55 {\scriptsize \textcolor{gray}{$\pm$1.52}} &
        61.52 {\scriptsize \textcolor{gray}{$\pm$0.29}} &
        54.23 {\scriptsize \textcolor{gray}{$\pm$0.51}} &
        51.81 {\scriptsize \textcolor{gray}{$\pm$0.80}} &
        78.34 {\scriptsize \textcolor{gray}{$\pm$2.21}} &
        51.99 {\scriptsize \textcolor{gray}{$\pm$1.45}} &
        59.96 {\scriptsize \textcolor{gray}{$\pm$2.04}} &
        60.10 {\scriptsize \textcolor{gray}{$\pm$2.10}} & 
        59.95 \\
        & RELIEF &
        \textbf{63.73} {\scriptsize \textcolor{gray}{$\pm$0.74}} &
        61.49 {\scriptsize \textcolor{gray}{$\pm$0.68}} &
        \underline{54.73} {\scriptsize \textcolor{gray}{$\pm$0.30}} &
        \underline{52.20} {\scriptsize \textcolor{gray}{$\pm$0.70}} &
        \underline{79.53} {\scriptsize \textcolor{gray}{$\pm$1.85}} &
        \underline{53.55} {\scriptsize \textcolor{gray}{$\pm$1.80}} &
        \underline{60.40} {\scriptsize \textcolor{gray}{$\pm$2.12}} &
        \underline{62.47} {\scriptsize \textcolor{gray}{$\pm$1.30}} & 
        \underline{61.01} \\
        & LEAP &
        \underline{63.55} {\scriptsize \textcolor{gray}{$\pm$1.08}} & \underline{61.54} {\scriptsize \textcolor{gray}{$\pm$0.39}} & \textbf{54.89} {\scriptsize \textcolor{gray}{$\pm$0.35}} & \textbf{52.33} {\scriptsize \textcolor{gray}{$\pm$1.01}} & \textbf{79.93} {\scriptsize \textcolor{gray}{$\pm$1.96}} & \textbf{53.59} {\scriptsize \textcolor{gray}{$\pm$1.91}}& \textbf{60.53} {\scriptsize \textcolor{gray}{$\pm$1.98}} &  \textbf{62.50} {\scriptsize \textcolor{gray}{$\pm$1.41}} & \textbf{61.11}\\
        
        \bottomrule
      \end{tabular}
  \label{tab:graph_few}
\end{table*}

\section{Additional Experimental Results}
\label{appendix:additional_experiment_results}

\subsection{Few-shot Scenario}
\label{appendix:few-shot}
We further conduct experiments in the few-shot scenario to validate the effectiveness of our LEAP. Table~\ref{tab:graph_few} and Table~\ref{tab:node_few} show that our LEAP achieves superior performance compared to baselines on graph- and node-level tasks in the few-shot scenario, with improvements in $27/32$ and $17/20$ tasks (metrics), respectively. Notably, RELIEF performs well in the few-shot scenario, but its performance in the full-shot scenario is unsatisfactory, which is consistent with the conclusion we derived in Theorem~\ref{theorem:1}. Its selective node-based graph prompt tuning facilitates prompt learning but undermines the theoretical foundation of universal graph prompt tuning.

\begin{table*}[htbp]
    \caption{Accuracy (\%) and Macro F1-score (\%) with respective standard deviation for node classification under 10-shot scenario with two pre-training and various tuning strategies.}
    \small
        \centering
        \begin{tabular}{cc<{\hspace{3pt}}|>{\hspace{3pt}}c@{\hspace{3pt}}c|c@{\hspace{3pt}}c|c@{\hspace{3pt}}c|c@{\hspace{3pt}}c|c@{\hspace{3pt}}c}

        \toprule
        
        & \multirow{2}{*}{\makecell{Tuning \\ Strategy}} & \multicolumn{2}{c|}{Cora} & \multicolumn{2}{c|}{CiteSeer} & \multicolumn{2}{c|}{PubMed} & \multicolumn{2}{c|}{Computers} & \multicolumn{2}{c}{Photos} \\
        & & Accuracy & Macro F1 & Accuracy & Macro F1 & Accuracy & Macro F1 & Accuracy & Macro F1 & Accuracy & Macro F1 \\
    
        \midrule

        \multirow{8}[2]{*}{\rotatebox{90}{MaskedEdge}} &
        FT & 
        54.21 {\scriptsize \textcolor{gray}{$\pm$1.80}} &
        53.30 {\scriptsize \textcolor{gray}{$\pm$1.65}} &
        60.83 {\scriptsize \textcolor{gray}{$\pm$1.35}} &
        56.25 {\scriptsize \textcolor{gray}{$\pm$1.23}} &
        70.91 {\scriptsize \textcolor{gray}{$\pm$1.89}} &
        65.42 {\scriptsize \textcolor{gray}{$\pm$1.73}} &
        74.61 {\scriptsize \textcolor{gray}{$\pm$1.80}} &
        70.52 {\scriptsize \textcolor{gray}{$\pm$1.73}} &
        83.52 {\scriptsize \textcolor{gray}{$\pm$2.02}} &
        81.19 {\scriptsize \textcolor{gray}{$\pm$2.13}} \\
        
        & GPPT &
        47.62 {\scriptsize \textcolor{gray}{$\pm$0.74}} &
        46.99 {\scriptsize \textcolor{gray}{$\pm$0.60}} &
        57.33 {\scriptsize \textcolor{gray}{$\pm$0.82}} &
        51.49 {\scriptsize \textcolor{gray}{$\pm$0.86}} &
        70.11 {\scriptsize \textcolor{gray}{$\pm$0.92}} &
        65.05 {\scriptsize \textcolor{gray}{$\pm$0.88}} &
        73.80 {\scriptsize \textcolor{gray}{$\pm$0.81}} &
        68.64 {\scriptsize \textcolor{gray}{$\pm$0.77}} &
        82.05 {\scriptsize \textcolor{gray}{$\pm$0.94}} &
        80.41 {\scriptsize \textcolor{gray}{$\pm$1.11}} \\
        
        & GPF &
        55.35 {\scriptsize \textcolor{gray}{$\pm$1.52}} &
        54.28 {\scriptsize \textcolor{gray}{$\pm$1.31}} &
        60.32 {\scriptsize \textcolor{gray}{$\pm$1.17}} &
        55.55 {\scriptsize \textcolor{gray}{$\pm$1.30}} &
        69.81 {\scriptsize \textcolor{gray}{$\pm$2.90}} &
        64.97 {\scriptsize \textcolor{gray}{$\pm$2.41}} &
        73.29 {\scriptsize \textcolor{gray}{$\pm$2.06}} &
        68.03 {\scriptsize \textcolor{gray}{$\pm$1.95}} &
        83.08 {\scriptsize \textcolor{gray}{$\pm$0.92}} &
        80.93 {\scriptsize \textcolor{gray}{$\pm$0.77}} \\
        
        & GPF-plus &
        56.72 {\scriptsize \textcolor{gray}{$\pm$1.26}} &
        55.03 {\scriptsize \textcolor{gray}{$\pm$1.03}} &
        \underline{63.36} {\scriptsize \textcolor{gray}{$\pm$0.91}} &
        56.92 {\scriptsize \textcolor{gray}{$\pm$0.75}} &
        \textbf{72.08} {\scriptsize \textcolor{gray}{$\pm$2.31}} &
        66.18 {\scriptsize \textcolor{gray}{$\pm$2.01}} &
        \underline{75.22} {\scriptsize \textcolor{gray}{$\pm$1.27}} &
        \textbf{72.02} {\scriptsize \textcolor{gray}{$\pm$1.35}} &
        \underline{84.82} {\scriptsize \textcolor{gray}{$\pm$1.30}} &
        82.61 {\scriptsize \textcolor{gray}{$\pm$1.04}} \\
        
        & SUPT\textsubscript{soft} &
        55.31 {\scriptsize \textcolor{gray}{$\pm$1.71}} &
        54.29 {\scriptsize \textcolor{gray}{$\pm$1.66}} &
        60.02 {\scriptsize \textcolor{gray}{$\pm$1.99}} &
        54.99 {\scriptsize \textcolor{gray}{$\pm$1.58}} &
        71.18 {\scriptsize \textcolor{gray}{$\pm$2.15}} &
        65.90 {\scriptsize \textcolor{gray}{$\pm$2.03}} &
        72.28 {\scriptsize \textcolor{gray}{$\pm$2.91}} &
        67.91 {\scriptsize \textcolor{gray}{$\pm$2.66}} &
        84.00 {\scriptsize \textcolor{gray}{$\pm$1.25}} &
        82.07 {\scriptsize \textcolor{gray}{$\pm$1.69}} \\
        
        & SUPT\textsubscript{hard} &
        56.31 {\scriptsize \textcolor{gray}{$\pm$1.08}} &
        \underline{55.45} {\scriptsize \textcolor{gray}{$\pm$1.37}} &
        61.88 {\scriptsize \textcolor{gray}{$\pm$1.43}} &
        56.73 {\scriptsize \textcolor{gray}{$\pm$1.25}} &
        70.59 {\scriptsize \textcolor{gray}{$\pm$1.86}} &
        65.29 {\scriptsize \textcolor{gray}{$\pm$1.99}} &
        72.56 {\scriptsize \textcolor{gray}{$\pm$1.31}} &
        68.20 {\scriptsize \textcolor{gray}{$\pm$1.28}} &
        84.33 {\scriptsize \textcolor{gray}{$\pm$0.88}} &
        82.45 {\scriptsize \textcolor{gray}{$\pm$0.72}} \\
        
        & RELIEF &
        \underline{57.03} {\scriptsize \textcolor{gray}{$\pm$1.21}} &
        55.39 {\scriptsize \textcolor{gray}{$\pm$0.99}} &
        63.33 {\scriptsize \textcolor{gray}{$\pm$1.40}} &
        \underline{57.51} {\scriptsize \textcolor{gray}{$\pm$1.05}} &
        71.16 {\scriptsize \textcolor{gray}{$\pm$1.68}} &
        \underline{66.37} {\scriptsize \textcolor{gray}{$\pm$1.21}} &
        73.02 {\scriptsize \textcolor{gray}{$\pm$1.33}} &
        69.97 {\scriptsize \textcolor{gray}{$\pm$1.60}} &
        84.73 {\scriptsize \textcolor{gray}{$\pm$1.29}} &
        \underline{83.24} {\scriptsize \textcolor{gray}{$\pm$0.88}} \\
        
        & \model  & \textbf{58.32} {\scriptsize \textcolor{gray}{$\pm$1.50}} & \textbf{56.81} {\scriptsize \textcolor{gray}{$\pm$0.94}} & \textbf{64.24} {\scriptsize \textcolor{gray}{$\pm$1.56}} & \textbf{57.67} {\scriptsize \textcolor{gray}{$\pm$1.54}} & \underline{71.68} {\scriptsize \textcolor{gray}{$\pm$1.48}} & \textbf{66.65} {\scriptsize \textcolor{gray}{$\pm$1.61}} & \textbf{75.48} {\scriptsize \textcolor{gray}{$\pm$1.33}} & \underline{71.09} {\scriptsize \textcolor{gray}{$\pm$1.09}} & \textbf{85.19} {\scriptsize \textcolor{gray}{$\pm$1.50}} & \textbf{83.35} {\scriptsize \textcolor{gray}{$\pm$1.42}} \\
        
        \addlinespace[1pt]
        \midrule[0.1pt]
        \addlinespace[2pt]
    
        \multirow{8}[2]{*}{\rotatebox{90}{ContraEdge}} &
        FT & 
        60.28 {\scriptsize \textcolor{gray}{$\pm$1.99}} &
        60.09 {\scriptsize \textcolor{gray}{$\pm$1.82}} &
        62.61 {\scriptsize \textcolor{gray}{$\pm$2.09}} &
        56.48 {\scriptsize \textcolor{gray}{$\pm$2.15}} &
        70.15 {\scriptsize \textcolor{gray}{$\pm$1.44}} &
        65.84 {\scriptsize \textcolor{gray}{$\pm$1.51}} &
        80.02 {\scriptsize \textcolor{gray}{$\pm$1.03}} &
        76.25 {\scriptsize \textcolor{gray}{$\pm$1.21}} &
        85.02 {\scriptsize \textcolor{gray}{$\pm$1.45}} &
        81.97 {\scriptsize \textcolor{gray}{$\pm$1.58}} \\
        
        & GPrompt &
        56.98 {\scriptsize \textcolor{gray}{$\pm$2.53}} &
        54.55 {\scriptsize \textcolor{gray}{$\pm$2.48}} &
        59.30 {\scriptsize \textcolor{gray}{$\pm$0.81}} &
        55.88 {\scriptsize \textcolor{gray}{$\pm$0.94}} &
        69.50 {\scriptsize \textcolor{gray}{$\pm$0.84}} &
        64.73 {\scriptsize \textcolor{gray}{$\pm$0.77}} &
        69.81 {\scriptsize \textcolor{gray}{$\pm$1.92}} &
        67.13 {\scriptsize \textcolor{gray}{$\pm$1.90}} &
        76.30 {\scriptsize \textcolor{gray}{$\pm$2.23}} &
        73.58 {\scriptsize \textcolor{gray}{$\pm$2.17}} \\
        
        & GPF &
        59.31 {\scriptsize \textcolor{gray}{$\pm$1.42}} &
        58.84 {\scriptsize \textcolor{gray}{$\pm$1.58}} &
        63.37 {\scriptsize \textcolor{gray}{$\pm$2.01}} &
        55.79 {\scriptsize \textcolor{gray}{$\pm$2.04}} &
        70.91 {\scriptsize \textcolor{gray}{$\pm$2.21}} &
        65.99 {\scriptsize \textcolor{gray}{$\pm$1.88}} &
        80.20 {\scriptsize \textcolor{gray}{$\pm$1.18}} &
        76.91 {\scriptsize \textcolor{gray}{$\pm$1.29}} &
        85.39 {\scriptsize \textcolor{gray}{$\pm$1.55}} &
        82.64 {\scriptsize \textcolor{gray}{$\pm$1.75}} \\
        
        & GPF-plus &
        \underline{60.93} {\scriptsize \textcolor{gray}{$\pm$1.04}} &
        60.40 {\scriptsize \textcolor{gray}{$\pm$1.10}} &
        65.03 {\scriptsize \textcolor{gray}{$\pm$1.89}} &
        57.14 {\scriptsize \textcolor{gray}{$\pm$1.72}} &
        71.18 {\scriptsize \textcolor{gray}{$\pm$3.73}} &
        66.39 {\scriptsize \textcolor{gray}{$\pm$3.67}} &
        80.85 {\scriptsize \textcolor{gray}{$\pm$1.03}} &
        77.05 {\scriptsize \textcolor{gray}{$\pm$1.12}} &
        86.12 {\scriptsize \textcolor{gray}{$\pm$1.30}} &
        83.43 {\scriptsize \textcolor{gray}{$\pm$1.61}} \\
        
        & SUPT\textsubscript{soft} &
        60.03 {\scriptsize \textcolor{gray}{$\pm$1.44}} &
        59.61 {\scriptsize \textcolor{gray}{$\pm$1.20}} &
        63.69 {\scriptsize \textcolor{gray}{$\pm$2.01}} &
        56.57 {\scriptsize \textcolor{gray}{$\pm$2.19}} &
        70.99 {\scriptsize \textcolor{gray}{$\pm$3.15}} &
        66.08 {\scriptsize \textcolor{gray}{$\pm$3.03}} &
        \underline{81.03} {\scriptsize \textcolor{gray}{$\pm$1.15}} &
        76.31 {\scriptsize \textcolor{gray}{$\pm$0.94}} &
        85.15 {\scriptsize \textcolor{gray}{$\pm$1.54}} &
        82.33 {\scriptsize \textcolor{gray}{$\pm$1.09}} \\
        
        & SUPT\textsubscript{hard} &
        59.37 {\scriptsize \textcolor{gray}{$\pm$0.96}} &
        59.09 {\scriptsize \textcolor{gray}{$\pm$0.89}} &
        64.50 {\scriptsize \textcolor{gray}{$\pm$1.89}} &
        56.26 {\scriptsize \textcolor{gray}{$\pm$1.94}} &
        70.86 {\scriptsize \textcolor{gray}{$\pm$2.27}} &
        66.23 {\scriptsize \textcolor{gray}{$\pm$2.30}} &
        80.20 {\scriptsize \textcolor{gray}{$\pm$1.04}} &
        76.43 {\scriptsize \textcolor{gray}{$\pm$1.27}} &
        84.68 {\scriptsize \textcolor{gray}{$\pm$1.35}} &
        82.12 {\scriptsize \textcolor{gray}{$\pm$1.10}} \\
      
        & RELIEF &
        \underline{60.93} {\scriptsize \textcolor{gray}{$\pm$1.40}} &
        \underline{60.77} {\scriptsize \textcolor{gray}{$\pm$1.08}} &
        \textbf{66.11} {\scriptsize \textcolor{gray}{$\pm$1.01}} &
        \underline{57.71} {\scriptsize \textcolor{gray}{$\pm$0.92}} &
        \underline{71.35} {\scriptsize \textcolor{gray}{$\pm$1.58}} &
        \underline{66.82} {\scriptsize \textcolor{gray}{$\pm$1.49}} &
        80.98 {\scriptsize \textcolor{gray}{$\pm$1.33}} &
        \underline{77.85} {\scriptsize \textcolor{gray}{$\pm$1.51}} &
        \underline{86.26} {\scriptsize \textcolor{gray}{$\pm$1.10}} &
        \underline{83.61} {\scriptsize \textcolor{gray}{$\pm$0.87}} \\

        & \model & \textbf{61.33} {\scriptsize \textcolor{gray}{$\pm$0.92}} & \textbf{61.12} {\scriptsize \textcolor{gray}{$\pm$1.14}} & \underline{65.92} {\scriptsize \textcolor{gray}{$\pm$1.01}} & \textbf{57.90} {\scriptsize \textcolor{gray}{$\pm$0.85}} & \textbf{71.84} {\scriptsize \textcolor{gray}{$\pm$1.14}} & \textbf{66.95} {\scriptsize \textcolor{gray}{$\pm$1.21}} & \textbf{81.32} {\scriptsize \textcolor{gray}{$\pm$1.55}} & \textbf{78.21} {\scriptsize \textcolor{gray}{$\pm$1.60}} & \textbf{86.41} {\scriptsize \textcolor{gray}{$\pm$1.39}} &\textbf{83.95} {\scriptsize \textcolor{gray}{$\pm$1.32}} \\

        \bottomrule
        \end{tabular}
    \label{tab:node_few}
\end{table*}

\subsection{Ablation Study}
\label{appendix:ablation_study}
To analyze the effectiveness of LEAP, we conduct extensive ablation studies to evaluate the necessity and contribution of each individual component within the model. Specifically, we compare our model with the following variants: 1) LEAP\textsubscript{GPF}, which replaces the basic universal graph prompt with GPF. 2) LEAP\textsubscript{GPF-plus}, which replaces the basic universal graph prompt with full GPF-plus. 3) LEAP $w/o$ ECR, which removes the editing coverage rate from the reward function. 
The results are demonstrated in Table~\ref{tab:ablation_graph} and Table~\ref{tab:ablation_node}. Specifically, we have the following observations. LEAP performs superior in most scenarios. We attribute this to learning different prompts for each node being inherently more challenging than learning $k$ basic universal graph prompt vectors and $k$ linear projections. On the other hand, LEAP $w/o$ ECR demonstrates the least satisfactory performance in the full-shot scenario. We ascribe this to the absence of the editing convergence rate constraint, which leads the model to repeatedly select and modify a small subset of nodes.

\subsection{Hyper-parameter Analysis}
\label{appendix:hyper-parameter_analysis}
We analyze the critical hyper-parameters in LEAP, including the basic universal graph prompt vector number $k$, the prompt edit range $\vartheta$, the policy networks update interval $h$, and the policy finite horizon (step) $T$. For both graph- and node-level tasks, we report the average performance across all datasets for each pre-training strategy.

\subsubsection{Basic Universal Graph Prompt Vector Number $k$}
\begin{figure*}[htbp]
    \centering
    \vspace{-0.5em}
    \subfigure {
        \includegraphics[width=0.23\linewidth]{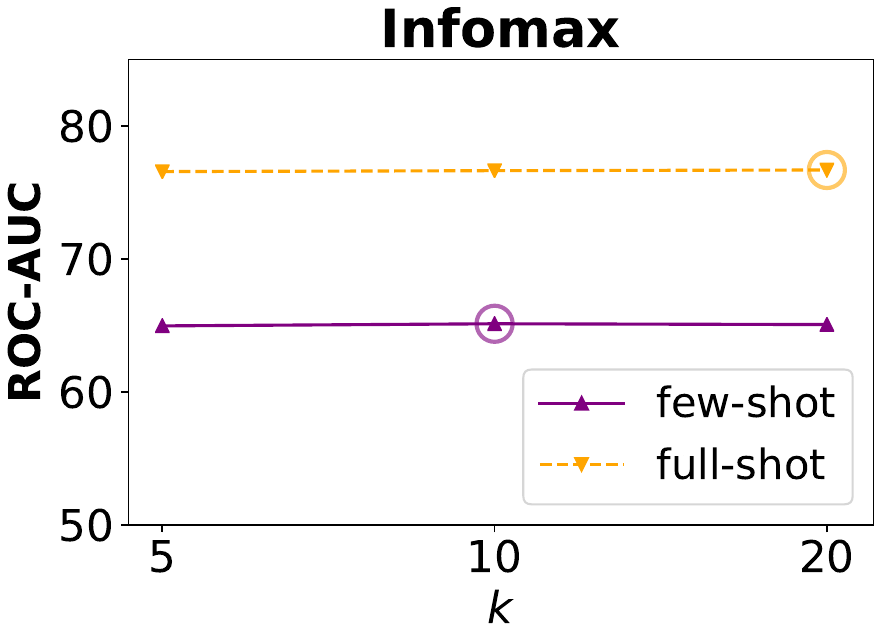}
        } 
    \subfigure {
        \includegraphics[width=0.23\linewidth]{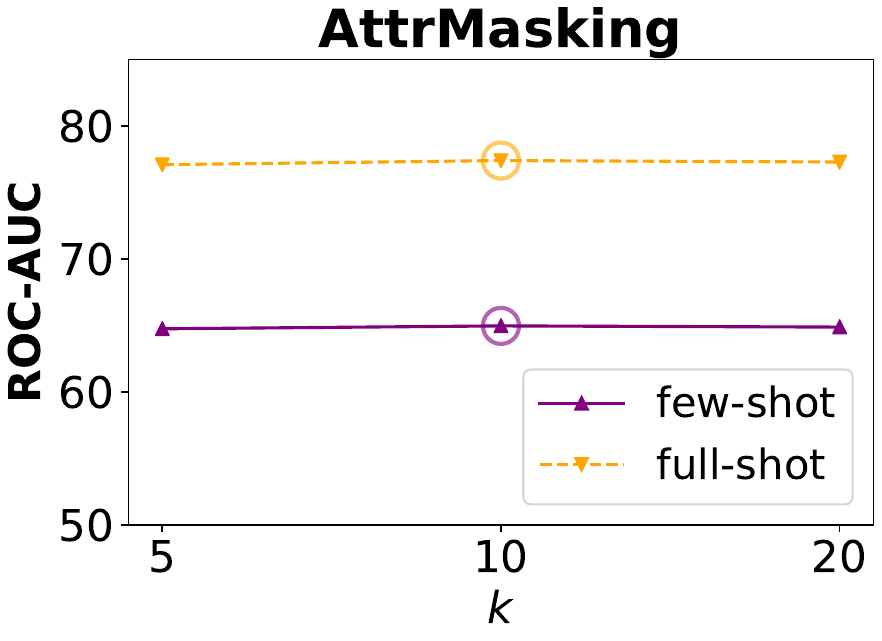}
        }
    \subfigure {
        \includegraphics[width=0.23\linewidth]{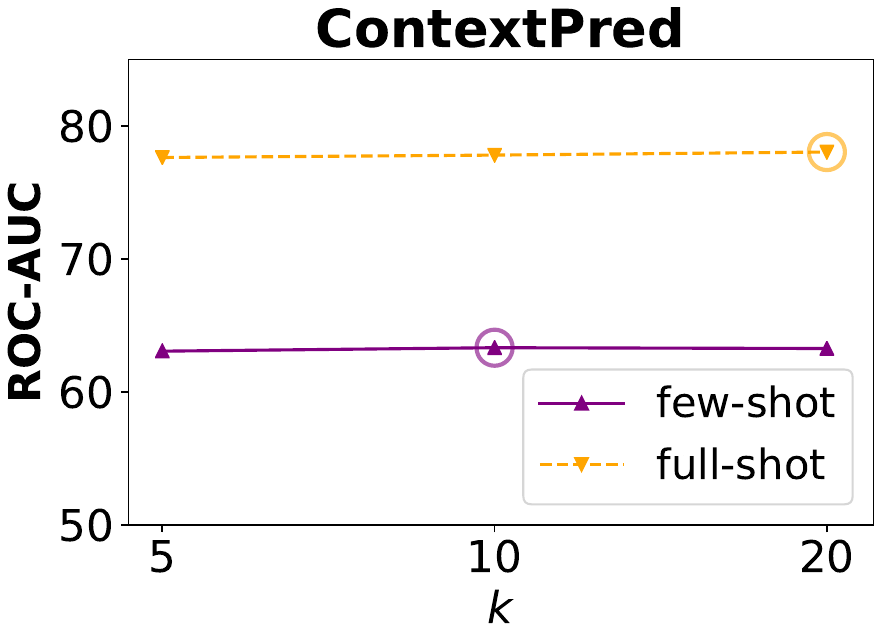}
        }
    \subfigure {
        \includegraphics[width=0.23\linewidth]{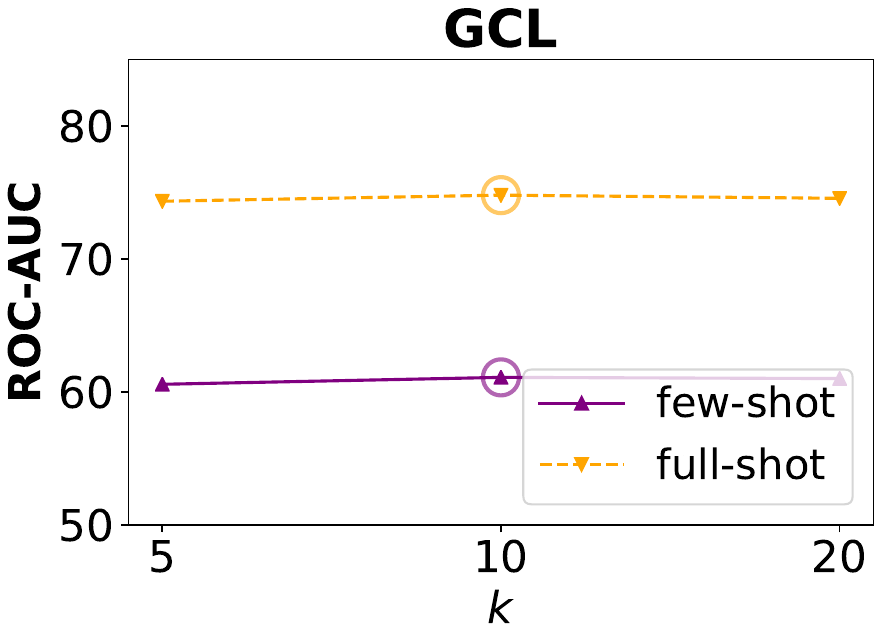}
        } 
    \vskip -0.1in
    \caption{The average ROC-AUC across all datasets for the graph-level task under different hyper-parameter $k$, with the circle marking the optimal results.}
    \label{fig:k-graph}
     \vskip -0.1in
\end{figure*}

\begin{figure*}[htbp]
    \centering
    \vspace{-0.5em}
    \subfigure {
        \includegraphics[width=0.23\linewidth]{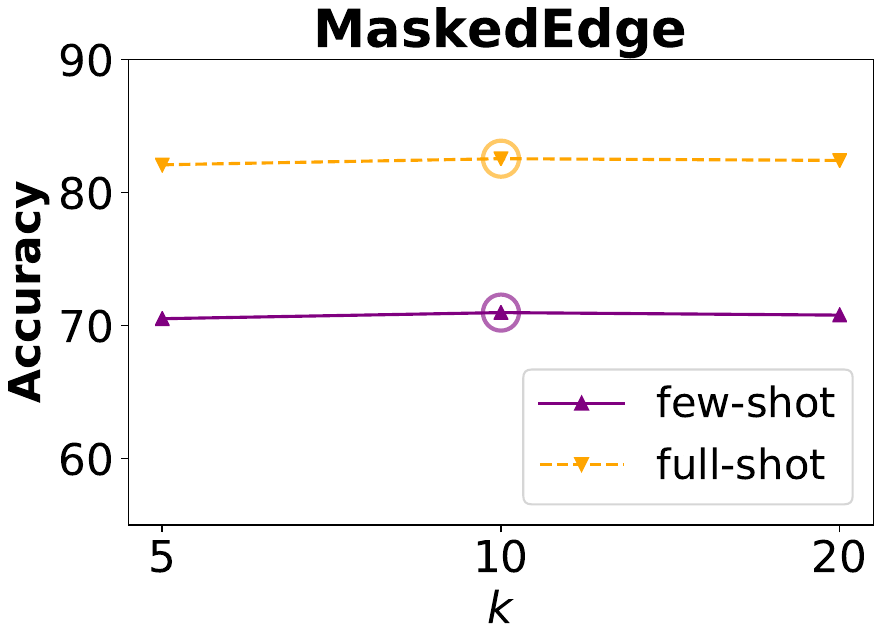}
        } 
    \subfigure {
        \includegraphics[width=0.23\linewidth]{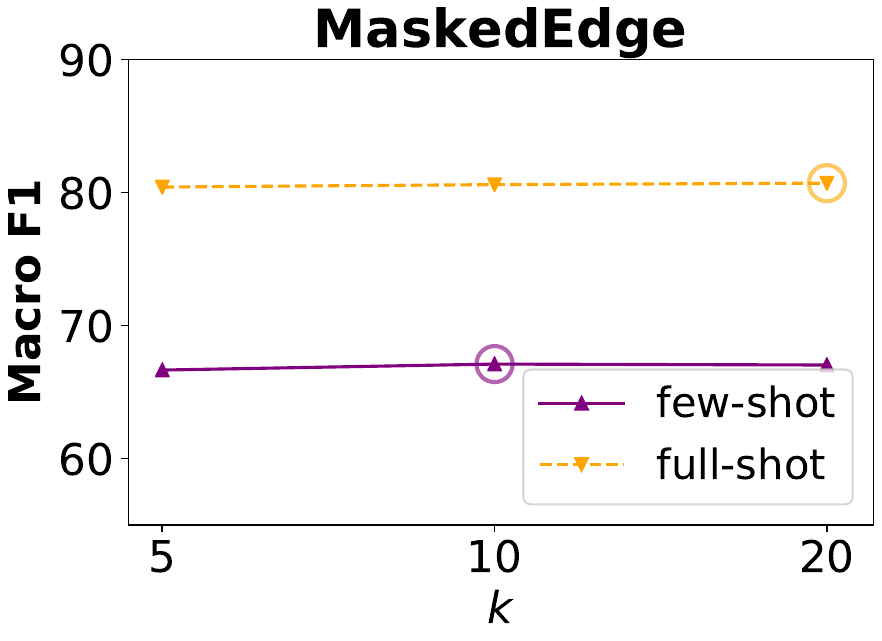}
        }
    \subfigure {
        \includegraphics[width=0.23\linewidth]{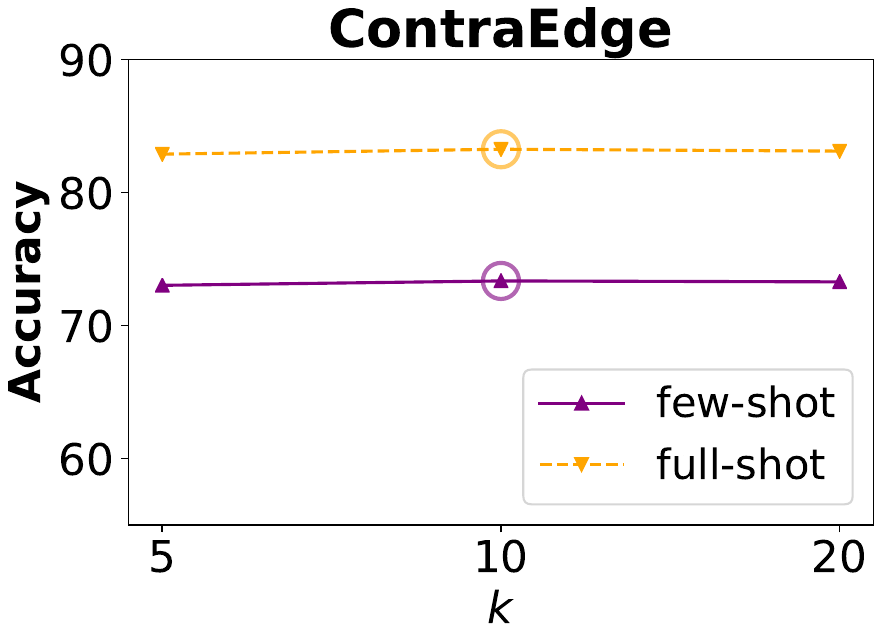}
        }
    \subfigure {
        \includegraphics[width=0.23\linewidth]{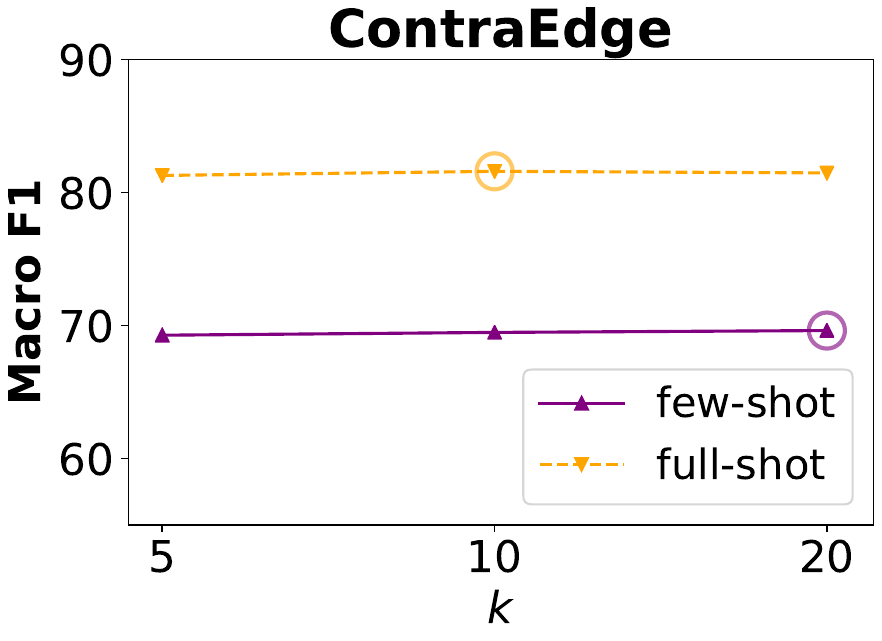}
        } 
    \vskip -0.1in
    \caption{The average Accuracy and Macro F1 across all datasets for the node-level task under different hyper-parameter $k$, with the circle marking the optimal results.}
    \label{fig:k-node}
     \vskip -0.1in
\end{figure*}

From Figure~\ref{fig:k-graph} and Figure~\ref{fig:k-node}, we observe that the optimal value of the number of basic universal graph prompt vectors $k$ is $10$. Moreover, selecting $k$ as $5$ or $20$ has no significant impact on performance, demonstrating LEAP's robustness to hyper-parameters $k$.

\subsubsection{Prompt Edit Range $\vartheta$}
\begin{figure*}[htbp]
    \centering
    \vspace{-0.5em}
    \subfigure {
        \includegraphics[width=0.23\linewidth]{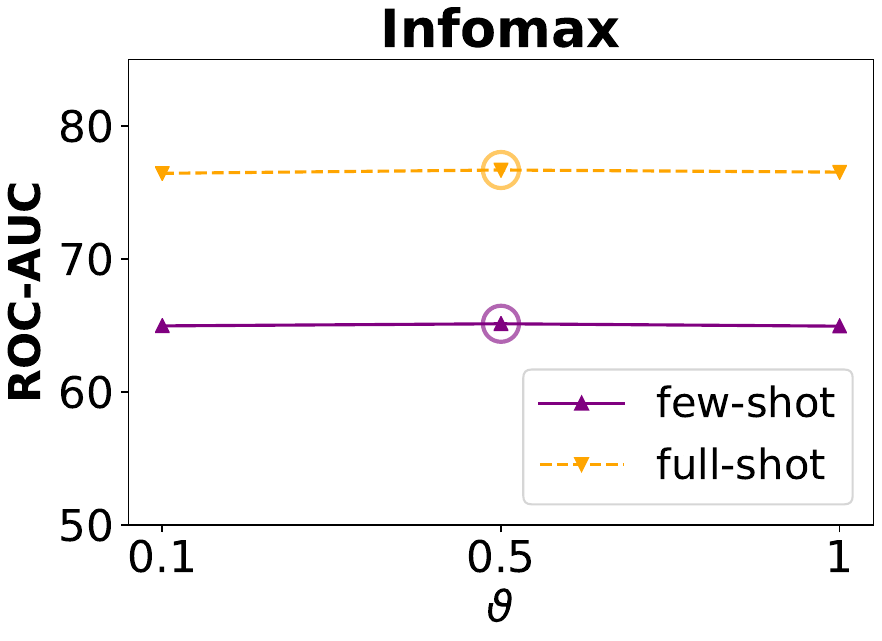}
        } 
    \subfigure {
        \includegraphics[width=0.23\linewidth]{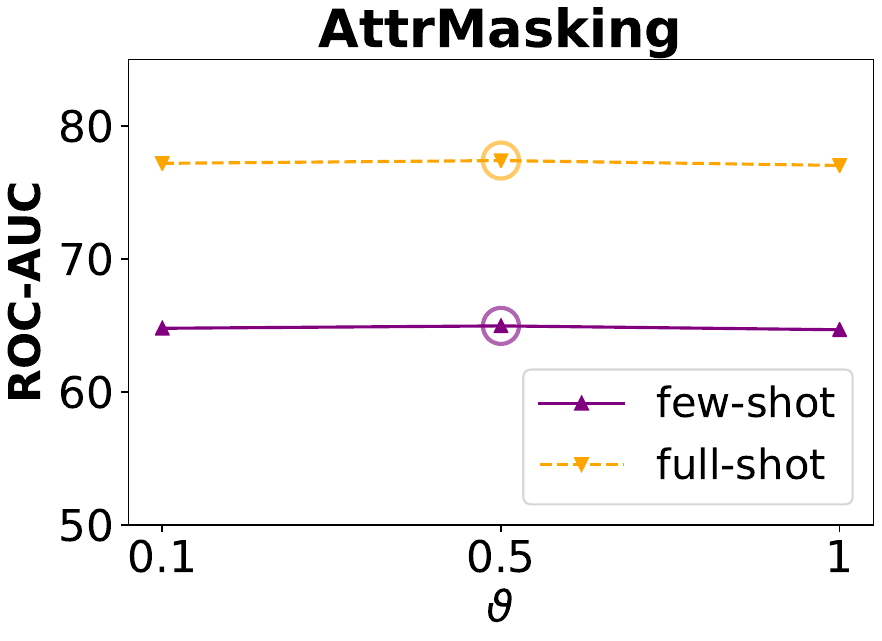}
        }
    \subfigure {
        \includegraphics[width=0.23\linewidth]{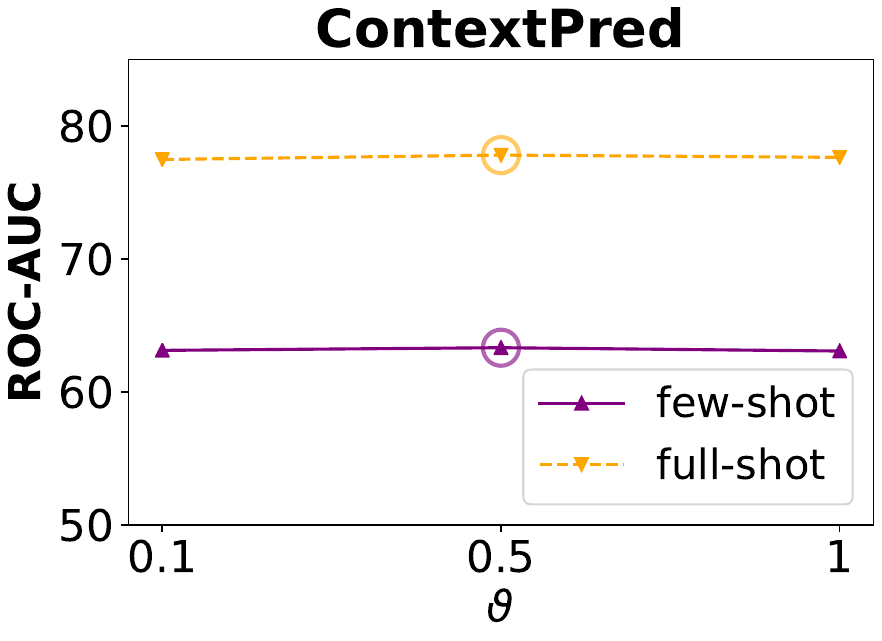}
        }
    \subfigure {
        \includegraphics[width=0.23\linewidth]{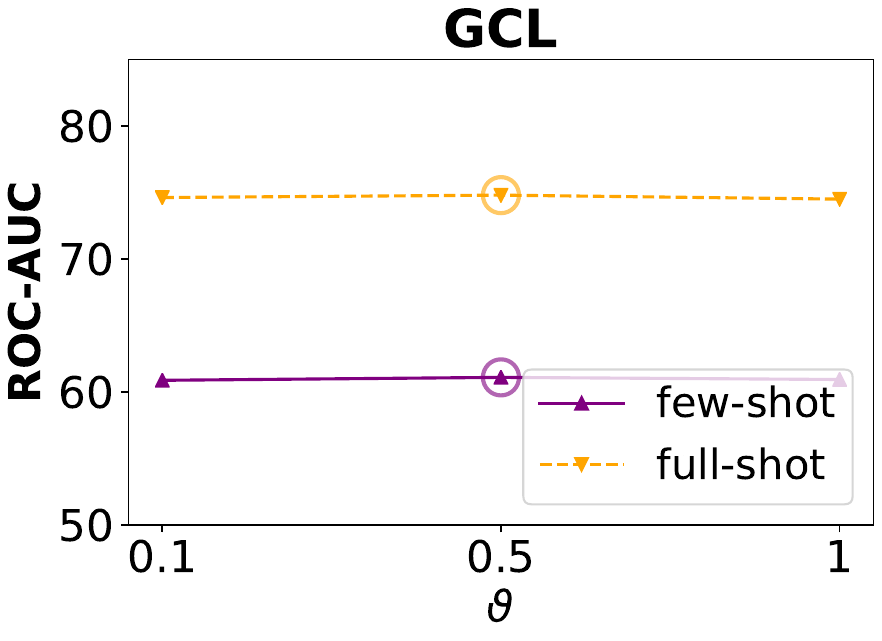}
        } 
    \vskip -0.1in
    \caption{The average ROC-AUC across all datasets for the graph-level task under different hyper-parameter $\vartheta$, with the circle marking the optimal results.}
    \label{fig:range-graph}
     \vskip -0.1in
\end{figure*}

\begin{figure*}[!htbp]
    \centering
    \vspace{-0.5em}
    \subfigure {
        \includegraphics[width=0.23\linewidth]{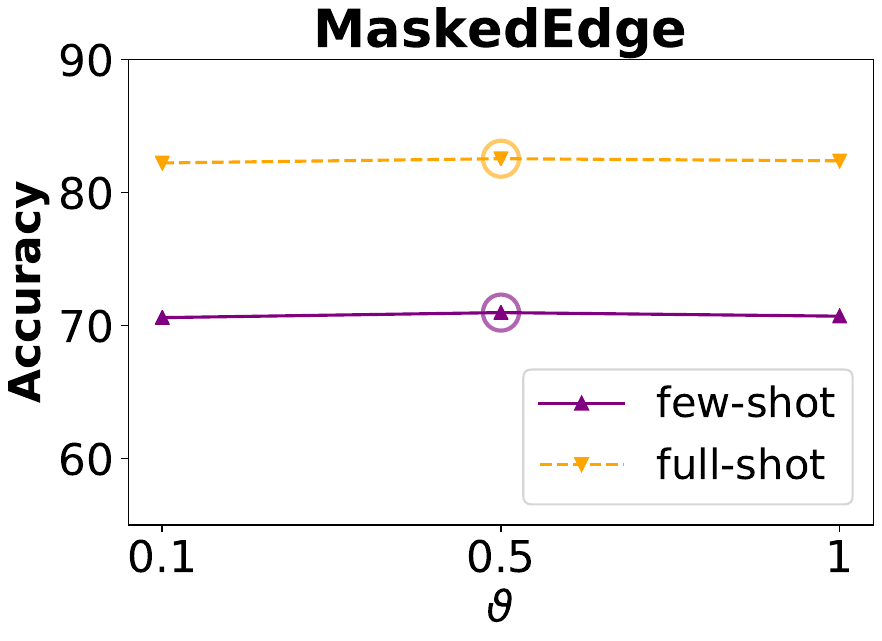}
        } 
    \subfigure {
        \includegraphics[width=0.23\linewidth]{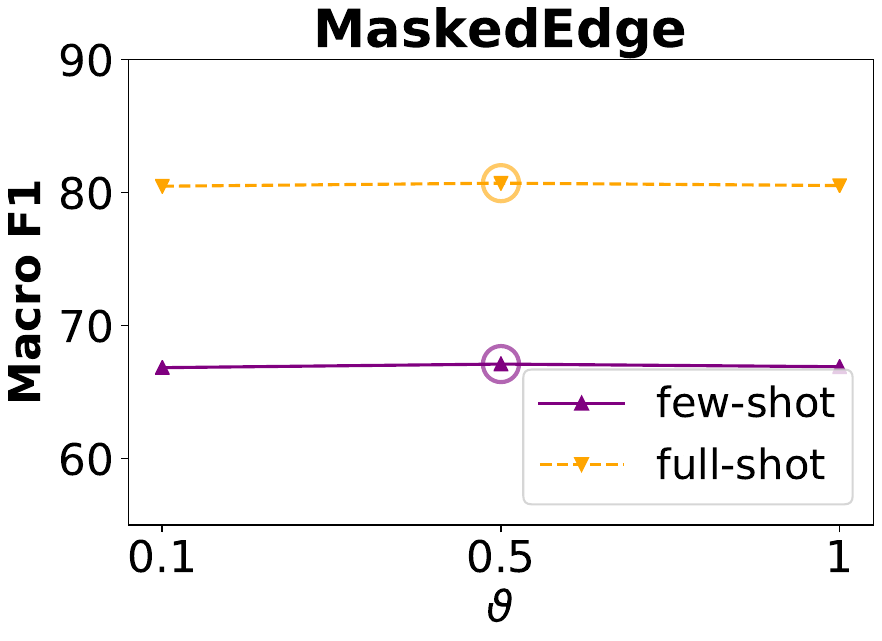}
        }
    \subfigure {
        \includegraphics[width=0.23\linewidth]{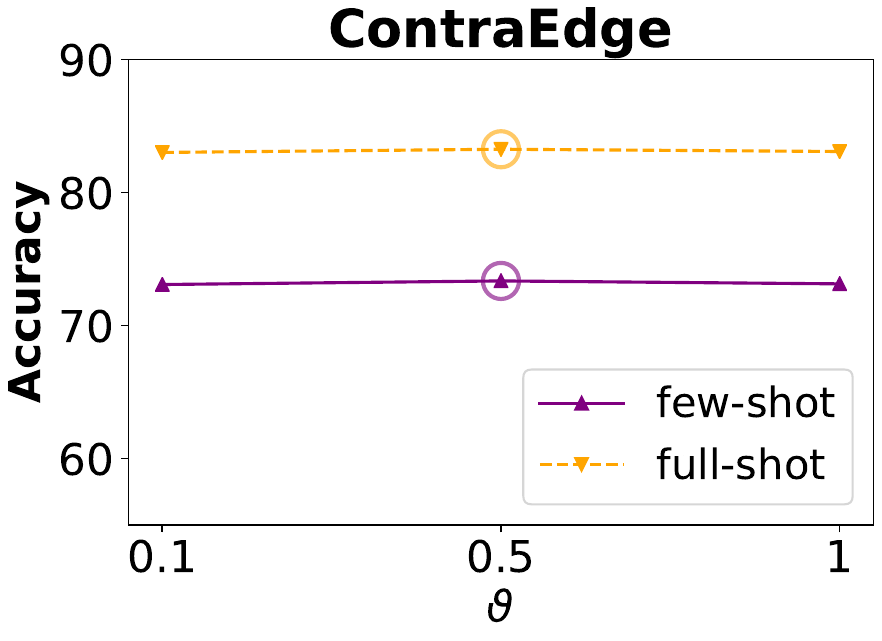}
        }
    \subfigure {
        \includegraphics[width=0.23\linewidth]{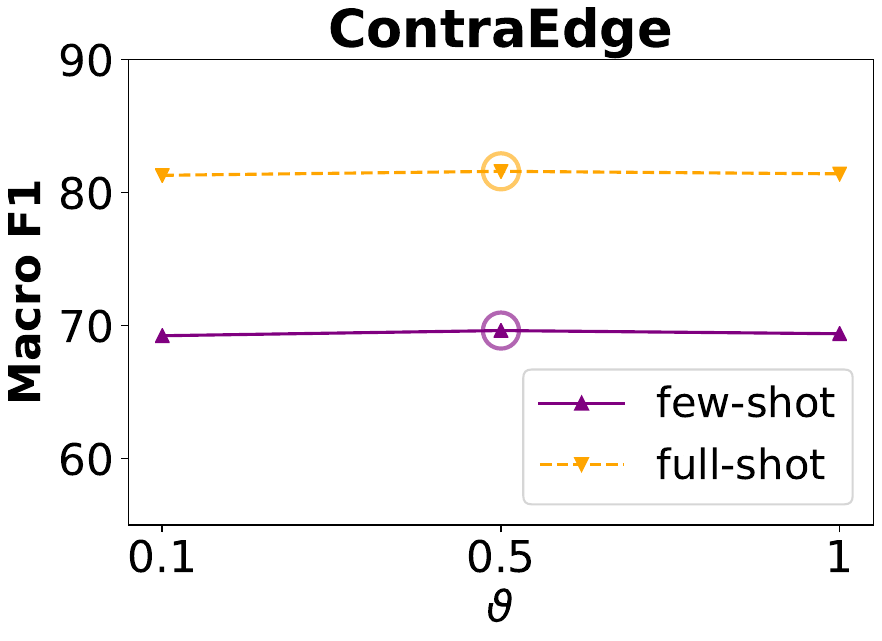}
        } 
    \vskip -0.1in
    \caption{The average Accuracy and Macro F1 across all datasets for the node-level task under different hyper-parameter $\vartheta$, with the circle marking the optimal results.}
    \label{fig:range-node}
     \vskip -0.1in
\end{figure*}

From Figure~\ref{fig:range-graph} and Figure~\ref{fig:range-node}, we observe that the optimal value of prompt edit range $\vartheta$ is $0.5$. Moreover, varying $\vartheta$ within reasonable limits has minimal impact on the model's performance, highlighting LEAP's robustness to the hyper-parameter $\vartheta$.

\subsubsection{Policy Networks Update Interval $h$}
\begin{figure*}[htbp]
    \centering
    \vspace{-0.5em}
    \subfigure {
        \includegraphics[width=0.23\linewidth]{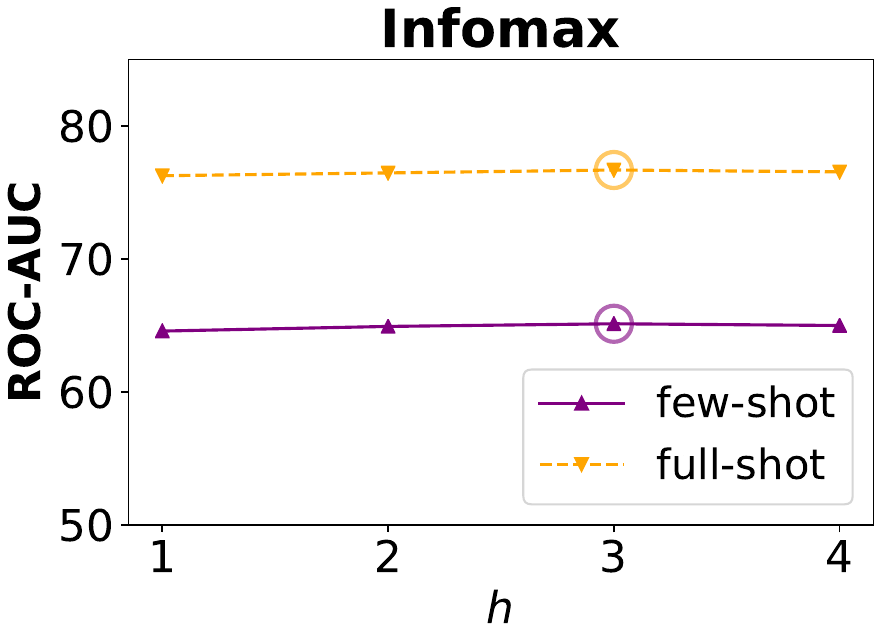}
        } 
    \subfigure {
        \includegraphics[width=0.23\linewidth]{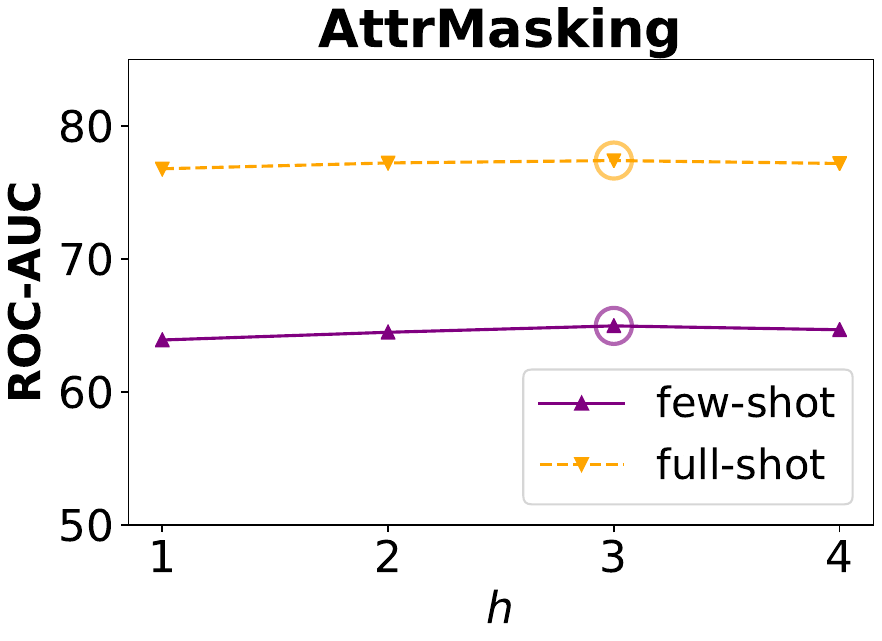}
        }
    \subfigure {
        \includegraphics[width=0.23\linewidth]{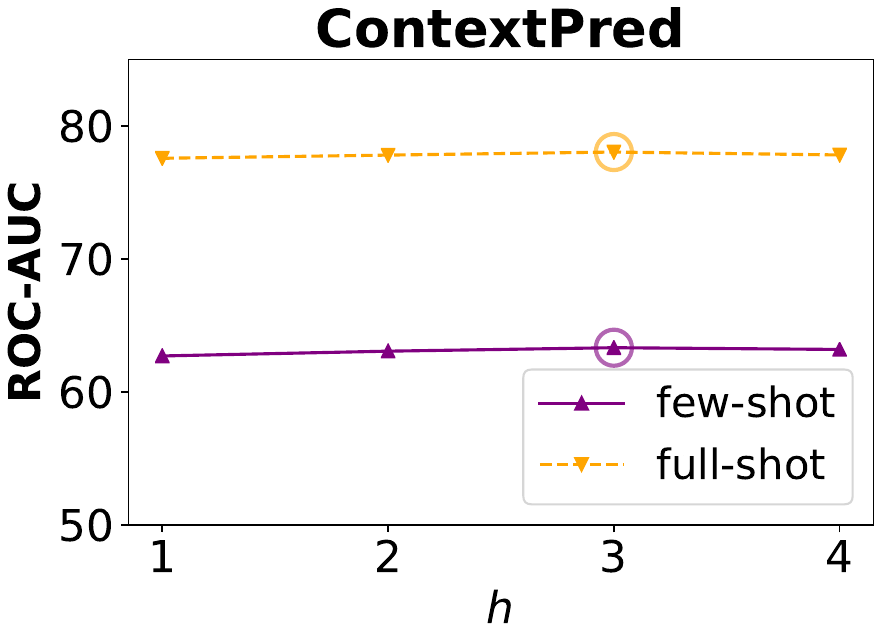}
        }
    \subfigure {
        \includegraphics[width=0.23\linewidth]{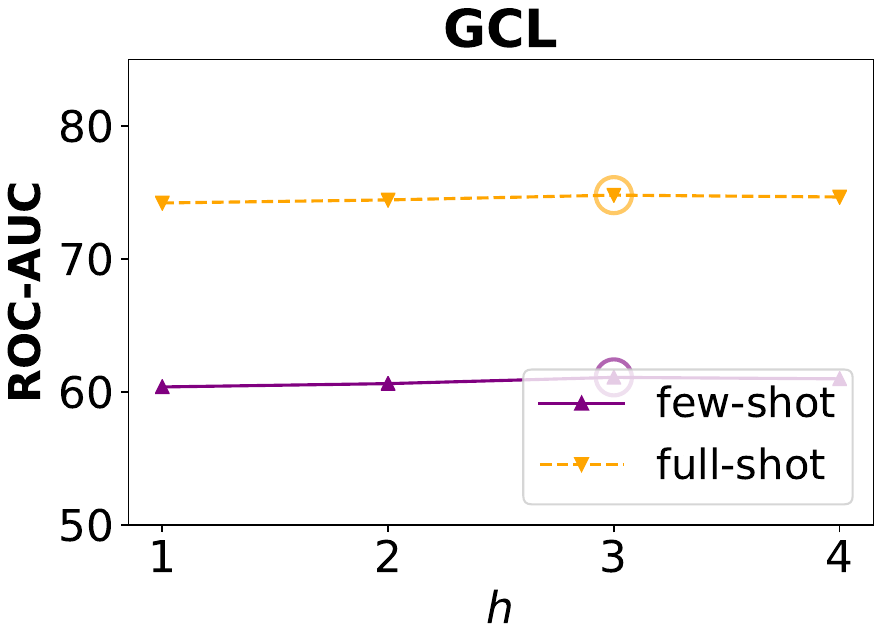}
        } 
    \vskip -0.1in
    \caption{The average ROC-AUC across all datasets for the graph-level task under different hyper-parameter $h$, with the circle marking the optimal results.}
    \label{fig:h-graph}
     \vskip -0.1in
\end{figure*}

\begin{figure*}[!htbp]
    \centering
    \vspace{-0.5em}
    \subfigure {
        \includegraphics[width=0.23\linewidth]{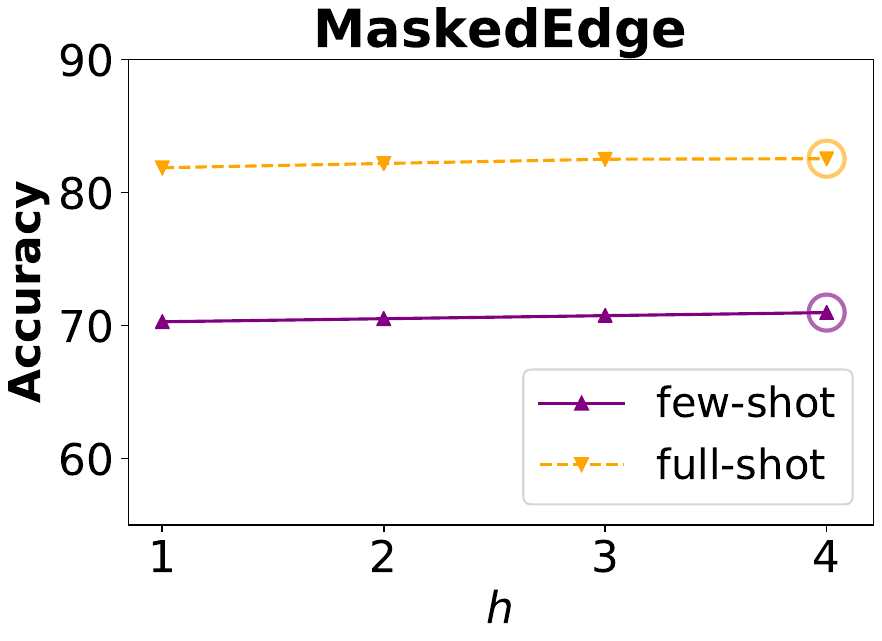}
        } 
    \subfigure {
        \includegraphics[width=0.23\linewidth]{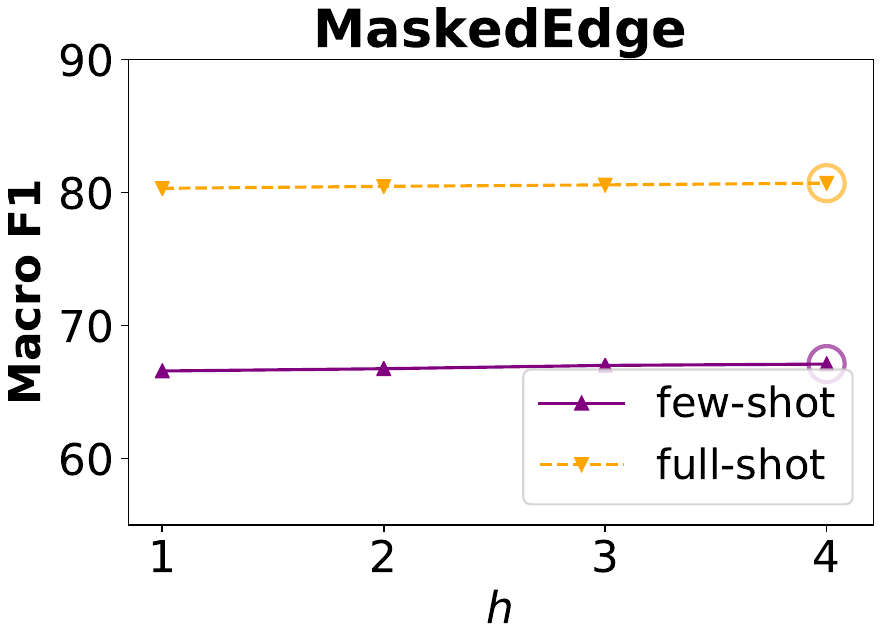}
        }
    \subfigure {
        \includegraphics[width=0.23\linewidth]{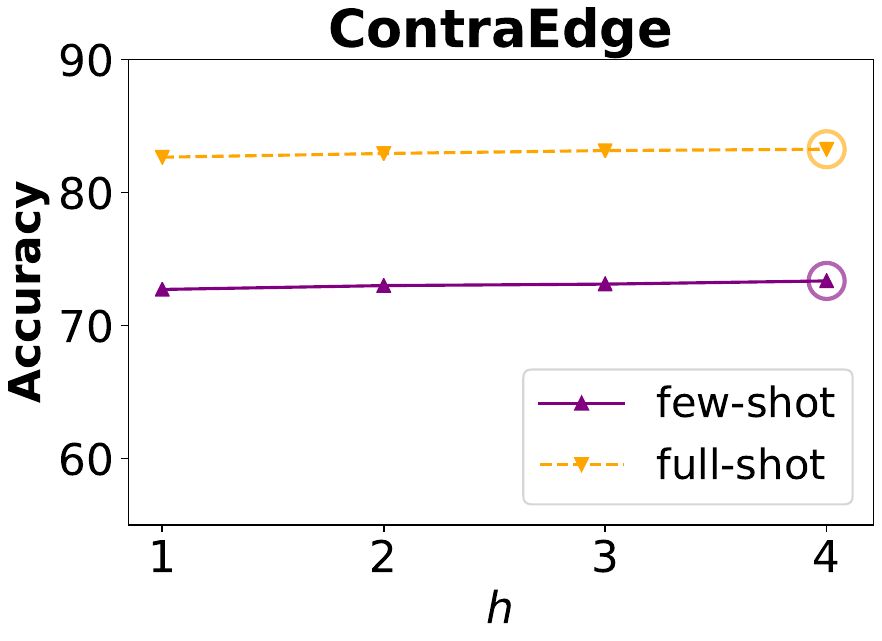}
        }
    \subfigure {
        \includegraphics[width=0.23\linewidth]{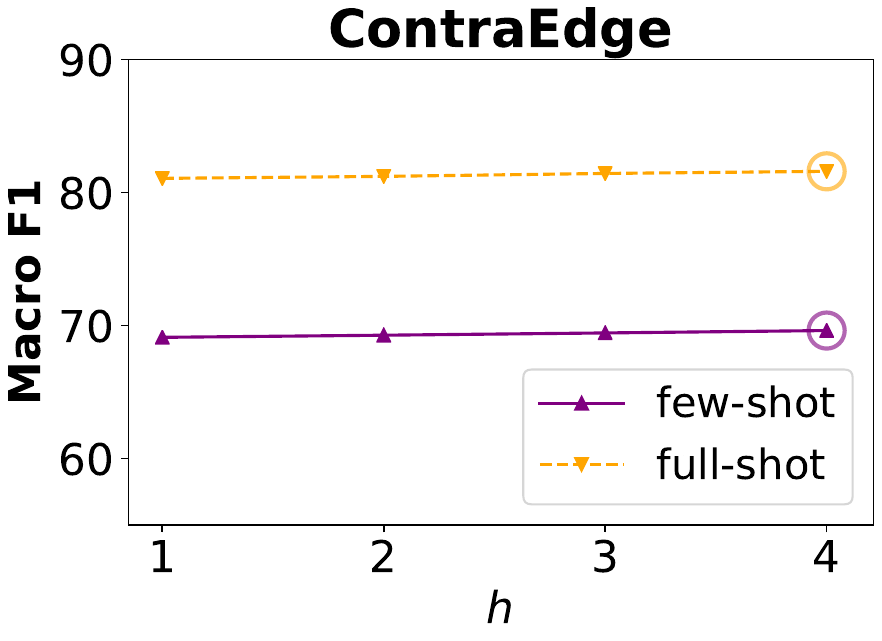}
        } 
    \vskip -0.1in
    \caption{The average Accuracy and Macro F1 across all datasets for the node-level task under different hyper-parameter $h$, with the circle marking the optimal results.}
    \label{fig:h-node}
     \vskip -0.1in
\end{figure*}

From Figure~\ref{fig:h-graph} and Figure~\ref{fig:h-node}, we observe that the optimal value of policy network update interval $h$ is $3$ for the graph-level task and $4$ for the node-level task. Moreover, varying $h$ within reasonable limits has minimal impact on the model's performance, showing LEAP's robustness to the hyper-parameter $h$.

\subsubsection{Policy Finite Horizon (Step) $T$}
\begin{figure*}[htbp]
    \centering
    \vspace{-0.5em}
    \subfigure {
        \includegraphics[width=0.23\linewidth]{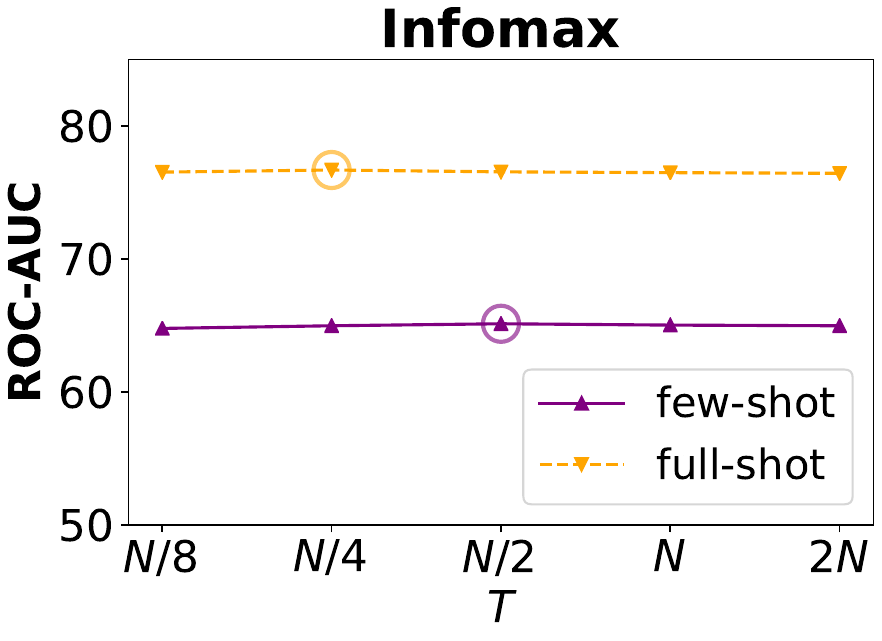}
        } 
    \subfigure {
        \includegraphics[width=0.23\linewidth]{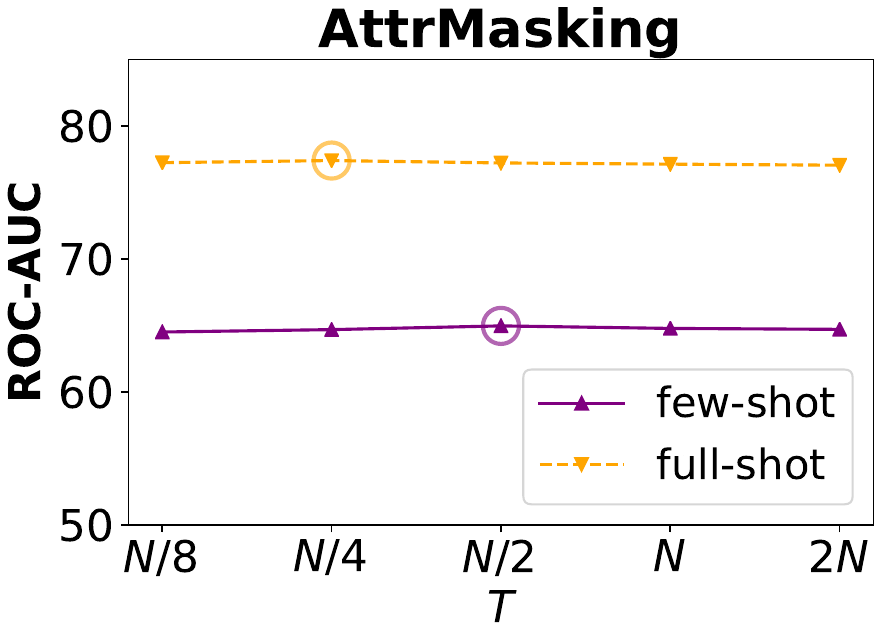}
        }
    \subfigure {
        \includegraphics[width=0.23\linewidth]{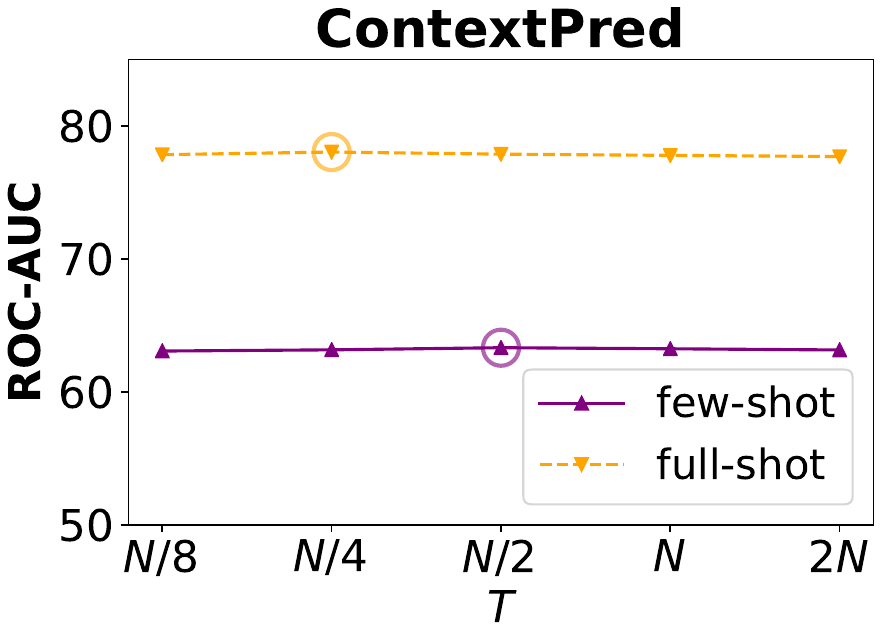}
        }
    \subfigure {
        \includegraphics[width=0.23\linewidth]{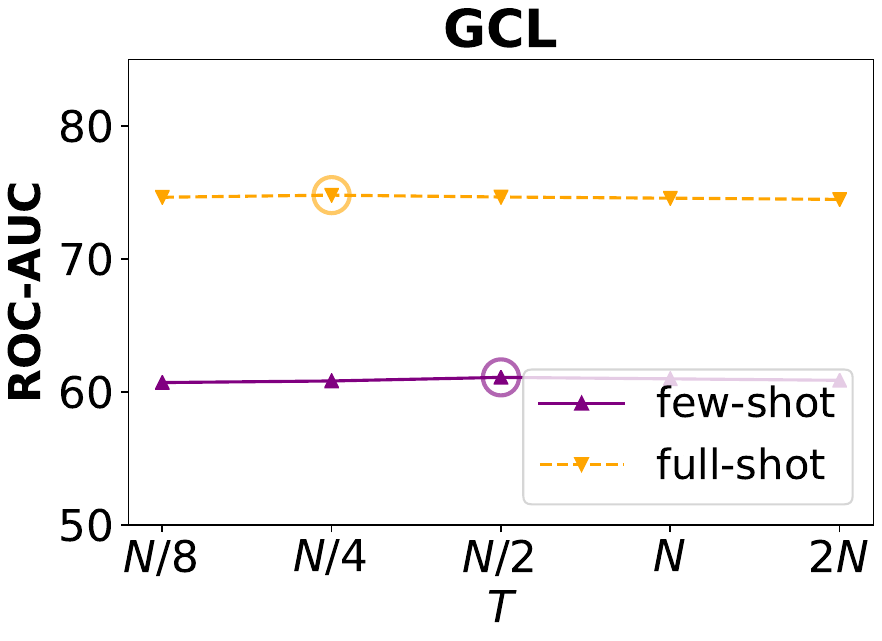}
        } 
    \vskip -0.1in
    \caption{The average ROC-AUC across all datasets for the graph-level task under different hyper-parameter $T$, with the circle marking the optimal results.}
    \label{fig:t-graph}
     \vskip -0.1in
\end{figure*}

\begin{figure*}[!htbp]
    \centering
    \vspace{-0.5em}
    \subfigure {
        \includegraphics[width=0.23\linewidth]{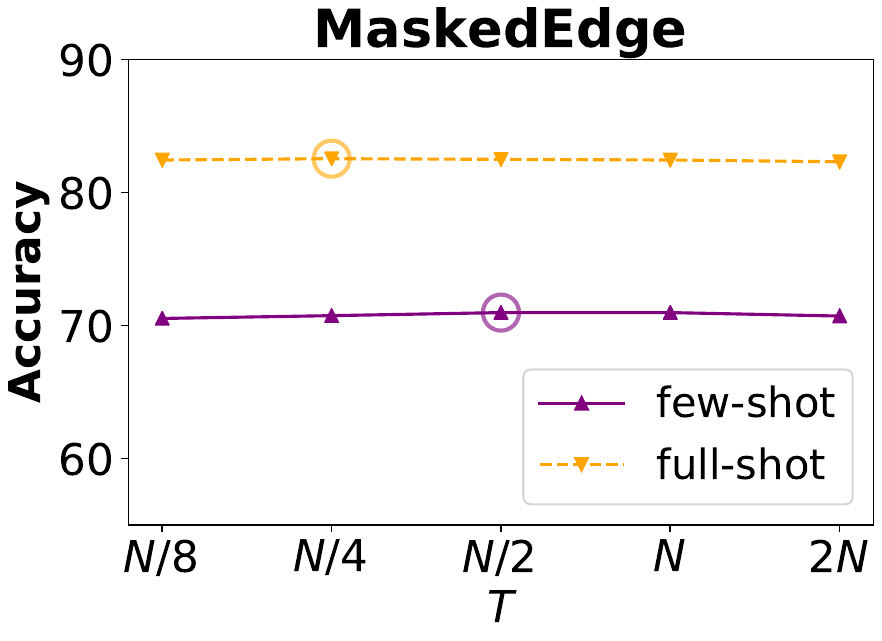}
        } 
    \subfigure {
        \includegraphics[width=0.23\linewidth]{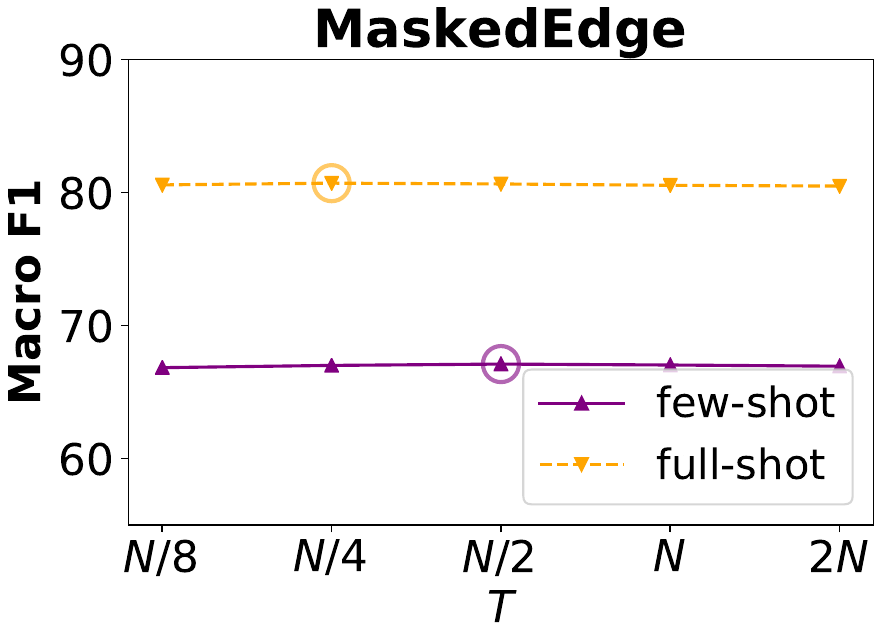}
        }
    \subfigure {
        \includegraphics[width=0.23\linewidth]{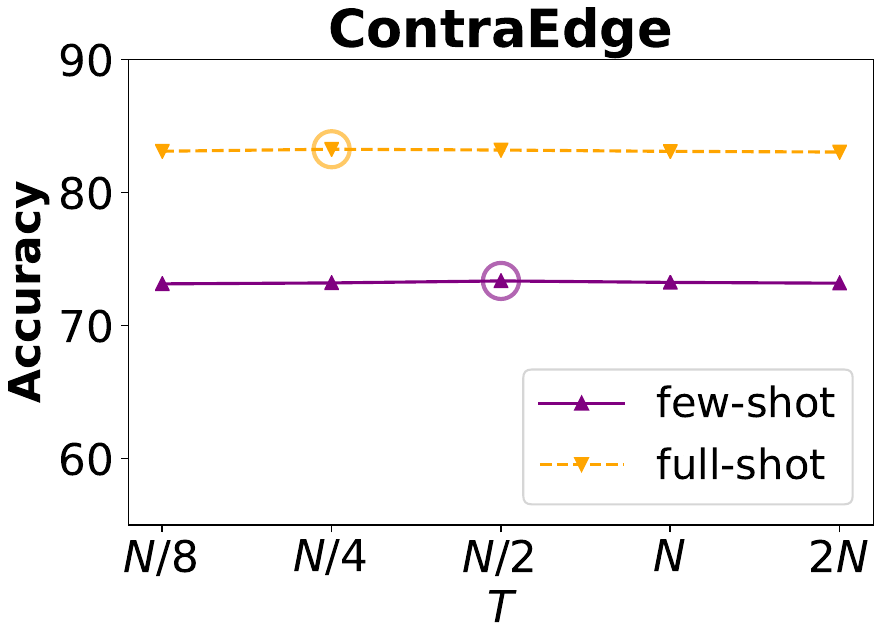}
        }
    \subfigure {
        \includegraphics[width=0.23\linewidth]{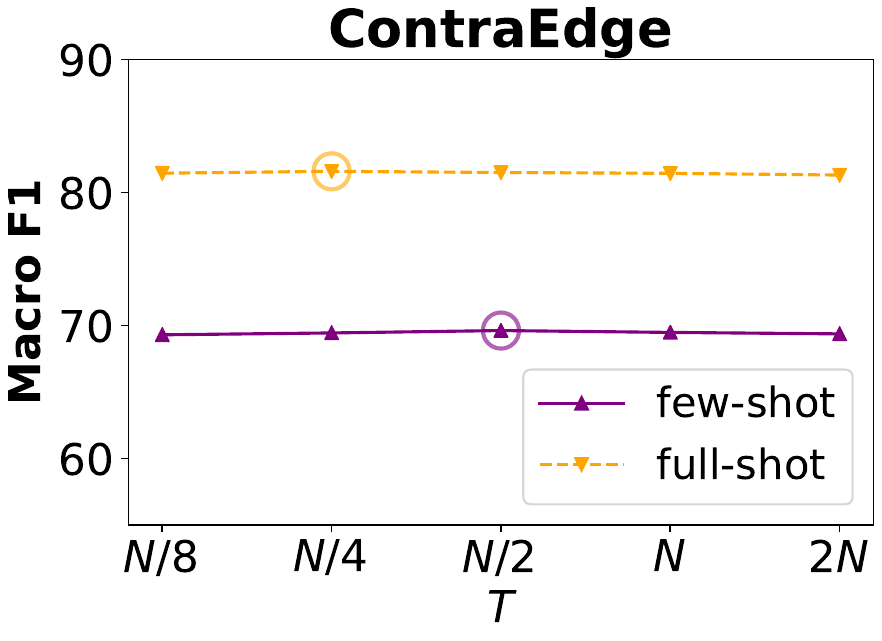}
        } 
    \vskip -0.1in
    \caption{The average Accuracy and Macro F1 across all datasets for the node-level task under different hyper-parameter $T$, with the circle marking the optimal results.}
    \label{fig:t-node}
     \vskip -0.1in
\end{figure*}

From Figure~\ref{fig:t-graph} and Figure~\ref{fig:t-node}, we observe that the optimal value of policy finite horizon (step) $T$ is $N/4$ in the full-shot scenario and $N/2$ in the few-shot scenario. Moreover, varying $T$ within reasonable limits has little effect on the model's performance, demonstrating LEAP's robustness to the hyper-parameter $T$.

\section{Discussion}
\label{appendix:discussion}
We categorize previous works \citep{fang2023universal,lee2024subgraph,zhu2024relief} into two paradigms of universal graph prompt tuning. Specifically, GPF and GPF-plus fall under the \textit{Conventional Graph Prompt Tuning} paradigm, which ensures that the universal graph can theoretically achieve the equivalent effect of any prompting function. However, this paradigm lacks a tailored design for pursuing more ideal prompts. On the other hand, SUPT and RELIEF belong to the \textit{Selective Node-based Graph Prompt Tuning} paradigm, which proposes various tailored designs to select and generate prompts for some nodes. However, as stated in Theorem~\ref{theorem:1}, it compromises the theoretical foundation of universal graph prompt tuning.

Our proposed LEAP represents an ideal paradigm, which we refer to as the \textit{Learning and Editing Graph Prompt Tuning} paradigm. It preserves the theoretical foundation of universal graph prompt tuning by constructing the basic universal graph prompt, while simultaneously pursuing more ideal prompts through prompt editing, effectively addressing the limitations of the previous two paradigms.

In the following, we will analyze the connections between LEAP and these previous works in detail.

\subsection{Compare with GPF and GPF-plus}
Compared to GPF and GPF-plus, LEAP builds upon the construction of a basic universal graph prompt to ensure a consistent theoretical foundation. Beyond this, LEAP further leverages RL to select and edit prompts for some nodes to pursue more ideal prompts. By optimizing prompts through maximizing performance gains and edit convergence rates, LEAP achieves practical performance that more closely approaches the theoretical upper bound.

\subsection{Compare with SUPT}
SUPT generates subgraph-level prompts to learn higher-quality prompts related to the context. However, the prompt is applied only to a subgraph rather than the entire node set of a given graph, which undermines the theoretical foundation of universal graph prompt tuning. In contrast, LEAP ensures the theoretical foundation of universal graph prompt tuning while leveraging RL to edit prompts for higher cumulative rewards.

\subsection{Compare with RELIEF}
RELIEF is inspired by NLP and utilizes RL to select and add prompts for certain nodes. However, as stated in Theorem~\ref{theorem:1}, this approach undermines the theoretical foundation of universal graph prompt tuning in the graph learning domain. Furthermore, as analyzed in Section~\ref{sec:experiments}, RELIEF performs well in few-shot scenarios but exhibits mediocre performance in full-shot settings. To mitigate the overfitting caused by RL, RELIEF employs ensemble actor policy networks to enhance generalization, which inevitably introduces additional computational overhead. In contrast, LEAP's basic universal graph prompt ensures the theoretical foundation of universal graph prompt tuning. Moreover, LEAP introduces a carefully designed reward function that addresses the issue of repeatedly selecting a small subset of nodes through the edit convergence rate. Additionally, unlike RELIEF, LEAP eliminates the need for ensemble actor policy networks to address overfitting, as its basic universal graph prompt and RL are naturally aligned to optimize a shared objective.

\section{Complexity Analysis}
\label{appendix:complexity}
We analyze the complexity of LEAP for one epoch. We list definitions used in our analysis. Specifically, let $L_1$ and $L_2$ denote the layer numbers of the pre-trained graph model and policy networks, respectively. Moreover, the hidden dimensions of the pre-trained graph model and policy networks are defined as $d_1$ and $d_2$, respectively. The projection head is a $L_3$ layer MLP with hidden size $d_3$. $N$ and $E$ represent the maximum number of nodes and edges, respectively. $D$ is the dimension of node features. $m$ is the total number of graphs. 

\subsection{Prompted Graph Building}
We summarize basic universal graph prompt building and RL-driven prompt editing as prompted graph building. First, we build the basic universal graph prompt via $k$ basic vectors and $k$ linear projections, with a complexity of $\mathcal{O}(k^2 N)$. Furthermore, for each step $t$, we get $f_a$ via both discrete and continuous actors, with a complexity of $\mathcal{O}(2 (2d_1 d_2 + L_2 d_2^2))$. Notably, we adopt $T$ steps within each epoch. Therefore, the total time complexity is:
\begin{equation}
    m \cdot (\mathcal{O}(k^2 N) + T \cdot \mathcal{O}(4d_1 d_2 + 2 L_2 d_2^2)). 
\end{equation}

\subsection{Policy Networks Training}
A total of $T$ steps are utilized for sequentially updating the actors and the critic, which primarily involves the training of policy networks based on MLP updates. Notably, the policy networks are updated at intervals of $h$, thereby reducing the per-epoch complexity by a factor of $h$. Therefore, the total time complexity is:
\begin{equation}
    mT \cdot \mathcal{O}(3(2d_1 d_2 + L_2 d_2^2))/h.
\end{equation}

\subsection{Projection Head and Basic Prompt Training}
In this process, we adopt pre-trained graph model to obtain node representations, with a complexity of $\mathcal{O}(L_1(E+ND)d_1)$. Moreover, we feed representations into the projection head, with a complexity of $\mathcal{O}(2d_1 d_3 + L_3 d_3^2)$. Final loss is calculated by label and output from the projection head, with a complexity of $\mathcal{O}(1)$. Therefore, the total time complexity is:
\begin{equation}
     m \cdot (\mathcal{O}(L_1(E + ND)d_1) + \mathcal{O}(2d_1 d_3 + L_3 d_3^2) + \mathcal{O}(1)).
\end{equation}

\subsection{Simplified Complexity}
For simplification, we adopt a unified dimension $d$ for all $d_1$, $d_2$, and $d_3$. Moreover, we adopt a unified layer $L$ for all $L_1$, $L_2$, and $L_3$. Therefore,  the simplified total time complexity is: 
\begin{equation}
     m \cdot (\mathcal{O}(k^2N + Ld(E + ND) + 1) + (\frac{3T}{h} + 2T + 1) \cdot \mathcal{O}((2+L)d^2)).
\end{equation}
In practice, we set $T = N/4$ and $T = N/2$ in the full-shot and few-shot scenarios, respectively. Therefore, in the full-shot scenario, the simplified total time complexity is: 
\begin{equation}
     m \cdot (\mathcal{O}(k^2N + Ld(E + ND) + 1) + (\frac{3N}{4h} + \frac{N}{2} + 1) \cdot \mathcal{O}((2+L)d^2)).
\end{equation}
Moreover, in the few-shot scenario, the simplified total time complexity is: 
\begin{equation}
     m \cdot (\mathcal{O}(k^2N + Ld(E + ND) + 1) + (\frac{3N}{2h} + N + 1) \cdot \mathcal{O}((2+L)d^2)).
\end{equation}
These two simplified total time complexities indicate that the computation time will noticeably increase with the number of large graphs as they scale up.



\section{Related Work}
\label{appendix:related_work}
\subsection{Graph Prompt Tuning}

\noindent \textbf{Task-specific Graph Prompt Tuning}
Task-specific approaches aim to align pretext tasks with downstream objectives. For instance, GPPT \citep{sun2022gppt} and GraphPrompt \citep{liu2023graphprompt} utilize edge prediction to bridge pre-training and downstream tasks, while All in One \citep{sun2023all} employs a prompt graph for graph-level contrastive learning. Similarly, SGL-PT \citep{zhu2023sgl} connects downstream tasks with masked node prediction via contrastive and generative objectives. However, these approaches may face challenges when GNNs are pre-trained with diverse self-supervised tasks through multi-task learning, rather than simpler objectives like link prediction.

\noindent \textbf{Universal Graph Prompt Tuning}
To address the limitations of task-specific methods, universal graph prompt tuning approaches have been introduced, offering compatibility across various pre-training strategies. GPF \citep{fang2023universal} pioneered this direction by demonstrating that adding a learnable, uniform feature prompt vector to each node is theoretically equivalent to any prompting function, irrespective of the pre-training strategy. GPF-plus \citep{fang2023universal} further extended this idea by introducing node-specific prompted features for more fine-grained and adaptable tuning. Recent works have explored selective node-based graph prompt tuning to improve optimization. For example, SUPT \citep{lee2024subgraph} assigns feature prompts at the subgraph level, capturing nuanced contextual information and enhancing performance. RELIEF \citep{zhu2024relief} takes a reinforcement learning approach to dynamically select nodes for prompt addition, enabling high-quality and adaptive prompt tuning. Although these selective node-based graph prompt tuning approaches achieve certain performance improvements, we point out that selective node-based graph prompt tuning undermines the theoretical foundation. This limitation constrains the theoretical representational capacity of universal graph prompt tuning. 

\subsection{RL for Graph Learning}
The integration of RL into graph representation learning has seen notable progress. MAG-GNN \citep{kong2023mag} leverages DQN to select optimal subgraph subsets, enhancing graph expressivity. WSD \citep{wang2023reinforcement} employs weighted sampling for subgraph counting, with edge weights determined using DDPG. SUGAR \citep{sun2021sugar} maintains hierarchical graph properties by selecting subgraphs via an RL-based pooling mechanism grounded in Q-learning. GPA \citep{hu2020graph} utilizes DQN to learn optimal annotation strategies for identifying valuable nodes in active search, while GraphCBAL \citep{yu2024graphcbal} extends this by employing A2C for a class-balanced active search strategy. Additionally, RELIEF \citep{zhu2024relief} incorporates an RL algorithm specifically designed for hybrid action spaces, integrating policy generalization techniques to select subsets of nodes and effectively incorporate prompts. However, RL inevitably introduces extra computational overhead to graph learning. To address these limitations, we propose the Learning and Editing Universal Prompt Tuning paradigm, which combines basic universal graph prompt learning with RL-driven editing to optimize prompts while maintaining a solid theoretical foundation and incurring affordable computational overhead.

\section{Proofs}
\label{appendix:proofs}
\subsection{Proof of Theorem~\ref{theorem:1}}
\noindent \textbf{Proof Sketch} The proof consists of two parts:
\begin{itemize}[leftmargin=*]
    \item \textbf{Sufficiency}: If every node is assigned a prompt, universal graph prompt tuning can simulate any prompting function $\psi(\cdot)$.
    \item \textbf{Necessity}: If any node lacks a prompt, there exists at least one $\psi(\cdot)$ that cannot be simulated.
\end{itemize}

\subsubsection{Sufficiency}

Assume each node $v_i$ is assigned a prompt $p_i$. Following the proof of the Theorem in the GPF, any prompting function $\psi(\cdot)$ can be decomposed into three atomic operations:
\begin{itemize}[leftmargin=*]
    \item \textbf{Feature Modification:} Modifying node features $\mathbf{X} \rightarrow \mathbf{\hat{X}}$.
    \item \textbf{Structure Modification:} Modifying adjacency matrix $\mathbf{A} \rightarrow \mathbf{\hat{A}}$.
    \item \textbf{Component Modification:} Adding or removing isolated components (sub-graphs) and generating the new adjacency matrix and feature matrix $(\mathbf{X}, \mathbf{A}) \rightarrow (\mathbf{\hat{X}}, \mathbf{\hat{A}})$.
\end{itemize}

Simulate each operation via universal graph prompts (For analytical simplicity, we initially assume that $f_\theta$ is a single-layer GNN with a linear transformation. Subsequently, we extend our conclusions to multi-layer models utilizing various transition matrices):

\noindent \underline{Case 1: Feature Modification:}

Set $p_i=\Delta \mathbf{X}_{[i,:]}$ for all $i$. Then:
\begin{equation}
\mathbf{X}+\{p_1, \ldots, p_N\}=\mathbf{X}+\Delta \mathbf{X}=\mathbf{\hat{X}}.
\end{equation}
This trivially replicates any prompting function $\psi(\cdot)$.

\noindent \underline{Case 2: Structure Modification:}

Structural changes in $\mathbf{A}$ affect the diffusion matrix \citep{gasteiger2019diffusion} $\mathbf{S}=\mathbf{A}+(1+\epsilon) \mathbf{I}$. Let $\mathbf{\hat{S}}=\mathbf{\hat{A}}+(1+\epsilon) \mathbf{I}$. The output of $f_\theta$ under $\mathbf{\hat{A}}$ is:
\begin{equation}
f_\theta(\mathbf{\hat{A}}, \mathbf{X})=\mathbf{\hat{S}} \mathbf{X} \mathbf{W}=\mathbf{S X} \mathbf{W}+\Delta \mathbf{S X} \mathbf{W}, \quad \Delta \mathbf{S}=\mathbf{\hat{S}}-\mathbf{S},
\end{equation}
where $\mathbf{W}$ is a frozen linear transformation. The parameters $\epsilon$ and $\mathbf{W}$ have been pre-trained in advance and remain fixed during downstream tuning. To simulate this with prompts, solve:
\begin{equation}
\mathbf{S}(\mathbf{X}+p) \mathbf{W}=\mathbf{S} \mathbf{X} \mathbf{W}+\Delta \mathbf{S X} \mathbf{W}.
\end{equation}
Simplify to:
\begin{equation}
\mathbf{S}p \mathbf{W}=\Delta \mathbf{S X W} .
\end{equation}
Since $\mathbf{W}$ is full-rank (pre-trained), we require:
\begin{equation}
\mathbf{S}p=\Delta \mathbf{S X} .
\end{equation}
Let $p=\{p_1, \ldots, p_N\}$. For each node $i$:
\begin{equation}
\sum_j \mathbf{S}_{[i, j]} p_j=\sum_j \Delta \mathbf{S}_{[i, j]} \mathbf{X}_{[j, :]}.
\end{equation}
This is a linear system with $N \times D$ variables $(p_j)$ and $N \times D$ equations. A solution exists if the system is consistent.

\noindent \underline{Case 3: Component Modification:}

Suppose $\psi(\cdot)$ adds a disconnected subgraph $\mathcal{C}=(\mathbf{A}_c, \mathbf{X}_c)$. Let $\mathbf{\hat{A}}=\left[\begin{array}{cc}\mathbf{A} & 0 \\ 0 & \mathbf{A}_c\end{array}\right], \mathbf{\hat{X}}=\left[\begin{array}{l}\mathbf{X} \\ \mathbf{X}_c\end{array}\right]$. The output of $f_\theta$ is:
\begin{equation}
f_\theta(\mathbf{A}, \mathbf{X})=\mathbf{S X W}, \quad \mathbf{S}=\mathbf{A}+(1+\epsilon) \mathbf{I}.
\end{equation}
The target output after adding $\mathcal{C}$ is:
\begin{equation}
f_\theta(\mathbf{\hat{A}}, \mathbf{\hat{X}})=\mathbf{\hat{S}} \mathbf{\hat{X}} \mathbf{W}, \quad \mathbf{\hat{S}}=\left[\begin{array}{cc}
\mathbf{S} & 0 \\
0 & \mathbf{S}_c
\end{array}\right], \quad \mathbf{S}_c=\mathbf{A}_c+(1+\epsilon) \mathbf{I}.
\end{equation}
To replicate this with prompts, we require:
\begin{equation}
\mathbf{S}(\mathbf{X}+p) \mathbf{W}=\mathbf{\hat{S}} \mathbf{\hat{X}} \mathbf{W}.
\end{equation}
Cancel $\mathbf{W}$ (full-rank) and expand both sides:
\begin{equation}
\mathbf{S X}+\mathbf{S}p=\left[\begin{array}{c}
\mathbf{S X} \\
\mathbf{S}_c \mathbf{X}_c
\end{array}\right].
\end{equation}
This implies:
\begin{equation}
\mathbf{S}p=\left[\begin{array}{c}
0 \\
\mathbf{S}_c \mathbf{X}_c
\end{array}\right] .
\end{equation}
For existing nodes $v_i \in \mathcal{V}$:
\begin{equation}
\sum_j \mathbf{s}_{[i, j]} p_j=0 \quad \Longrightarrow \quad p_j=0 \quad \text { (since } \mathbf{S} \text { is invertible). }
\end{equation}
For virtual nodes $v_j \in \mathcal{C}$. To simulate $\mathbf{S}_c \mathbf{X}_c$, define prompts for virtual nodes as:
\begin{equation}
p_j=\mathbf{X}_{c[j,:]} \quad(\text { for } j=N+1, \ldots, N+M) ,
\end{equation}
where $M$ denotes the number of virtual nodes $v_j \in \mathcal{C}$. This is a system of $M \times D$ equations. A solution exists if the prompt dimension $D$ is sufficiently large (e.g., $D \geq M$).

\noindent \textbf{Extension to Multi-Layer Scenario}

The final node representation after $L$ layers is:
\begin{equation}
\mathbf{H}_{(L)}=\underbrace{\mathbf{S}_{(L)} \mathbf{S}_{(L-1)} \cdots \mathbf{S}_{(1)}}_{\mathbf{S}^{\prime}} \mathbf{X} \underbrace{\mathbf{W}_{(1)} \mathbf{W}_{(2)} \cdots \mathbf{W}_{(L)}}_{\mathbf{W}^{\prime}}.
\end{equation}
Let $\mathbf{S}^{\prime}=\prod_{l=1}^L \mathbf{S}_{(l)}$ and $\mathbf{W}^{\prime}=\prod_{l=1}^L \mathbf{W}_{(l)}$. The output simplifies to:
\begin{equation}
\mathbf{H}_{(L)}=\mathbf{S}^{\prime} \mathbf{X} \mathbf{W}^{\prime} .
\end{equation}

\noindent \underline{Case 1: Feature Modification:}

Let $\mathbf{\hat{X}}=\mathbf{X}+\Delta \mathbf{X}$. The target output is:
\begin{equation}
\mathbf{\hat{H}}_{(L)}=\mathbf{S}^{\prime} \mathbf{\hat{X}} \mathbf{W}^{\prime}=\mathbf{S}^{\prime}(\mathbf{X}+\Delta \mathbf{X}) \mathbf{W}^{\prime}.
\end{equation}
Achieving equivalence via prompts needs to satisfy the following:
\begin{equation}
\mathbf{S}^{\prime}(\mathbf{X}+p) \mathbf{W}^{\prime}=\mathbf{S}^{\prime}(\mathbf{X}+\Delta \mathbf{X}) \mathbf{W}^{\prime}.
\end{equation}
We require:
\begin{equation}
\mathbf{S}^{\prime} p=\mathbf{S}^{\prime} \Delta \mathbf{X}.
\end{equation}
Since $\mathbf{S}^{\prime}$ is invertible (diagonally dominant), the solution is:
\begin{equation}
p=\Delta \mathbf{X} .
\end{equation}
Thus, setting $p_i=\Delta \mathbf{X}_{[i,:]}$ for all $i$ replicates the feature modification.

\noindent \underline{Case 2: Structure Modification:}

Let $\mathbf{S}_{(l)}^{\prime}=\mathbf{\hat{A}}+(1+\epsilon_l) \mathbf{I}$. The target output is:
\begin{equation}
\mathbf{H}_{(L)}^{\prime}=\left(\prod_{l=1}^N \mathbf{S}_{(l)}^{\prime}\right) \mathbf{X} \mathbf{W}^{\prime} \triangleq \mathbf{S}^{\prime \prime} \mathbf{X} \mathbf{W}^{\prime} .
\end{equation}
To replicate $\mathbf{H}_{(L)}^{\prime}$, solve:
\begin{equation}
\mathbf{S}^{\prime}(\mathbf{X}+p) \mathbf{W}^{\prime}=\mathbf{S}^{\prime \prime} \mathbf{X} \mathbf{W}^{\prime} .
\end{equation}
Canceling $\mathbf{W}^{\prime}$:
\begin{equation}
\mathbf{S}^{\prime} p=(\mathbf{S}^{\prime \prime}-\mathbf{S}^{\prime}) \mathbf{X}
\end{equation}
For each node $i$:
\begin{equation}
\sum_j \mathbf{S}_{[i, j]}^{\prime} p_j=\sum_j(\mathbf{S}_{[i, j]}^{\prime \prime}-\mathbf{S}_{[i, j]}^{\prime}) \mathbf{X}_{[j,:]} .
\end{equation}
This linear system has a unique solution $p$ due to $\mathbf{S}^{\prime}$'s invertibility.

\noindent \underline{Case 3: Component Modification:}

Suppose $\psi(\cdot)$ adds a disconnected subgraph $\mathcal{C}=(\mathbf{A}_c, \mathbf{X}_c)$. Let $\mathbf{\hat{A}}=\left[\begin{array}{cc}\mathbf{A} & 0 \\ 0 & \mathbf{A}_c\end{array}\right], \mathbf{\hat{X}}=\left[\begin{array}{l}\mathbf{X} \\ \mathbf{X}_c\end{array}\right]$. The output is:
\begin{equation}
\mathbf{\hat{H}}_{(L)}=\mathbf{\hat{S}}^{\prime} \mathbf{\hat{X}} \mathbf{W}^{\prime}=\left[\begin{array}{c}
\mathbf{S}^{\prime} \mathbf{X} \\
\mathbf{S}_c^{\prime} \mathbf{X}
\end{array}\right] \mathbf{W}^{\prime},
\end{equation}
where $\mathbf{S}_c^{\prime}=\prod_{l=1}^L(\mathbf{A}_c+\left(1+\epsilon_l\right) \mathbf{I})$.
To simulate $\mathbf{\hat{H}}_{(L)}$, enforce:
\begin{equation}
\mathbf{S}^{\prime}(\mathbf{X}+p) \mathbf{W}^{\prime}=\left[\begin{array}{c}
\mathbf{S}^{\prime} \mathbf{X} \\
\mathbf{S}_c^{\prime} \mathbf{X}_c
\end{array}\right] \mathbf{W}^{\prime} .
\end{equation}
Canceling $\mathbf{W}^{\prime}$:
\begin{equation}
\mathbf{S}^{\prime} p=\left[\begin{array}{c}
0 \\
\mathbf{S}_c^{\prime} \mathbf{X}_c
\end{array}\right] .
\end{equation}
Therefore, for existing nodes $v_i \in \mathcal{V}$, set $p_i=0$. For virtual nodes $v_j \in \mathcal{C}$, inject their impact into existing nodes by solving:
\begin{equation}
\sum_{i=1}^N \mathbf{S}_{[k, i]}^{\prime} p_i=\mathbf{S}_c^{\prime} \mathbf{X}_{c[j,:]} \quad(j=N+1, \ldots, N+M)
\end{equation}
This distributes the subgraph’s effect across the original graph through prompts.

\subsubsection{Necessity}
Prove that if any node lacks a prompt $(\exists j, p_j=0)$, there exists $\psi(\cdot)$ that cannot be simulated. 
First, we define a specific $\psi(\cdot)$ as: 1) Add an edge $\left(v_j, v_k\right)$ to $\mathbf{A}$, resulting in $\mathbf{\hat{A}}$. 2) Modify $\mathbf{X}_{[j,:]} \rightarrow \mathbf{X}_{[j,:]}+\delta$, where $\delta \neq 0$. To simulate $\psi(\cdot)$ by universal graph prompts, we need:
\begin{equation}
f_\theta(\mathbf{A}, \mathbf{X} + p)=f_\theta(\mathbf{\hat{A}}, \mathbf{X}+\delta \mathbf{e}_j),
\end{equation}
where $\mathbf{e}_j$ is a one-hot vector. Expanding both sides:
\begin{equation}
f_\theta(\mathbf{A}, \mathbf{X} + p)=\mathbf{S}(\mathbf{X}+p)\mathbf{W} = (\mathbf{S}+\Delta \mathbf{S})\left(\mathbf{X}+\delta \mathbf{e}_j\right) \mathbf{W} = f_\theta(\mathbf{\hat{A}}, \mathbf{X}+\delta \mathbf{e}_j).
\end{equation}
Equating the two:
\begin{equation}
\mathbf{S}p\mathbf{W}=\Delta \mathbf{S X W}+\Delta \mathbf{S} \delta \mathbf{e}_j \mathbf{W}+\mathbf{S} \delta \mathbf{e}_j \mathbf{W}.
\end{equation}
Dividing both sides by $\mathbf{W}$ (full-rank):
\begin{equation}
\mathbf{S} p=\Delta \mathbf{S X}+(\Delta \mathbf{S}+\mathbf{S}) \delta \mathbf{e}_j.
\end{equation}
If $p_j=0$ ($j$-th row of $p$ is zero). From the equation $\mathbf{S}p=\Delta \mathbf{S X}+(\Delta \mathbf{S}+\mathbf{S}) \delta \mathbf{e}_j$, the $j$-th row gives:
\begin{equation}
\sum_k \mathbf{S}_{[j, k]} p_k=\sum_k \Delta \mathbf{S}_{[j, k]} \mathbf{X}_{[k, j]}+(\Delta \mathbf{S}_{[j, j]}+\mathbf{S}_{[j, j]}) \delta.
\end{equation}
Since $p_k=0$ for all $k$, the left-hand side is zero. However, 1)$\Delta \mathbf{S}_{[j, j]}=0$ (adding an edge does not change self-loops) and 2) $\mathbf{S}_{[j, j]}=1+\epsilon \neq 0$. Thus:
\begin{equation}
0=\sum_k \Delta \mathbf{S}_{[j, k]} \mathbf{X}_{[k,:]}+(1+\epsilon) \delta.
\end{equation}
This implies $\delta=-\frac{1}{1+\epsilon} \sum_k \Delta \mathbf{S}_{[j, k]} \mathbf{X}_{[k, :]}$, which contradicts the arbitrary choice of $\delta \neq 0$.

\noindent \textbf{Extension to Multi-Layer Scenario}

Construct $\psi(\cdot)$ by: 1) Add an edge $(v_j, v_k)$ to $\mathbf{A}$, resulting in $\mathbf{\hat{A}}$ and modifying all $\mathbf{S}_{(l)}$ to $\mathbf{S}_{(l)}^{\prime}=\mathbf{\hat{A}}+(1+\epsilon_l) \mathbf{I}$. 2) Modify node features: $\mathbf{X}_{[j,:]} \rightarrow \mathbf{X}_{[j,:]}+\delta$, where $\delta \neq 0$. The output is:
\begin{equation}
\mathbf{H}_{(L)}^{\prime}=\left(\prod_{l=1}^L \mathbf{S}_{(l)}^{\prime}\right)(\mathbf{X}+\delta \mathbf{e}_j).
\end{equation}
To simulate $\mathbf{H}_{(L)}^{\prime}$, enforce:
\begin{equation}
\underbrace{\left(\prod_{l=1}^L \mathbf{S}_{(l)}\right)}_{S^{\prime}}(\mathbf{X}+p).
\end{equation}
Canceling $\mathbf{W}^{\prime}$ (full-rank):
\begin{equation}
\mathbf{S}^{\prime}(\mathbf{X}+p)=\left(\prod_{l=1}^L \mathbf{S}_{(l)}^{\prime}\right)(\mathbf{X}+\delta \mathbf{e}_j).
\end{equation}
Expanding the right-hand side:
\begin{equation}
\mathbf{S}^{\prime \prime} \mathbf{X}+\mathbf{S}^{\prime \prime} \delta \mathbf{e}_j, \quad \mathbf{S}^{\prime \prime}=\prod_{l=1}^L \mathbf{S}_{(l)}^{\prime}.
\end{equation}
Rearranging terms:
\begin{equation}
\mathbf{S}^{\prime} p=\left(\mathbf{S}^{\prime \prime}-\mathbf{S}^{\prime}\right) \mathbf{X}+\mathbf{S}^{\prime \prime} \delta \mathbf{e}_j .
\end{equation}
For node $v_j$, the $j$-th row equation is:
\begin{equation}
\sum_{n=1}^N \mathbf{S}_{[j, n]}^{\prime} \boldsymbol{p}_n=\sum_{n=1}^N\left(\mathbf{S}_{[j, n]}^{\prime \prime}-\mathbf{S}_{[j, n]}^{\prime}\right) \mathbf{X}_{[n,:]}+\mathbf{S}_{[j, j]}^{\prime \prime} \delta .
\end{equation}
If node $v_j$ lacks a prompt $(p_j=0)$, the equation reduces to:
\begin{equation}
\sum_{n \neq j} \mathbf{S}_{[j, n]}^{\prime} p_n=\sum_{n=1}^N\left(\mathbf{S}_{[j, n]}^{\prime \prime}-\mathbf{S}_{[j, n]}^{\prime}\right) \mathbf{X}_{[n,:]}+\mathbf{S}_{[j, j]}^{\prime \prime} \delta .
\end{equation}
This implies:
\begin{equation}
\mathbf{S}_{[j, j]}^{\prime \prime} \delta=-\sum_{n=1}^N\left(\mathbf{S}_{[j, n]}^{\prime \prime}-\mathbf{S}_{[j, n]}^{\prime}\right) \mathbf{X}_{[n, j]},
\end{equation}
which forces $\delta$ to depend on $\mathbf{X}$, contradicting the arbitrary choice of $\delta \neq 0$.

\subsection{Connection to Recent Theoretical Work}

A recent theoretical work \cite{wang2025does}, which proves that any prompt design preserves the GPF foundation as long as the model’s weight matrix is row full-rank. In practice, however, pre-trained GNNs seldom satisfy the row full-rank condition. Although that study offers solid theoretical insights, it provides limited practical guidance for future research directions.

Our Theorem~\ref{theorem:1} and their results mutually reinforce each other. Specifically, when the pre-trained graph does not satisfy the row full-rank condition, selectively adding prompts to only a subset of nodes may prevent the model from achieving row full-rank. In contrast, if prompts are added to all nodes, a solution theoretically exists that ensures the graph satisfies the row full-rank condition, thereby preserving the GPF foundation. Our "Learning and Editing" paradigm offers a concrete and actionable solution aligned with this theoretical insight.


\end{document}